\documentclass{article} 
\usepackage{iclr2024_conference_tinypaper,times}

\usepackage{amsmath,amsfonts,bm}

\def\eqref#1{equation~\ref{#1}}

\def\1{\bm{1}}

\DeclareMathAlphabet{\mathsfit}{\encodingdefault}{\sfdefault}{m}{sl}
\SetMathAlphabet{\mathsfit}{bold}{\encodingdefault}{\sfdefault}{bx}{n}

\usepackage{hyperref}
\usepackage{url}
\usepackage{graphicx}
\usepackage{float}

\title{Toward Learning Latent-Variable Representations of Microstructures by Optimizing in Spatial Statistics Space}

\iclrfinalcopy

\author{Sayed Sajad Hashemi   \\
Dept. of Mechanical \& Industrial Engineering \\
University of Toronto\\
\texttt{sajjads.hashemi@gmail.com} \\
\And
Michael Guerzhoy \\
Division of Engineering Science and \\
Dept. of Mechanical \& Industrial Engineering \\
University of Toronto\\
\texttt{guerzhoy@cs.toronto.edu} \\
\And
Noah Paulson \\
Argonne National Laboratory \\
\texttt{npaulson@anl.gov} \\
}

\begin{document}

\maketitle
\begin{abstract} 
In Materials Science, material development involves evaluating and optimizing the internal structures of the material, generically referred to as microstructures. Microstructures structure is stochastic, analogously to image textures. A particular microstructure can be well characterized by its \textit{spatial statistics}~\cite{paulson2017reduced}, analogously to image texture being characterized by the response to a Fourier-like filter bank~\cite{varma2002classifying}.  Material design would benefit from low-dimensional representation of microstructures~\cite{paulson2017reduced}. 

In this work, we train a Variational Autoencoders (VAE) to produce reconstructions of textures that preserve the spatial statistics of the original texture,  while not necessarily reconstructing the same image in data space. We accomplish this by adding a differentiable term to the cost function in order to minimize the distance between the original and the reconstruction in spatial statistics space. 

Our experiments indicate that it is possible to train a VAE that minimizes the distance in spatial statistics space between the original and the reconstruction of synthetic images. In future work, we will apply the same techniques to microstructures, with the goal of obtaining low-dimensional representations of material microstructures.

\end{abstract}

\section{Introduction}

Computer-aided materials design holds the promise of designing materials with favorable properties by enabling researchers to computationally optimize potential materials. 

Material microstructure is often represented as a large 3D image. 2D images are often also used. In research, both real images (e.g., X-ray) and simulated microstructures are used. 

In ``data space," the dimensionality of the microstructure representation is the number of pixels/voxels, which can be quite large. This project is motivated by the need for working in low-dimensional latent space when optimizing material properties. Previous work includes representing the microstructure using the first principal components of the spatial statistics representation~\cite{paulson2017reduced}. 

In this paper, we modify the Variational Autoencoder (VAE)~\cite{kingma2014auto} to learn a low-dimensional representation that preserves the statistical properties of the material by making the original and the reconstruction be similar in spatial statistics space. We work with 2D images.

\section{Background}
In Materials Science, the microstructure $m$ is often characterized using  spatial statistics~\cite{cecen2016versatile}, 
$f\left(h, h^{\prime} \mid \boldsymbol{r}\right)=\frac{1}{\operatorname{Vol}\left(\Omega_{\boldsymbol{r}}\right)} \int_{\Omega_{\boldsymbol{r}}} m(h, \boldsymbol{x}) m\left(h^{\prime}, \boldsymbol{x}+\boldsymbol{r}\right) d \boldsymbol{x}.$

Here, $m(h, \boldsymbol{x}$) is the local state $h$ (e.g., the crystal lattice orientation) at location $\boldsymbol{x}$ in the microstructure. The spatial statistics (or correlations) can be interpreted as the set of correlations between all locations at distance $r$, in states $h$ and $h'$ respectively. In practice, the spatial statistics can be computed as the inverse Fourier transform of the magnitudes of the Fourier transform of the microstructure~\cite{paulson2017reduced}~\cite{einstein1914methode}.

Generation of images with a specific texture that is specified by the response to a Fourier-like filter bank (or indeed the Fourier transform) goes back decades~\cite{matsuyama1983structural}. Recently, the literature on neural style transfer~\cite{gatys2016image} constrained the image generator with two cost functions: one that controlled the content and one that controlled the style. The style cost function constrained the response to a ConvNet in the lower layers, some of which are known to be Fourier-response-like~\cite{zeiler2014visualizing}.
                 
Prior work includes incorporating the loss from the neural style transfer literature to train a VAE that reconstructs microstructure~\cite{sardeshmukh2021texturevae} and using a Generative Adversarial Network (GAN) when training a VAE to produce reconstructions of microstructure that preserve the texture~\cite{zhang2024vegan}. Unlike previous work, we directly optimize in spatial statistics space. We aim to use limited embedding space to only store texture information.

Embedding microstructure in a low-dimensional space while preserving microstructure information is useful since it can enable researchers to rapidly explore different microstructures in the process of materials design~\cite{kalidindi2015hierarchical}~\cite{mcdowell2009integrated}~\cite{sundararaghavan2009statistical}. 

\section{Methods}
We train a VAE to reconstruct images of texture such that, in spatial statistics space, the original and the reconstruction are close. That is, the VAE loss we use is

$\mathcal{L} = \alpha \cdot ||f(\boldsymbol{x}) - f(\boldsymbol{x}_{recon})||_2 + \beta \cdot \text{Loss}_{KL}$.

That is, instead of minimizing the distance between the input $x$ and the reconstruction $\boldsymbol{x}_{recon}$, we minimize the distance between the spatial statistics $f$ of $\boldsymbol{x}$ and $\boldsymbol{x}_{recon}$. The spatial statistics function can be efficiently computed using FFT and is differentiable.

\section{Results}


On a dataset of images of 100,000 randomly-placed vertical and horizontal lines of various lengths (details in Appendix A), we show that our network successfully produces reconstructions that are close to the original in spatial statistics space, but not necessarily in data space, and does so better than the baseline VAE. Qualitative and quantitative comparisons are in Appendix B. More samples are in Appendix C.

\begin{table}[H]
\begin{center}
\begin{tabular}{cccc}
\hline
\multicolumn{2}{c}{\bf Data} & \multicolumn{2}{c}{\bf Spatial statistics} \\
\hline
{\bf Original} & {\bf Reconstructed} & {\bf Original} & {\bf Reconstructed} \\
\hline
\includegraphics[width=0.15\linewidth]{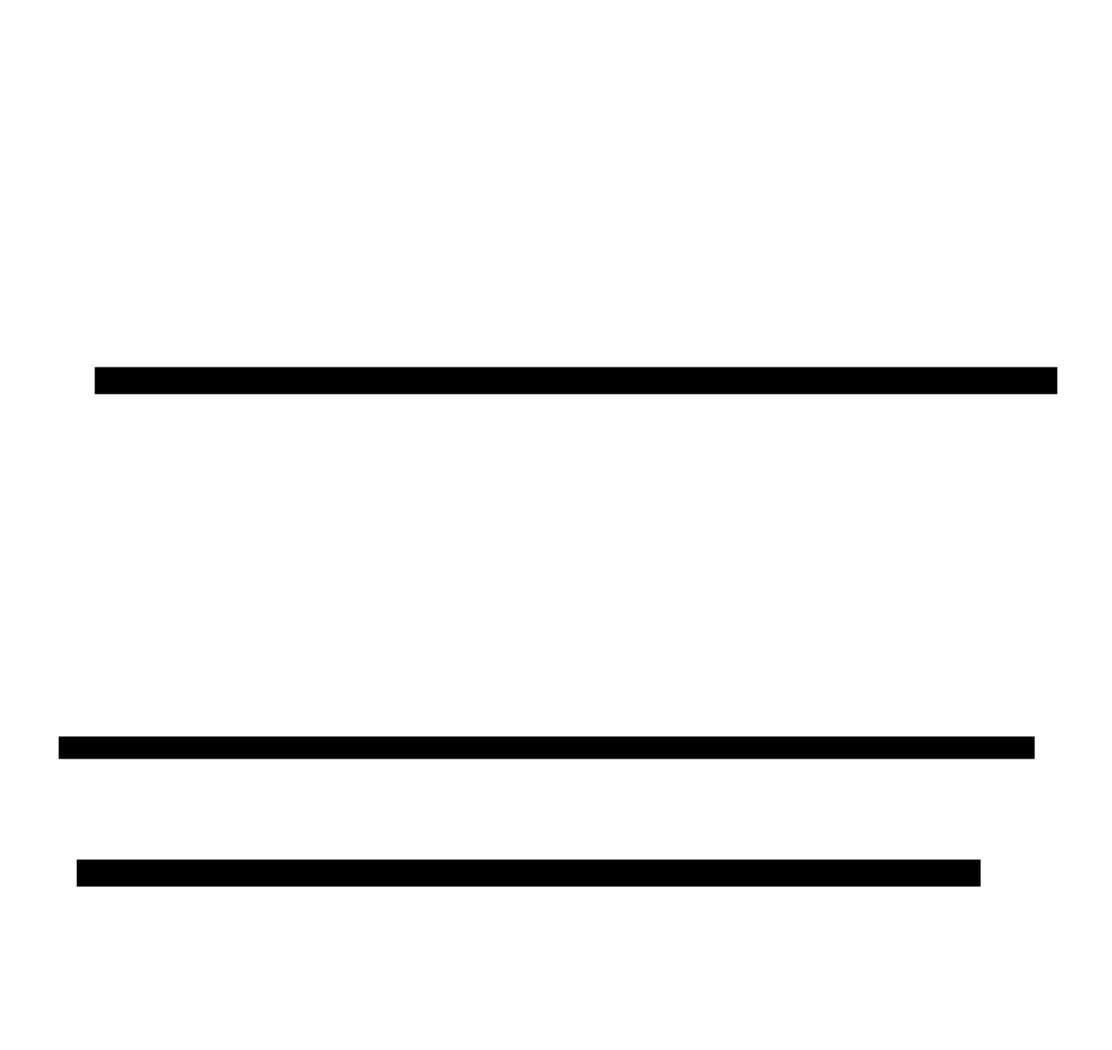} & \includegraphics[width=0.15\linewidth]{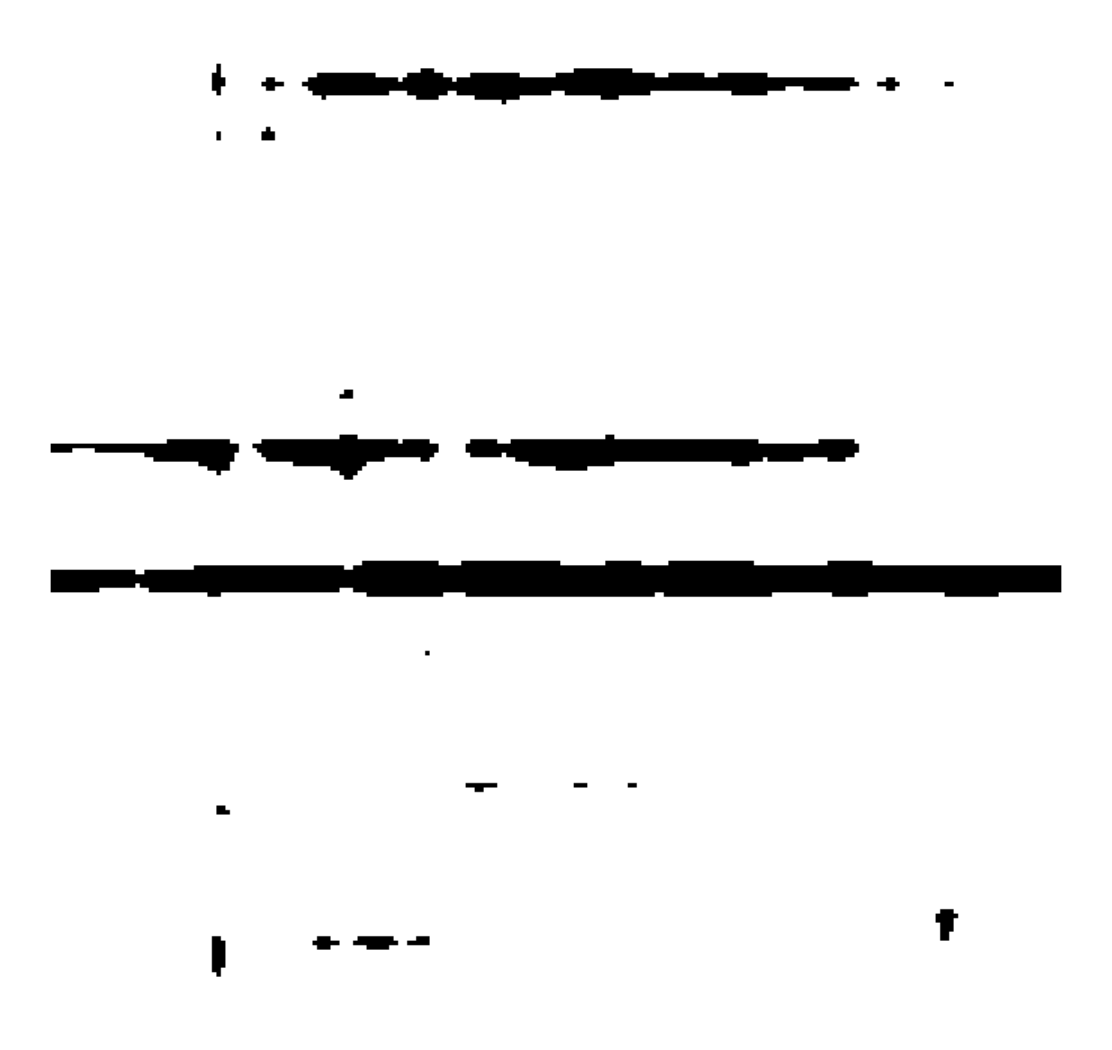} & \includegraphics[width=0.15\linewidth]{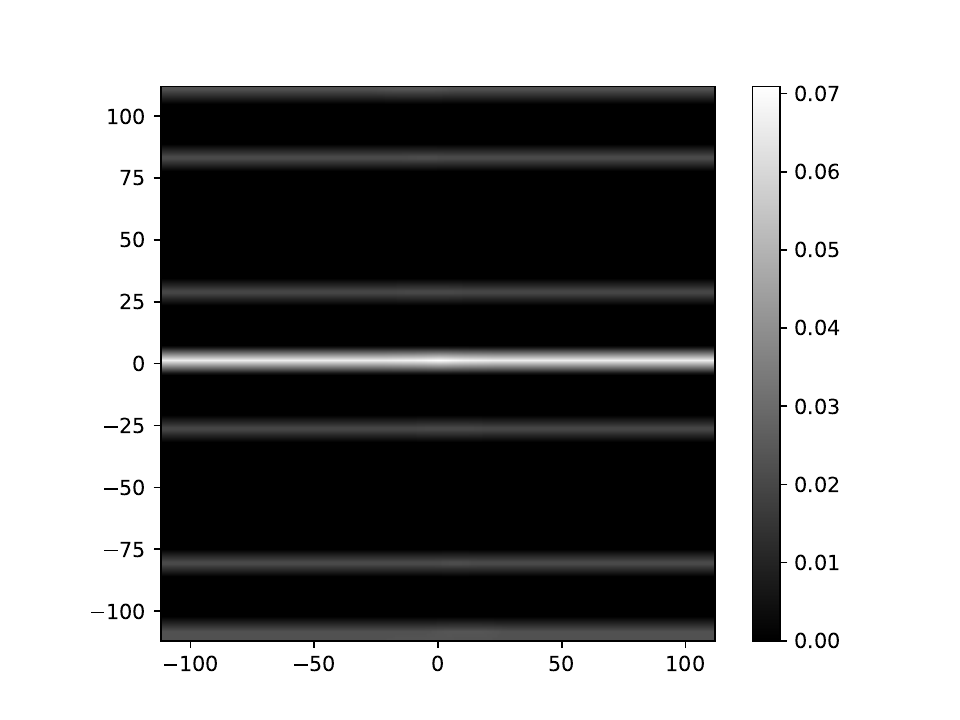} & \includegraphics[width=0.15\linewidth]{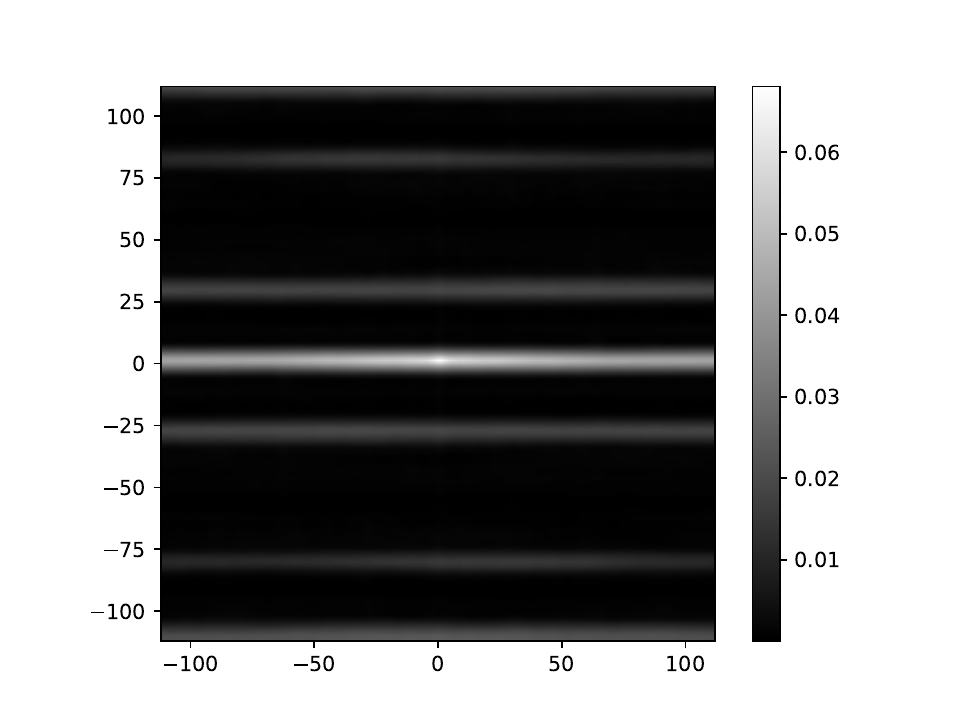} \\
\includegraphics[width=0.12\linewidth]{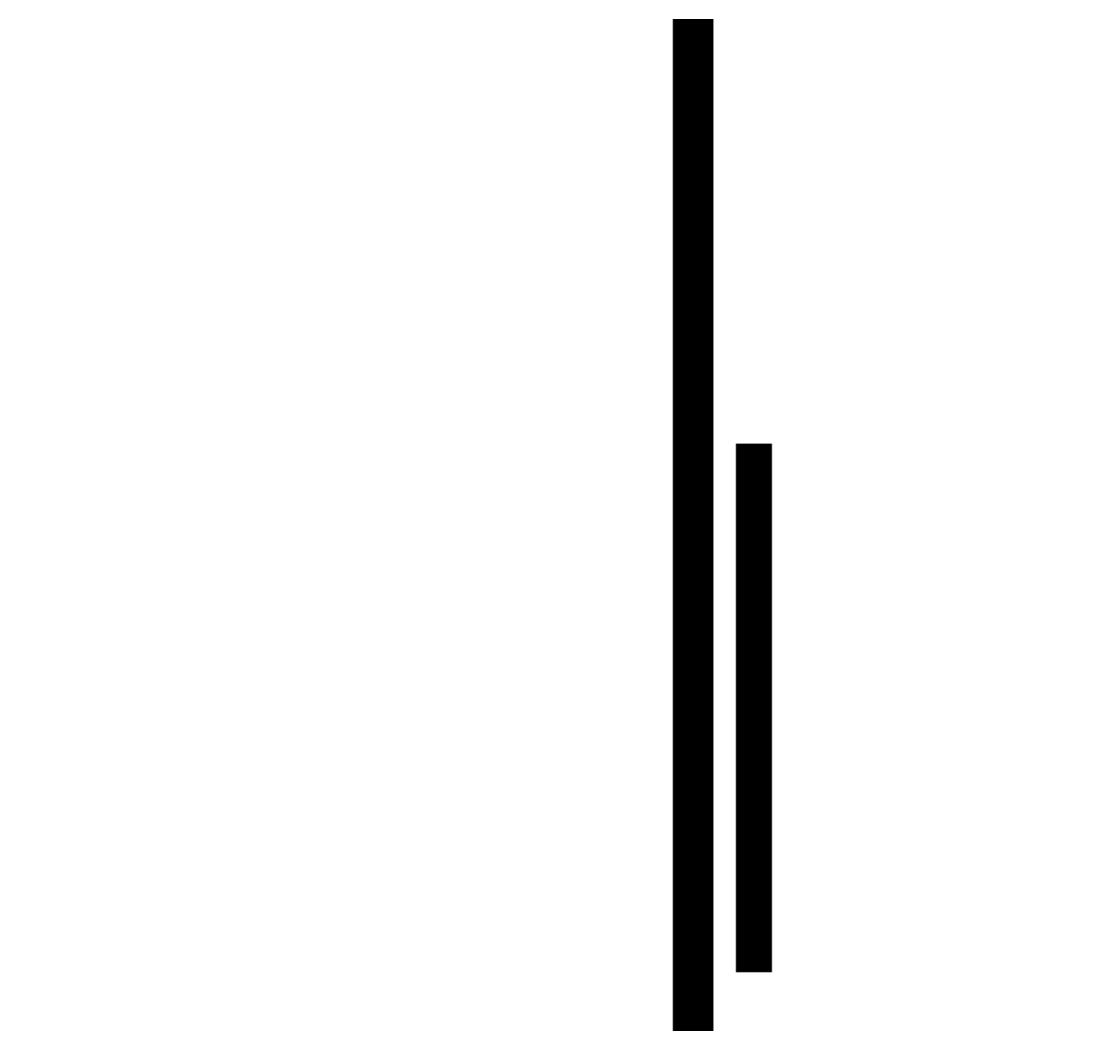} & \includegraphics[width=0.12\linewidth]{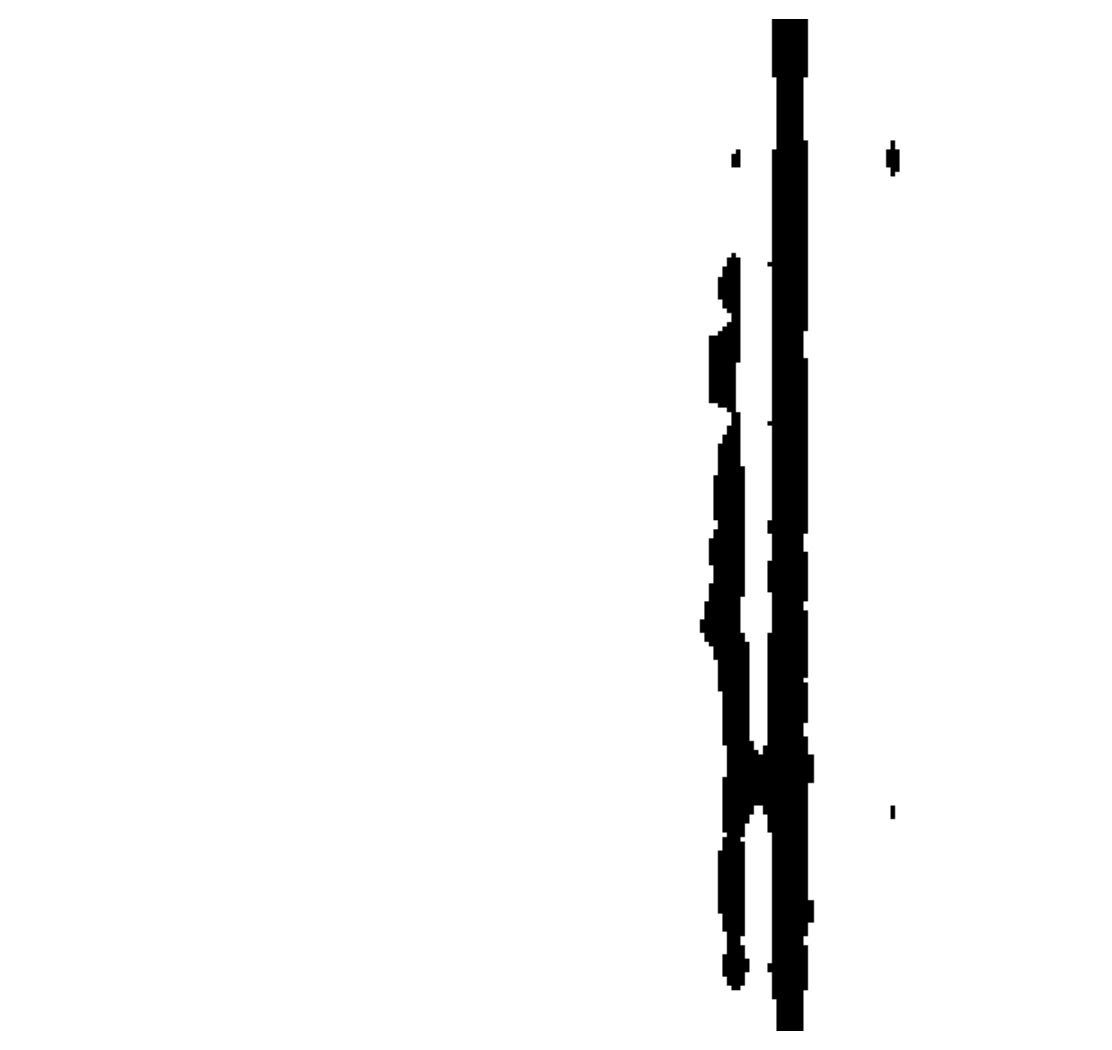} & \includegraphics[width=0.15\linewidth]{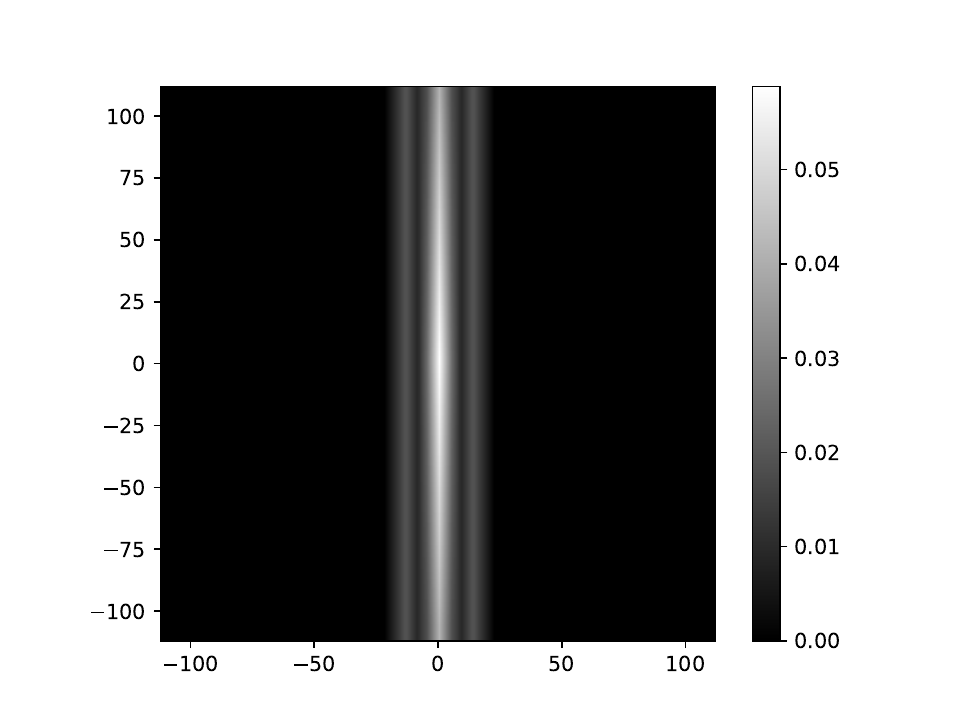} & \includegraphics[width=0.15\linewidth]{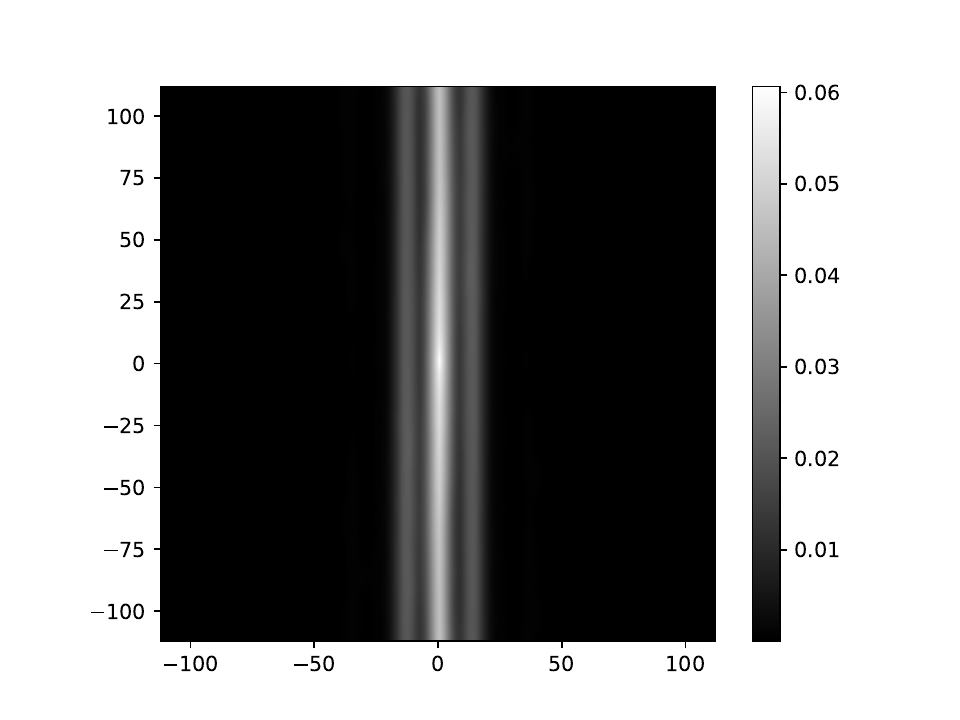} \\
\hline
\end{tabular}
\end{center}
\end{table}


\bibliography{iclr2024_conference_tinypaper}
\bibliographystyle{iclr2024_conference_tinypaper}
\newpage
\appendix

\section{Experimental Setup \label{setup}}
In our experiment, we trained a VAE on a dataset consisting of 100,000 images, each sized 224x224 pixels. The VAE utilizes a pretrained ResNet-152-based encoder with a bottleneck of size 9 and an untrained decoder with a spatial statistics loss. The training hyper-parameters were set to the following: a total of 1500 epochs, a learning rate of 0.001, a batch size of 32, and a KLD beta value of 1. 
For detailed insights into the experimental configuration, including access to the code, models, and dataset, interested readers are directed to the README.md file of the GitHub repository: \url{https://github.com/spatial-stats-vae/spatial-stats-vae.git}. This repository provides comprehensive resources for replicating this experiment.

    \begin{figure}[H]
        \begin{center}
        \includegraphics[width=5.in]{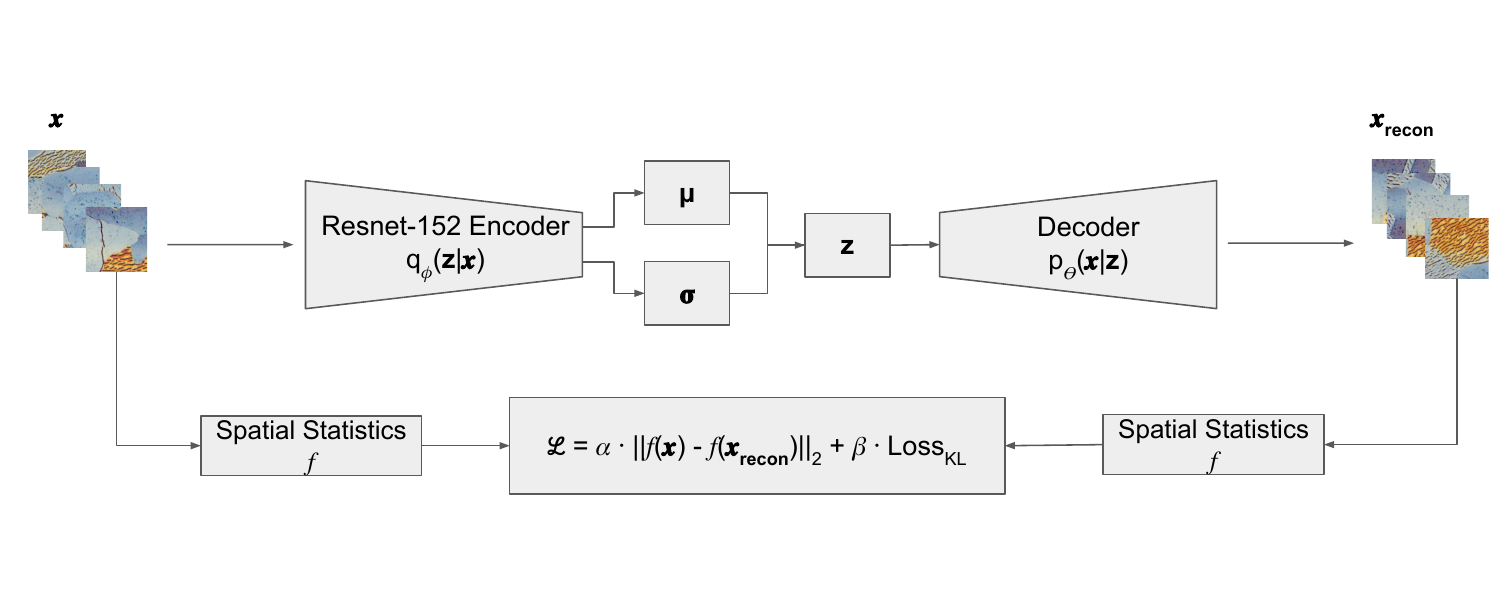}
        \end{center}
        \caption{The VAE with spatial statistics loss.}
    \end{figure}

\section{Evaluation}

We assessed the performance of the VAE with 100 image pairs from the test set. From the 100 reconstructions, 98 had the correct line orientation, and 63 had the correct number of lines as the original image

The reconstructions are characterized by randomly arranged lines, except one line remains constant across all images, varying only in size and orientation in response to the source image. The black pixels in the reconstructions, which represent lines, exhibited an average difference in volume fraction of 17\%.

Our model demonstrated a consistent ability to accurately reconstruct images that originally contained a single line. However, its performance varied with images having more lines. Notably, in instances where the original images included two or three lines, our model correctly identified the number of lines in most cases. It is also worth mentioning that when comparing the original images to those with spatial statistics most similar to the reconstructed images, there was a noticeable increase in accuracy regarding the correct identification of the number of lines (see A3). 

To determine if our model was overfitting, we analyzed 1000 reconstructed samples from both the training and validation datasets, comparing them against their respective complete sets using Mean Squared Error (MSE) as the metric. Our focus was on the distribution of average MSE values for each reconstruction, where each average denotes the mean MSE of a single reconstruction compared against all samples in the respective dataset. Our findings indicate that the distribution of the average MSEs for training reconstructions relative to the training set closely mirrors the distribution of the average MSEs for validation reconstructions relative to the validation set (refer to Figure 4). This similarity suggests that our model did not overfit to the training data. We also analyzed the MSE of spatial statistics, comparing the spatial statistics of 1000 training reconstruction samples with the spatial statistics of the entire training set, and similarly for the validation set. The findings revealed similar results for both training and validation sets (refer to Figure 5).

Lastly, we have tabulated reconstruction examples in Appendix B. In the post-processing stage of the VAE reconstructions, a threshold value of 0.05 was utilized on the reconstructed tensors. This thresholding transforms grayscale images into binary ones, a step taken to improve the clarity of the output images. Crucially, this thresholding process preserves the integrity of the reconstructions, ensuring that the resulting binary images are appropriate for further visual analysis and comparison. It should be noted that the reported MSE scores below do not depend on this thresholding function, and that the thresholding was solely done for visual purposes. For additional details, refer to the Appendix A2. 


\subsection{Calculation of volume fraction difference}
The volume fraction difference percentage is a  metric in evaluating the accuracy of the reconstructed images in terms of the black pixel content. This metric quantifies the difference in the number of black pixels (representing lines in images) between the original and the thresholded (or reconstructed) images. The calculation of this percentage is vital for understanding the degree to which the reconstructed images deviate from the original in terms of pixel composition.

The formula for calculating the volume fraction difference percentage is as follows:
\begin{equation}
    \left| \frac{\text{original\_black\_pixel\_count} - \text{thresholded\_reconstruction\_black\_pixel\_count}}{\text{original\_black\_pixel\_count}} \right| \times 100
\end{equation}

After calculating the percentage difference, the value is rounded to four decimal places for precision and ease of interpretation. This rounding process helps in standardizing the results for comparison and further analysis.

\textbf{Note:} The calculated volume fraction percentage is rounded to four decimal places.

\begin{figure}[H]
\begin{center}
\includegraphics[width=3.in]{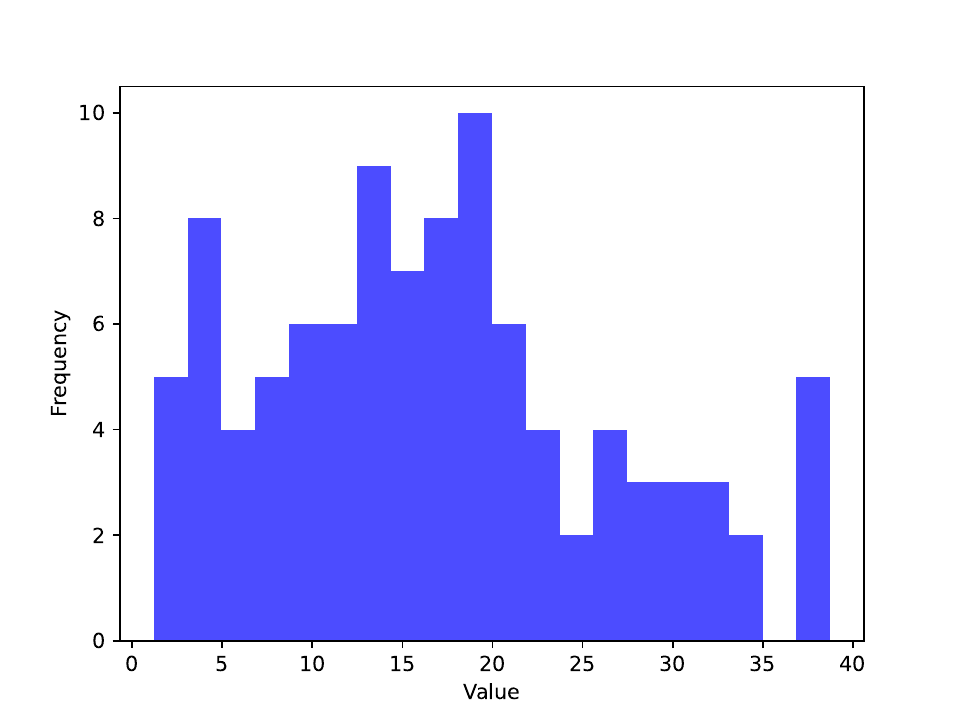}
\end{center}
\caption{The difference between percentage of black pixels in the input image and the percentage of black pixels in the reconstruction.}
\end{figure}

\subsection{Image thresholding function}
The threshold function is implemented to modify the image based on a predefined threshold value, setting pixel values below the threshold to zero and those equal to or above the threshold to one. This process is crucial for enhancing image clarity in the VAE's reconstructed outputs.

The mathematical representation of the threshold function is as follows:
\begin{equation}
    \text{threshold\_pixel}(pixel, threshold) = 
    \begin{cases} 
      0 & \text{if } pixel < threshold, \\
      1 & \text{otherwise.} 
    \end{cases}
\end{equation}

\subsection{Number of lines in original images vs. reconstructions \label{lines}}
We randomly select 100 images in the test set, compute their reconstructions, and also find the closest image in spatial statistics space to the original. We then compare the number of lines in the test image to the number of lines in the reconstruction (left) and the number of lines in the closest image in spatial statistics (right).

We show that images that are close together in spatial statistics space have similar numbers of lines, and that our reconstructions produce images with numbers of lines related to the number of lines in the original.
\begin{figure}[H]
    \begin{center}
    \includegraphics[width=5in]{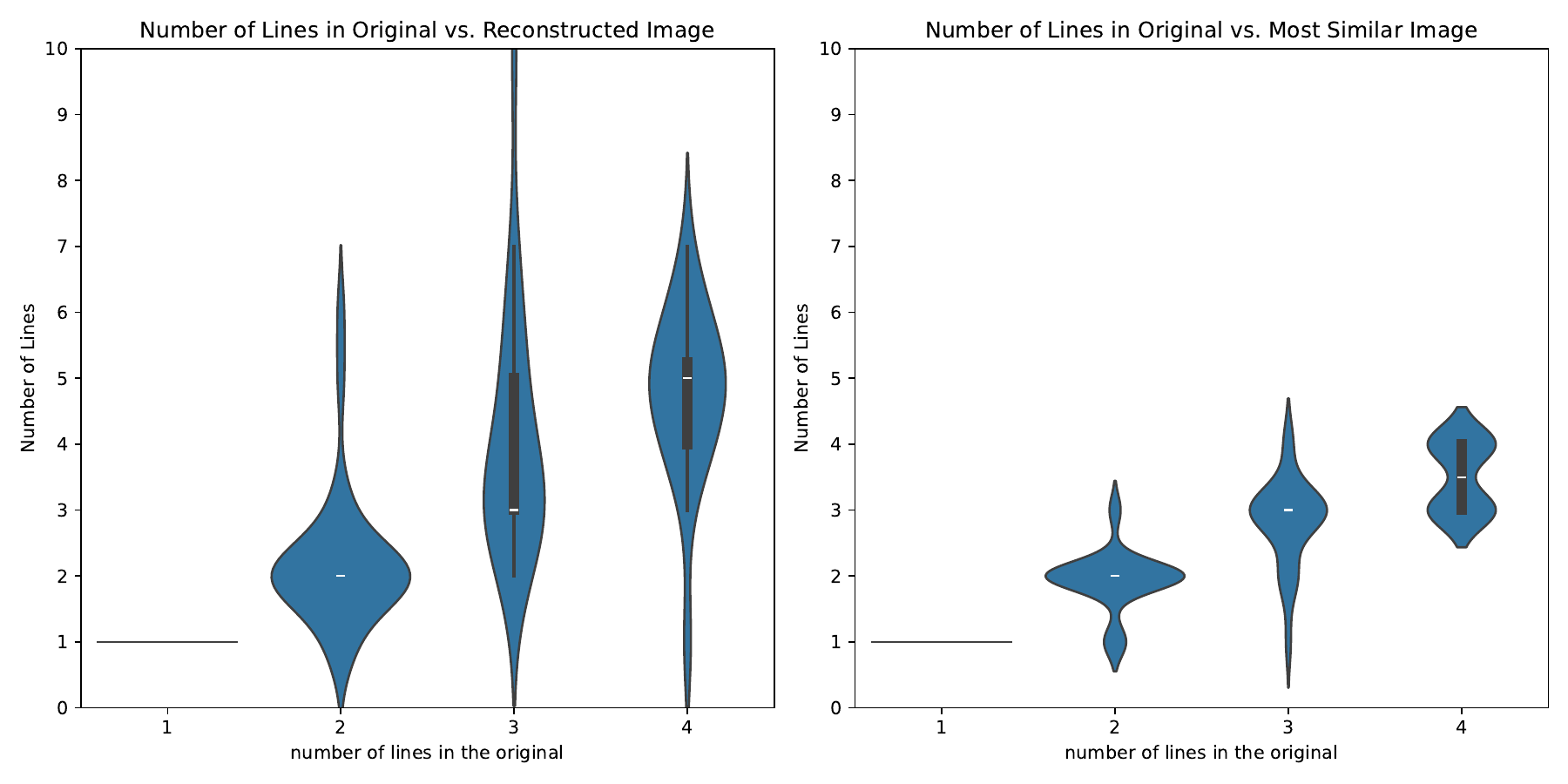}
    \end{center}
    \caption{Line count in the reconstructed image vs original image (left), line count in image closes in spatial statistics space vs line in count in original image (right)}
\end{figure}

\subsection{Quantitative results for original VAE and our VAE \label{quantresults}}
We train a ``vanilla" VAE that minimizes the distance in data space between the original and the reconstruction (``data space loss"), and our proposed model, which minimizes the distance in spatial statistics space between the original and the reconstruction (``spatial statistics loss"). Our method produces reconstructions that are closer together in spatial statistics space, but further apart in data space.

This is to be expected: we directly try to make the reconstructions be as close as possible as the original in spatial statistics space, and do not try to match the original in data space at all.

Note the exact reconstructions that are close to the original in data space would also be close to the original in spatial statistics space. For that reason, the spatial statistics error when training with data space loss is larger, but not much larger, than when training with data space loss.

\begin{figure}[H]
\begin{center}
\includegraphics[width=5.5in]{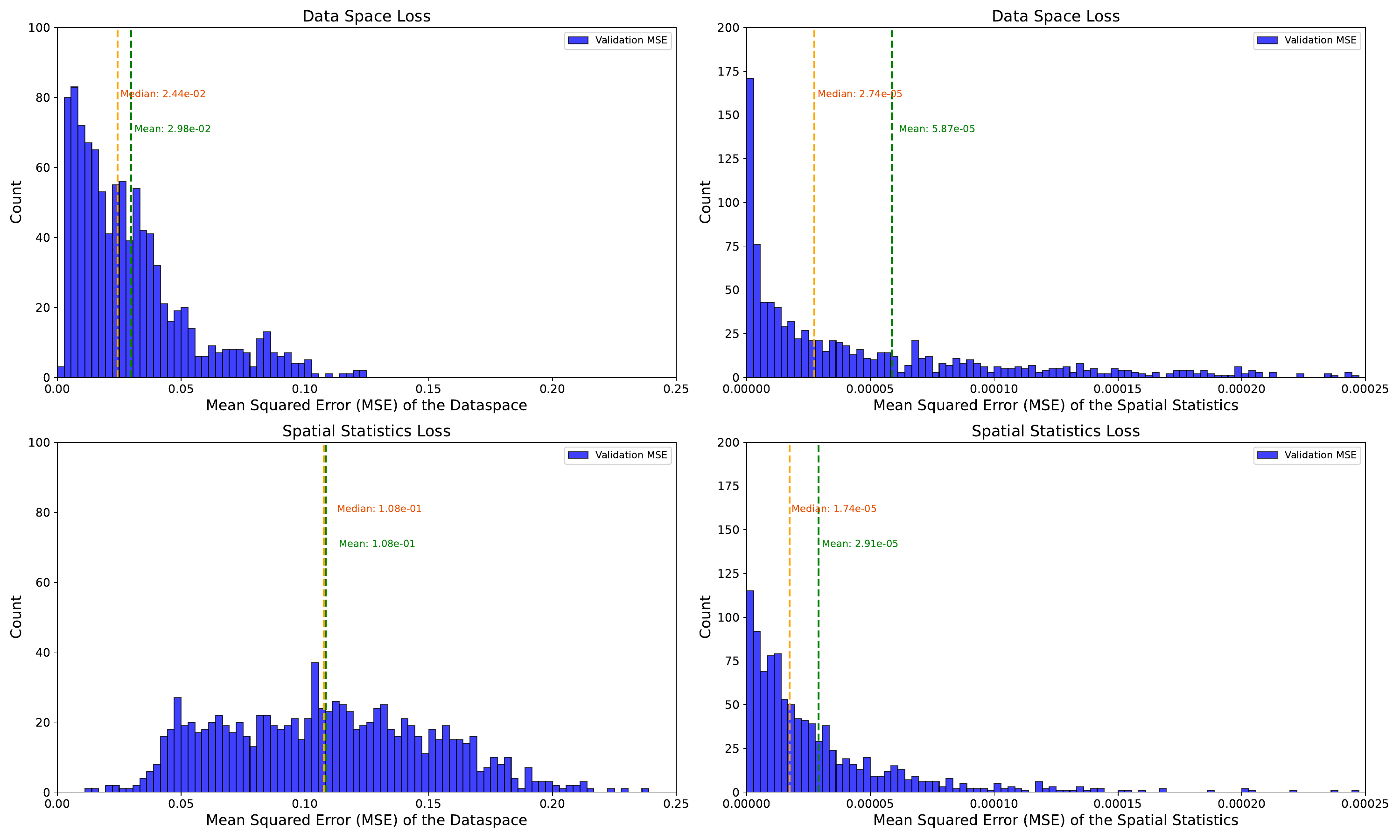}
\end{center}
\end{figure}




\section{More illustrative examples}
\subsection{Reconstructions and their spatial statistics}

The following table contains illustrative instances of reconstructions from the 100 samples. The table shows the original image from the dataset, and the reconstructed image by our model in the two columns on the left. The two columns on the right show the auto-correlations (aka spatial statistics) of the original and the reconstructed images, respectively. We have also included the 2 reconstruction samples (out of 100) that did have the correct line orientation at the end of the table.
\begin{table}[H]
\caption{Original, reconstructed, and spatial statistics difference}
\begin{center}
\begin{tabular}{cccc}
\hline
\multicolumn{2}{c}{\bf Data} & \multicolumn{2}{c}{\bf Auto-correlations} \\
\hline
{\bf Original} & {\bf Reconstructed} & {\bf Original} & {\bf Reconstructed} \\
\hline
\includegraphics[width=0.2\linewidth]{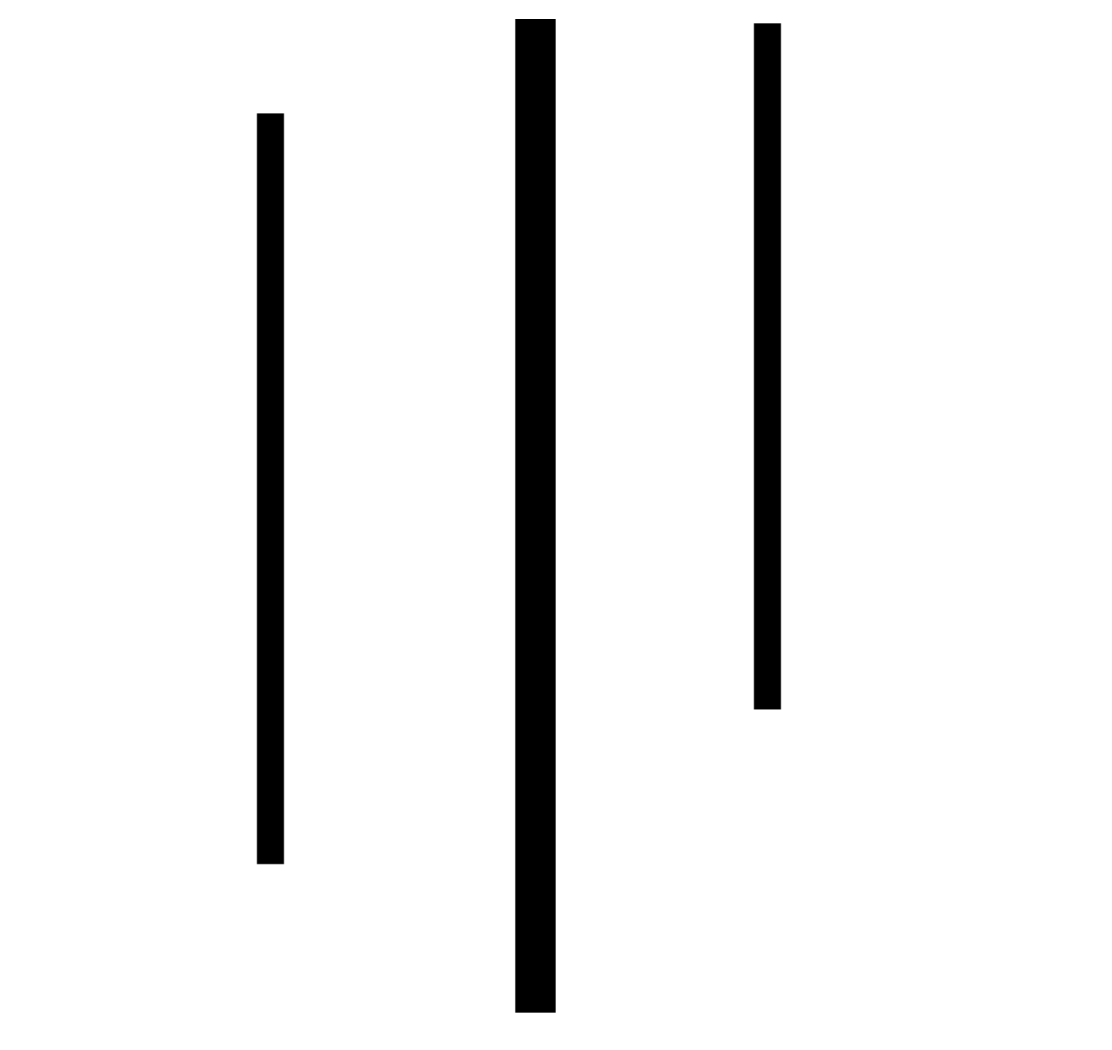} & \includegraphics[width=0.2\linewidth]{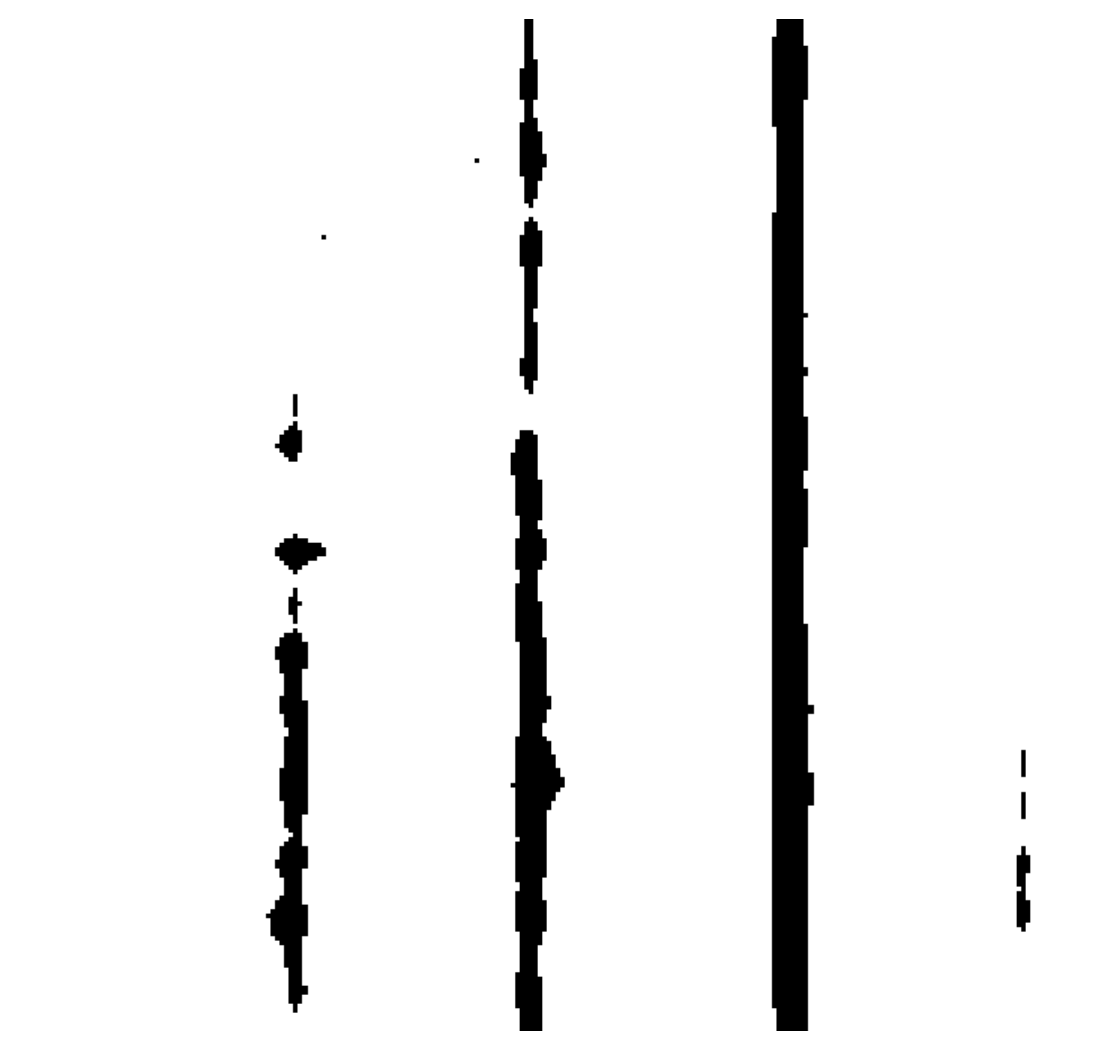} & \includegraphics[width=0.2\linewidth]{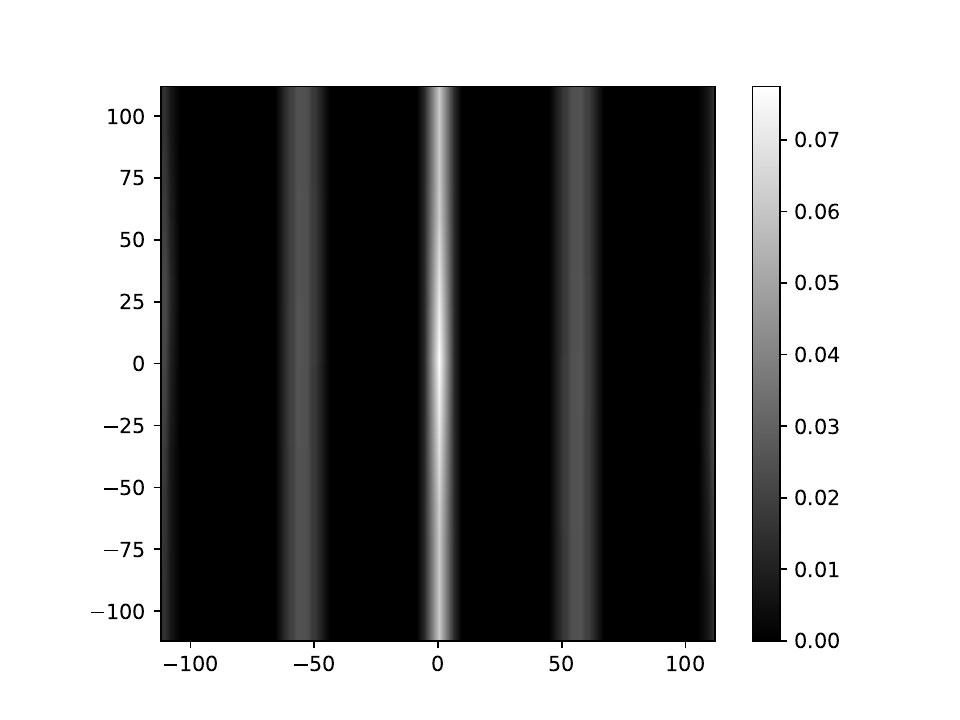} & \includegraphics[width=0.2\linewidth]{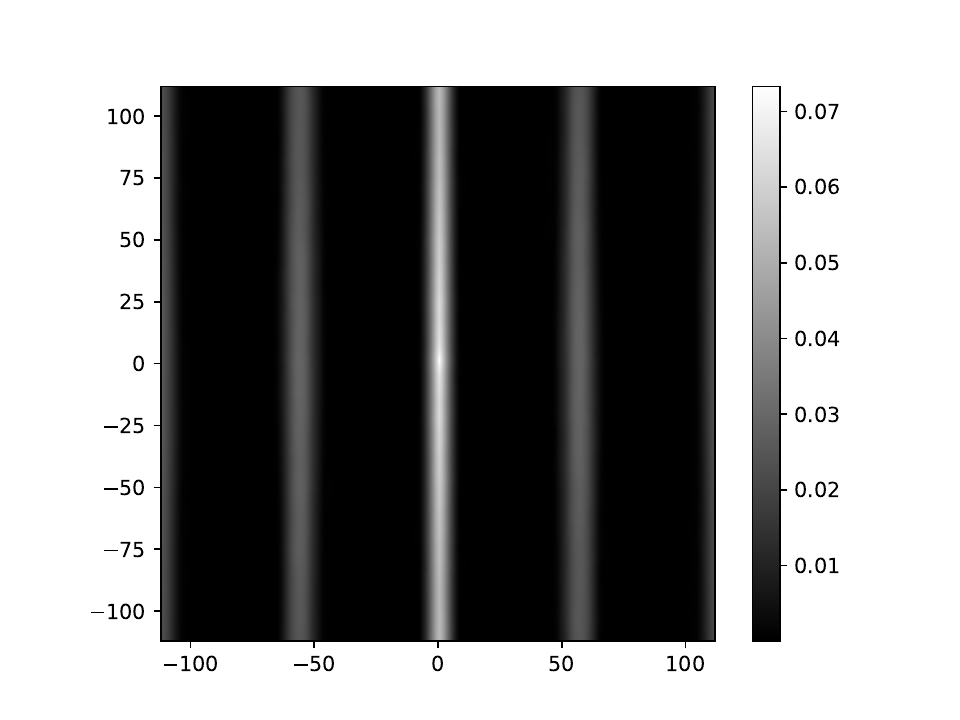} \\
\includegraphics[width=0.2\linewidth]{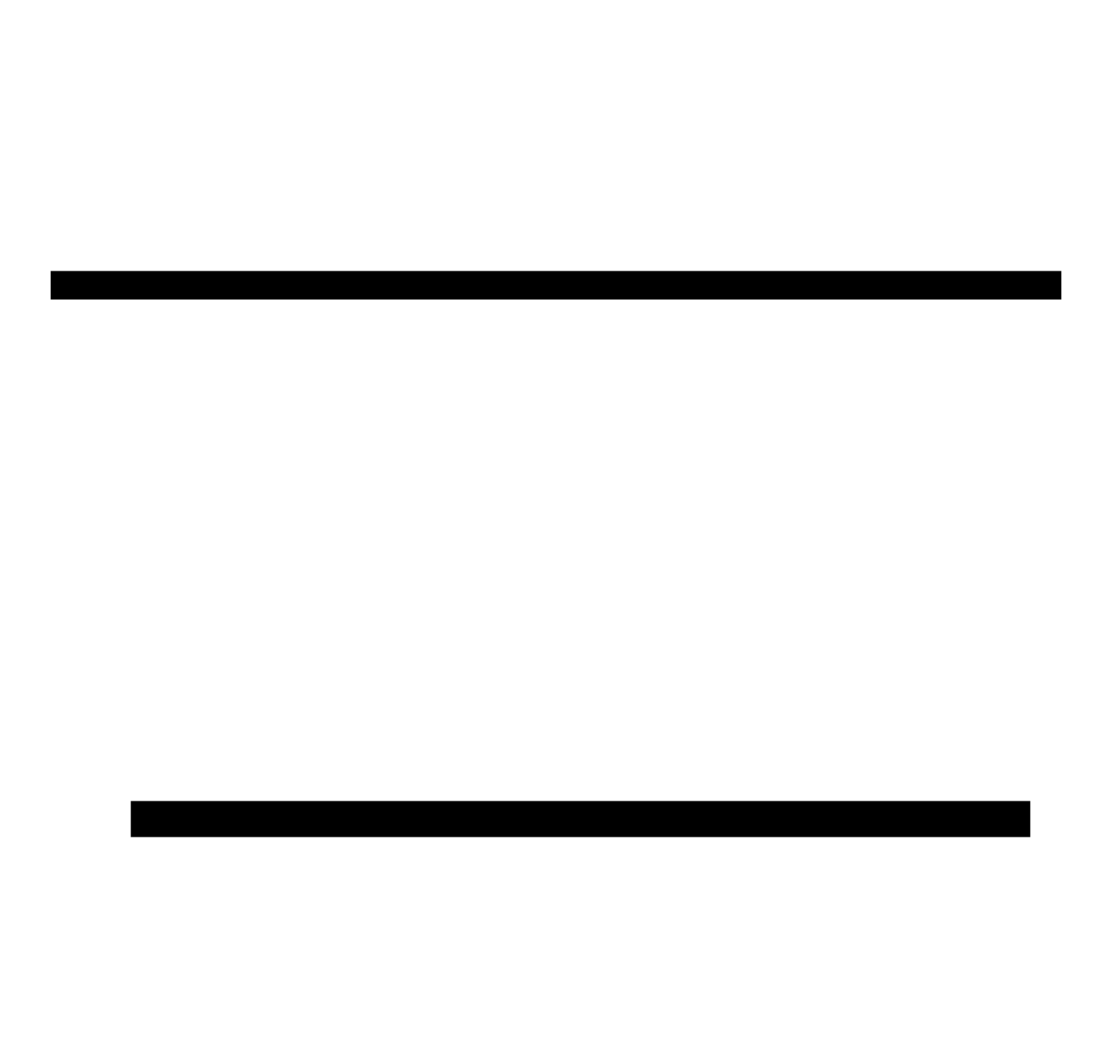} & \includegraphics[width=0.2\linewidth]{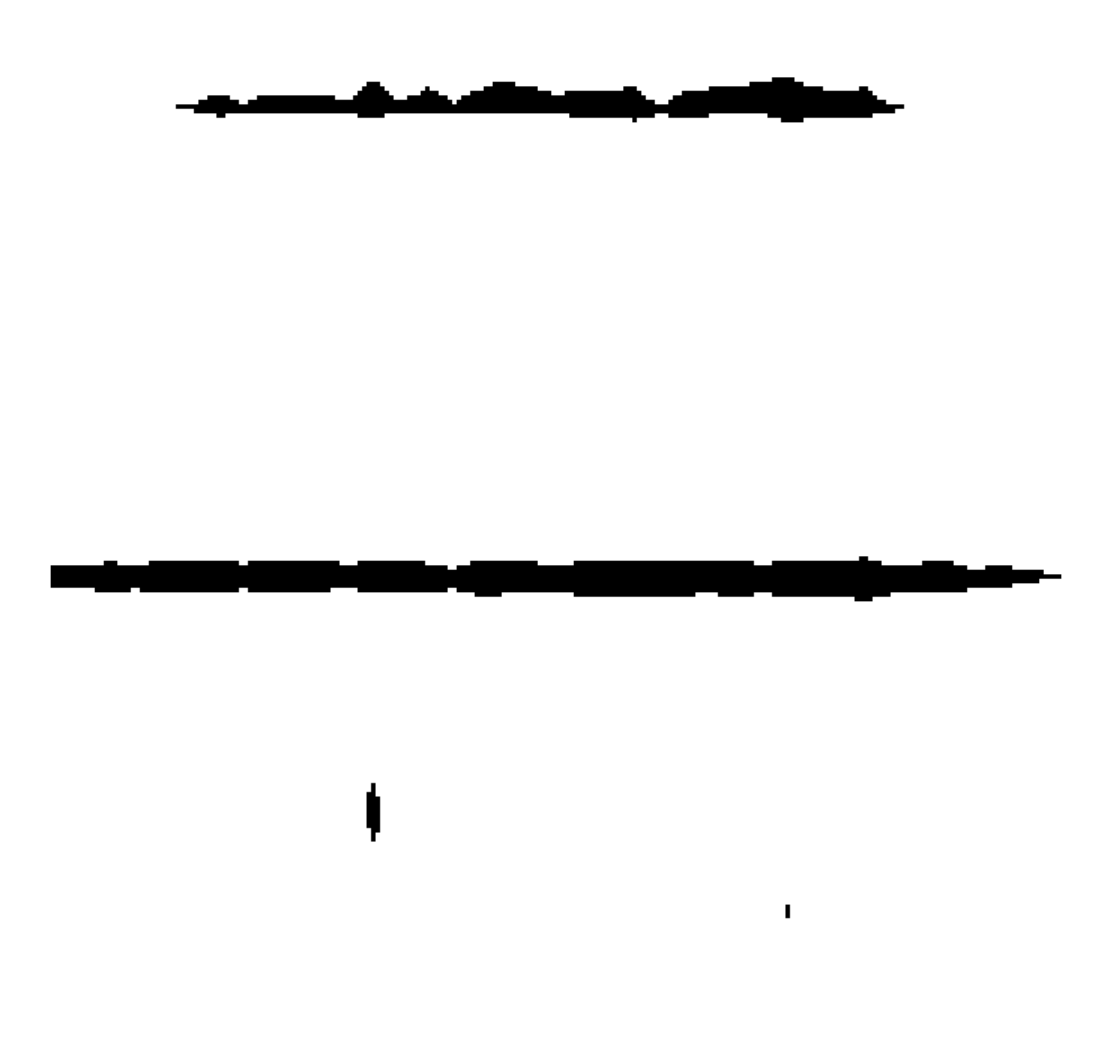} & \includegraphics[width=0.2\linewidth]{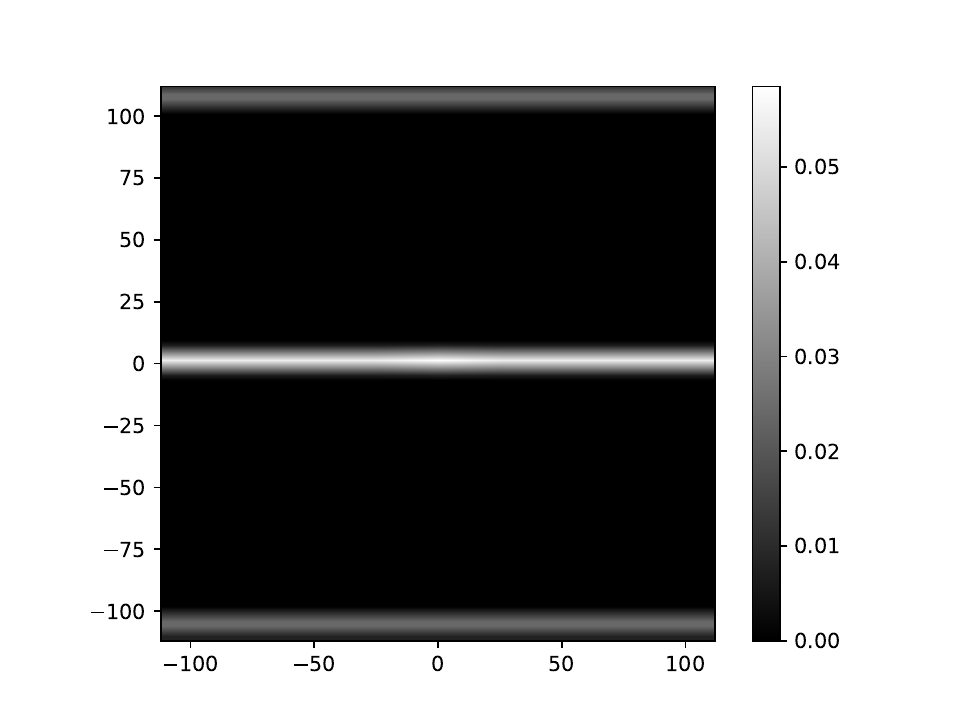} & \includegraphics[width=0.2\linewidth]{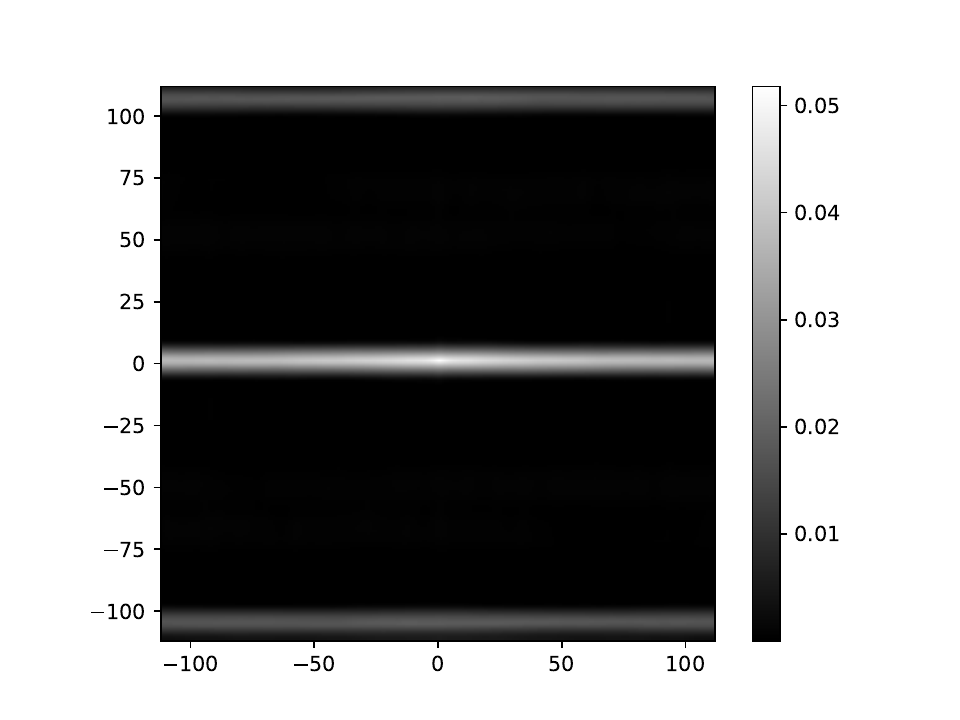} \\
\includegraphics[width=0.2\linewidth]{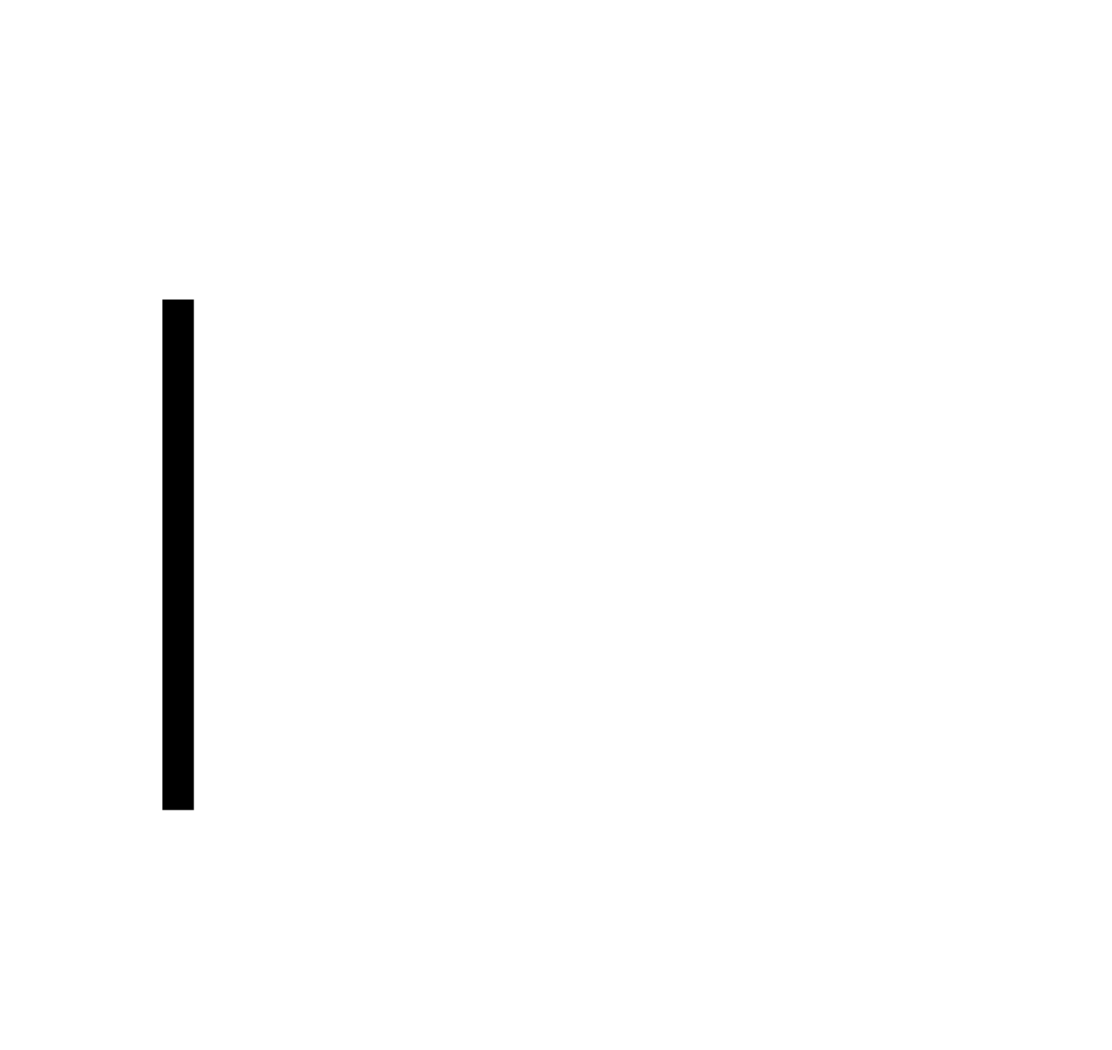} & \includegraphics[width=0.2\linewidth]{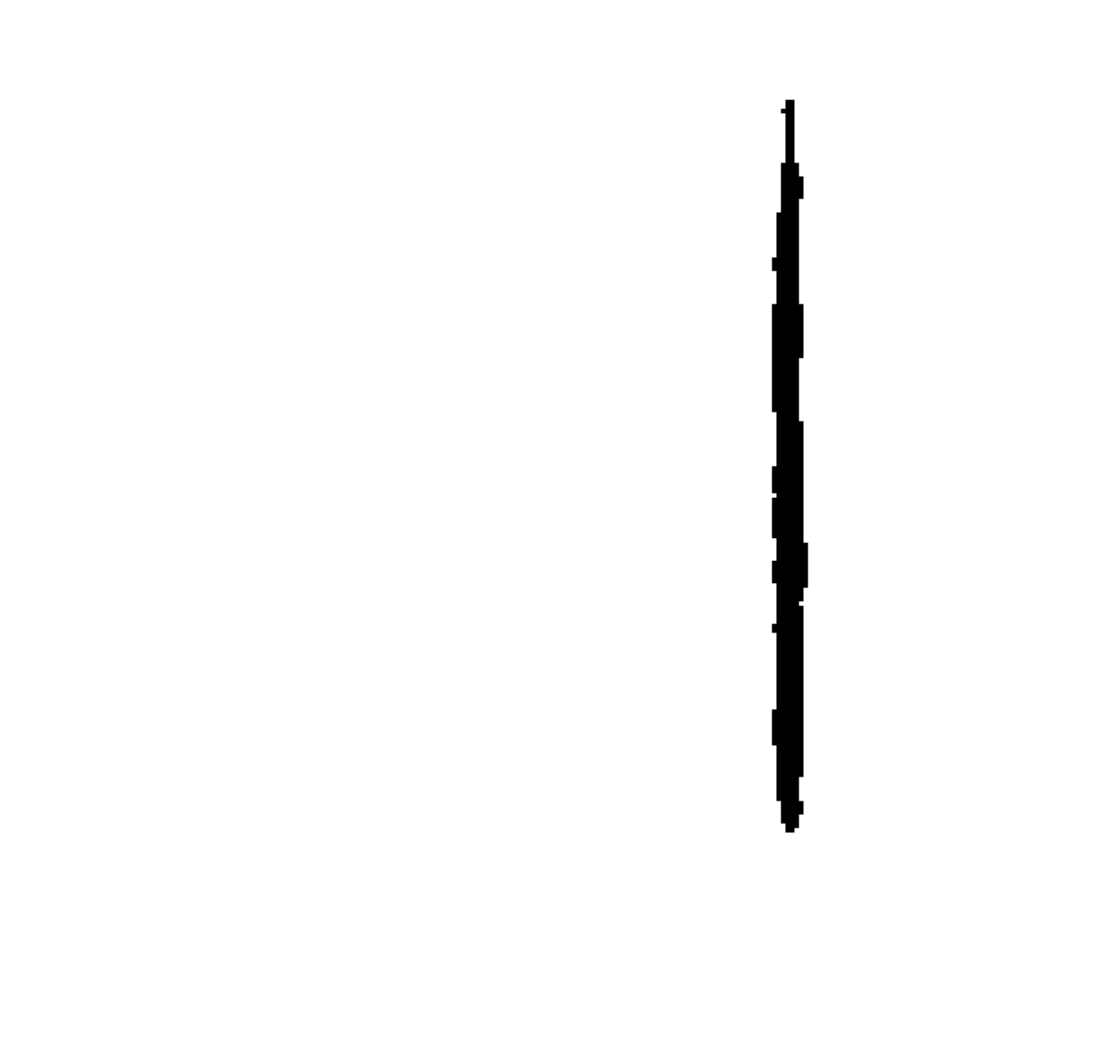} & \includegraphics[width=0.2\linewidth]{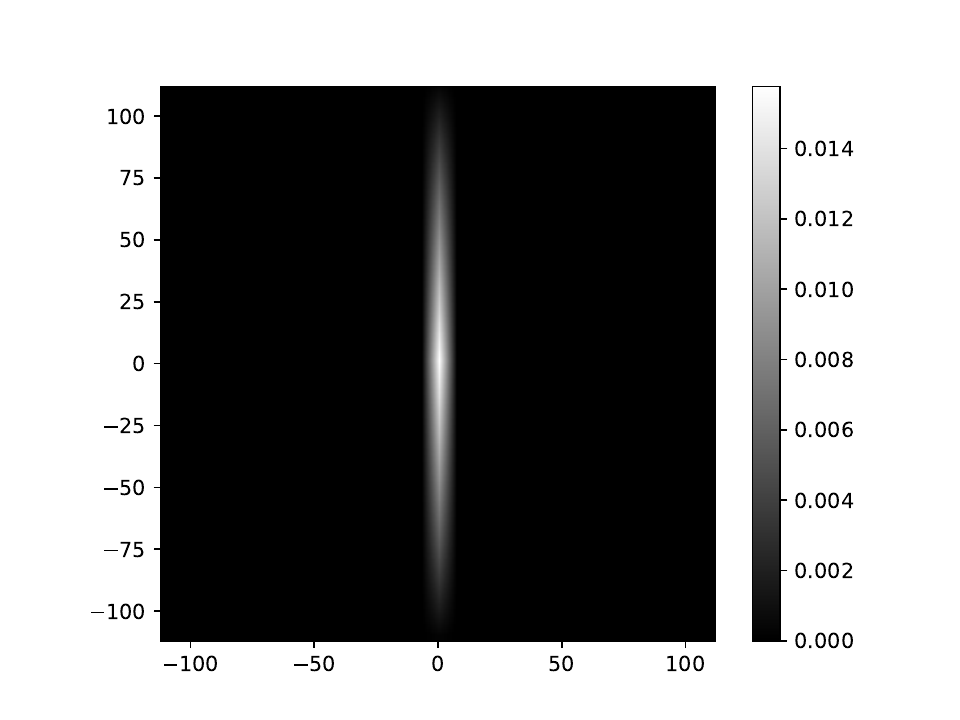} & \includegraphics[width=0.2\linewidth]{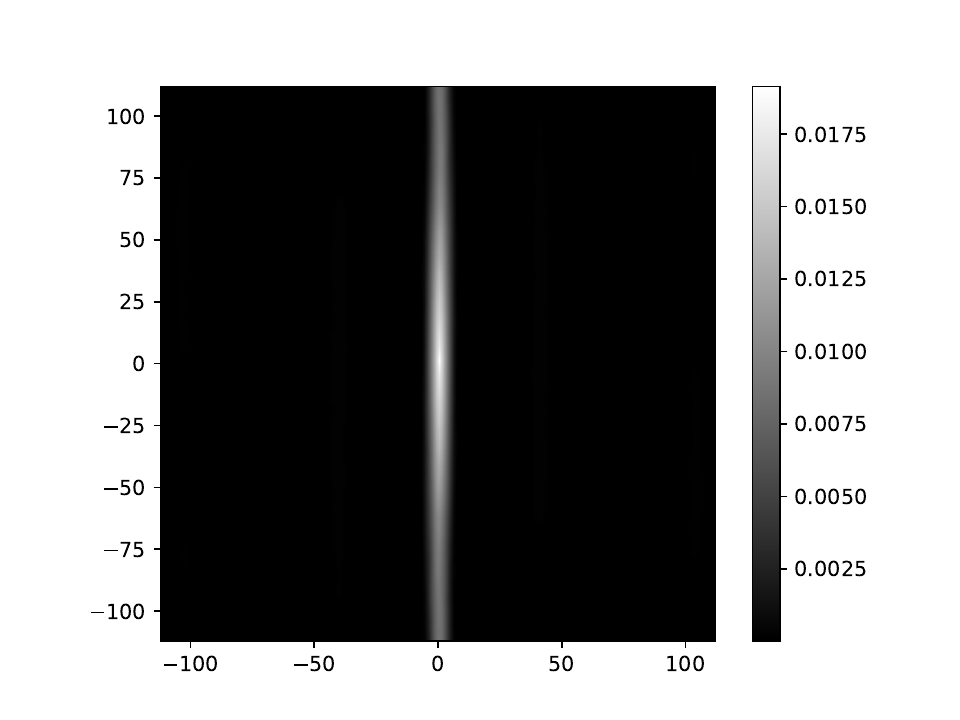} \\
\includegraphics[width=0.2\linewidth]{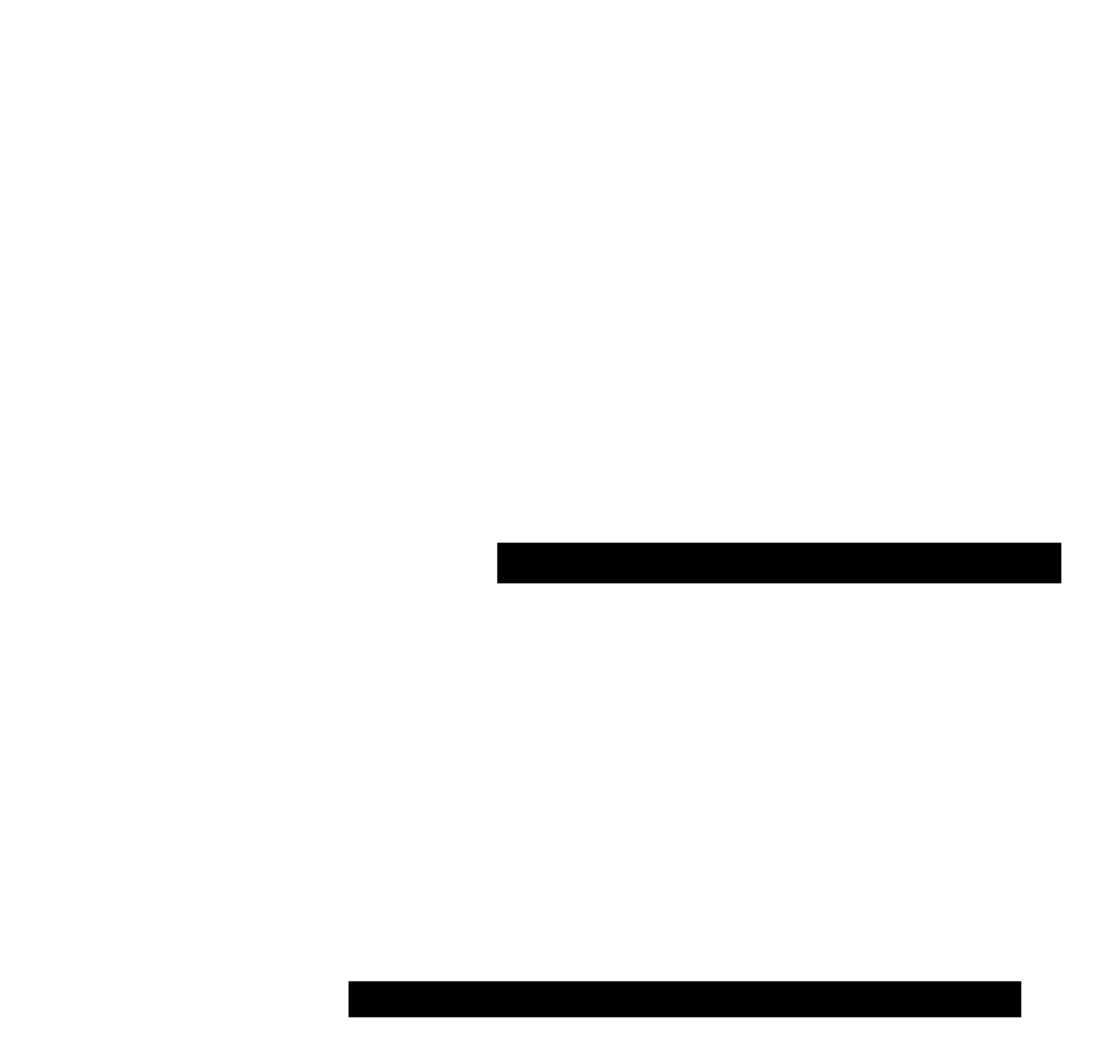} & \includegraphics[width=0.2\linewidth]{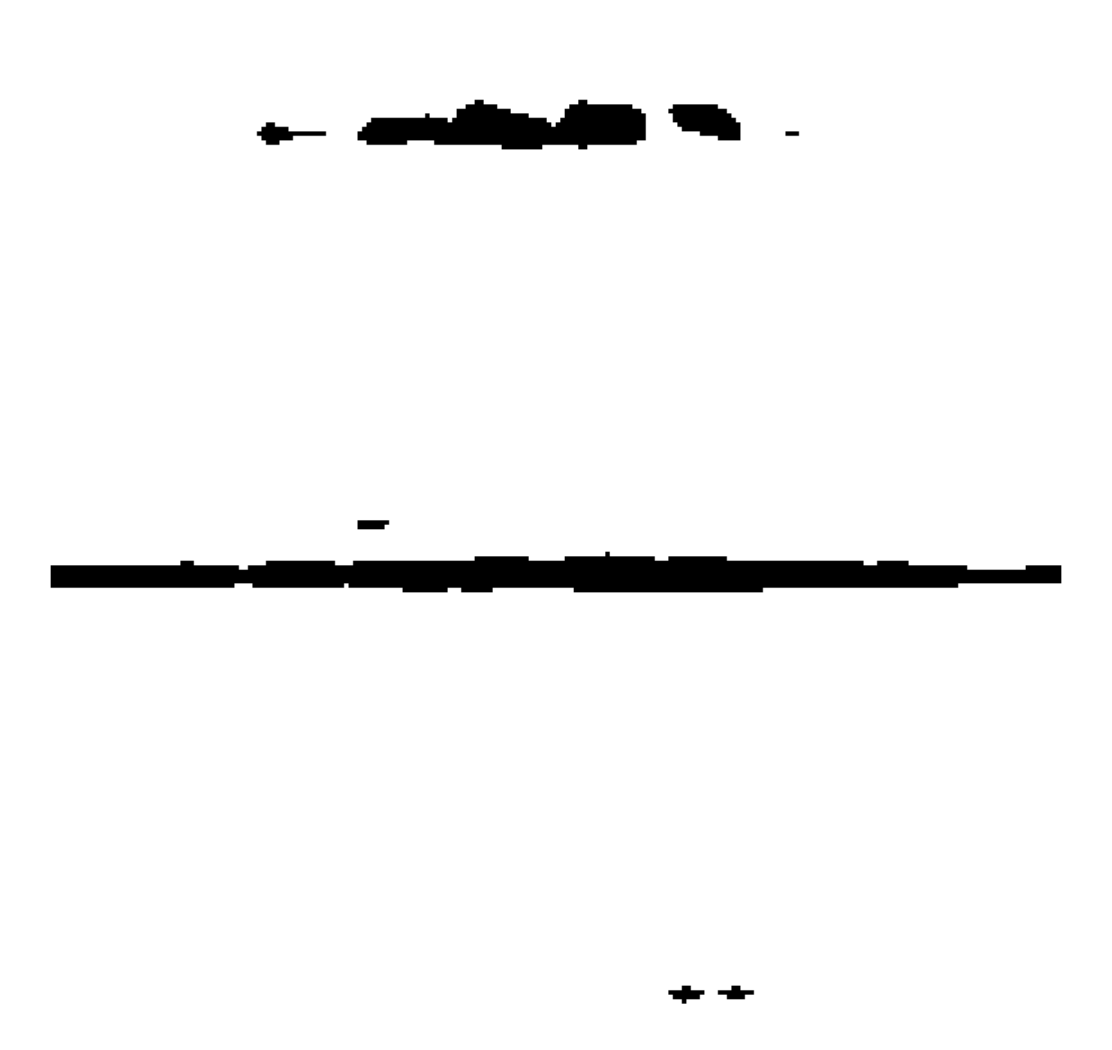} & \includegraphics[width=0.2\linewidth]{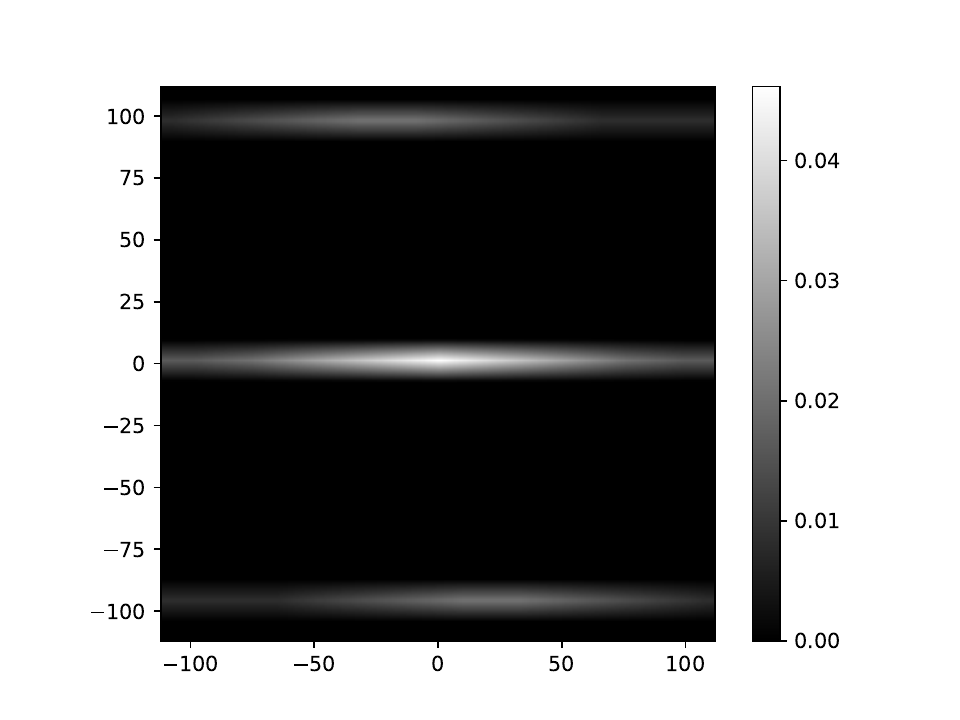} & \includegraphics[width=0.2\linewidth]{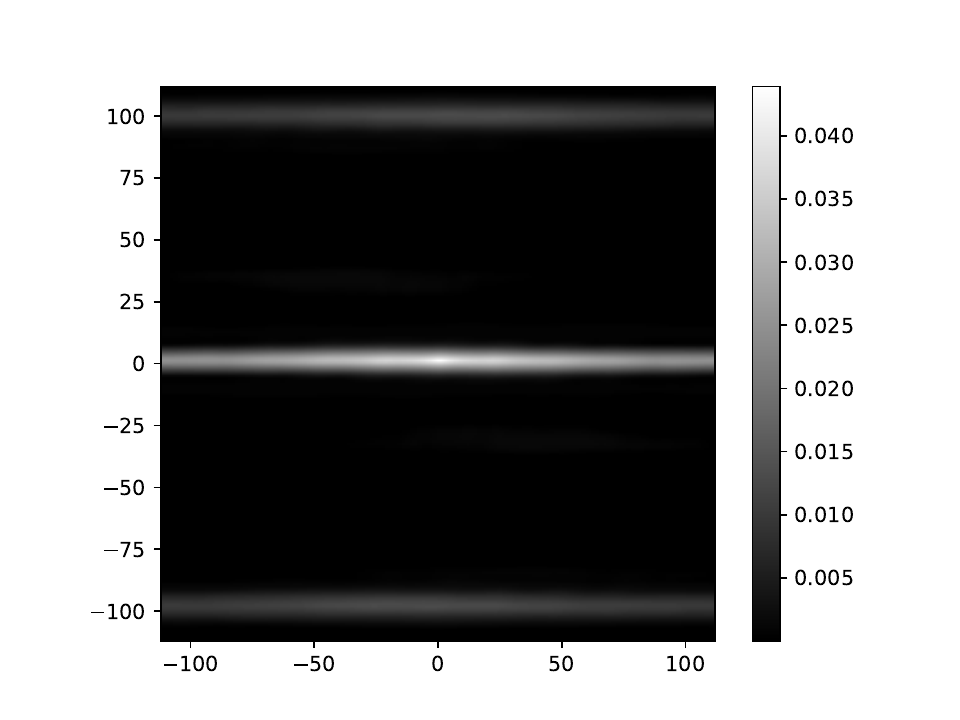} \\
\includegraphics[width=0.2\linewidth]{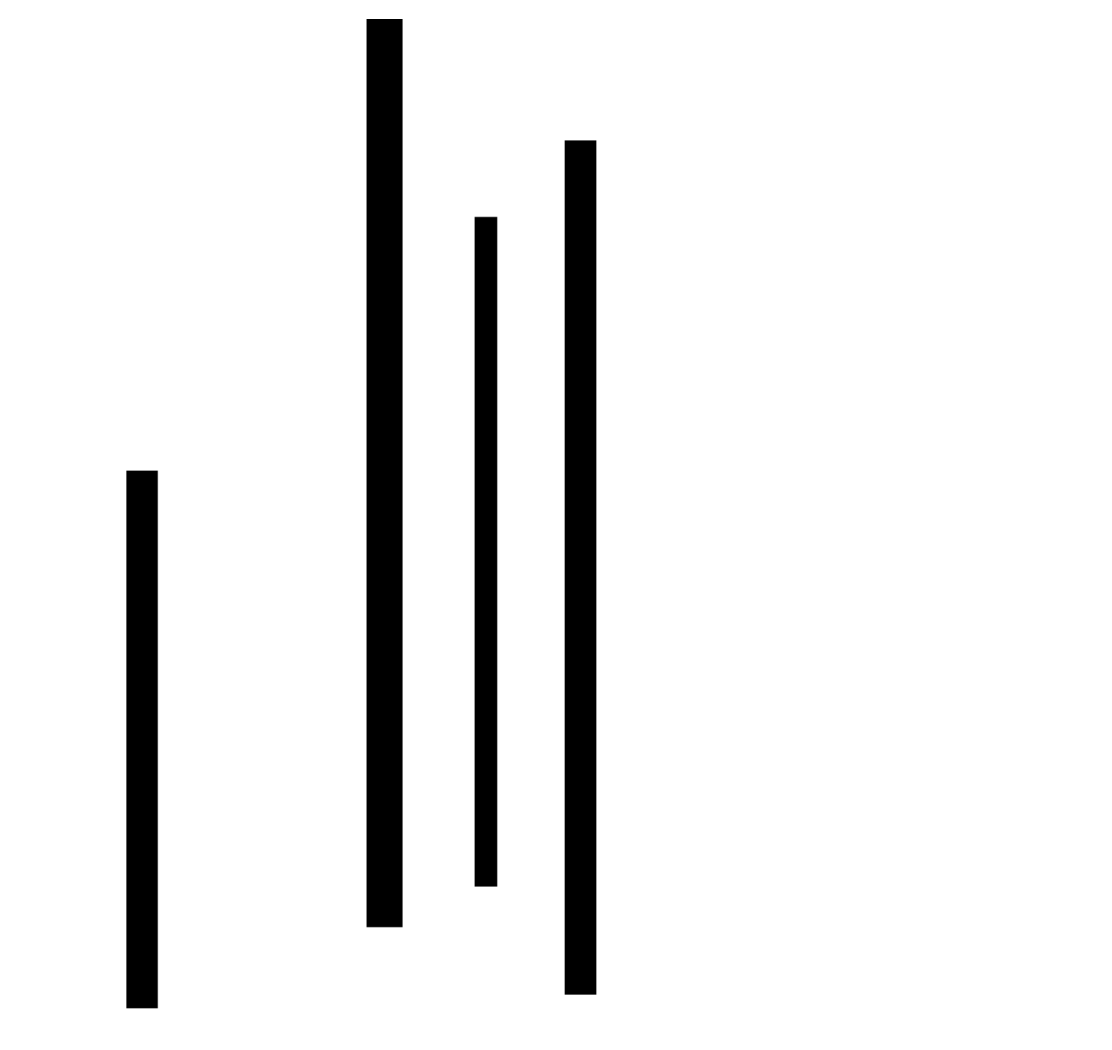} & \includegraphics[width=0.2\linewidth]{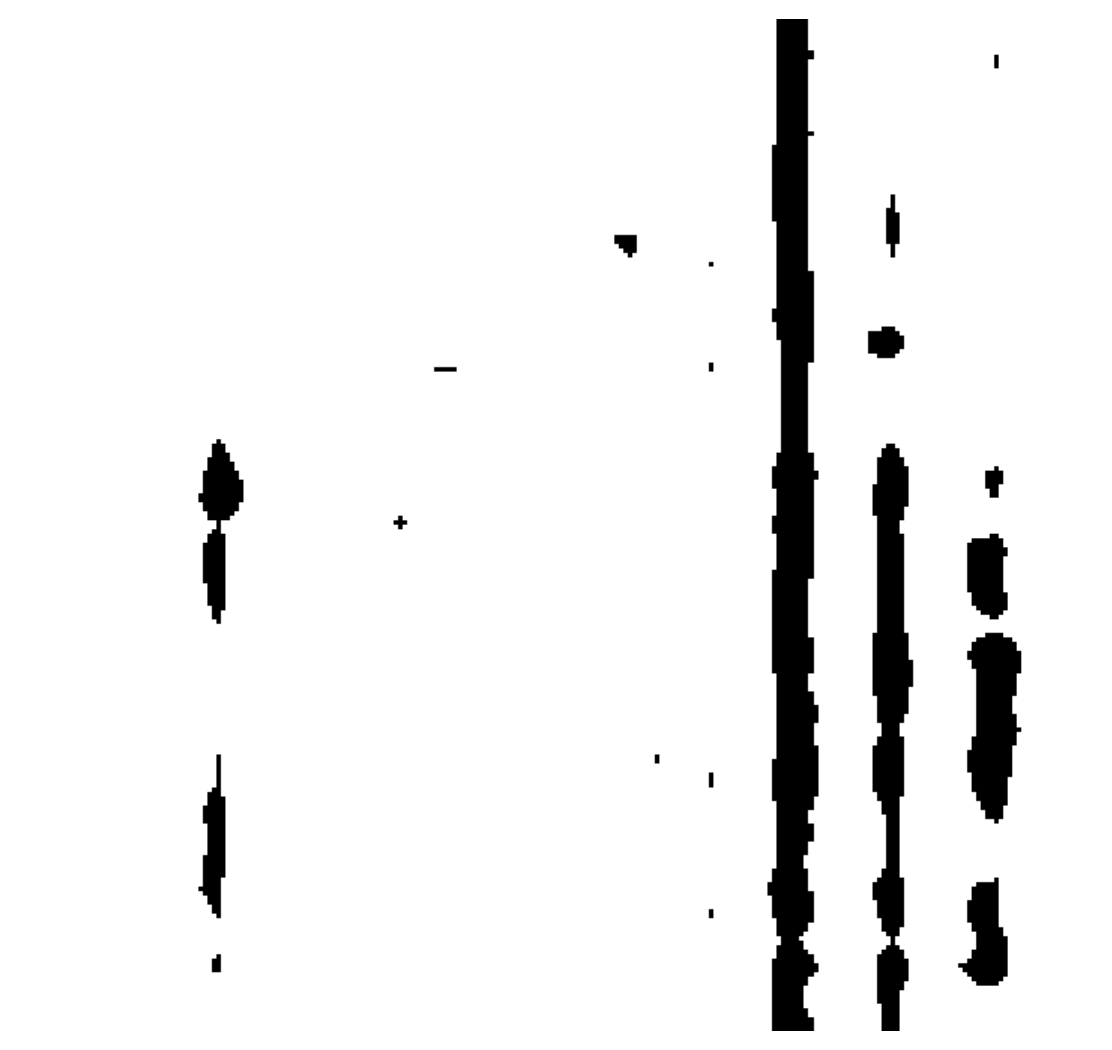} & \includegraphics[width=0.2\linewidth]{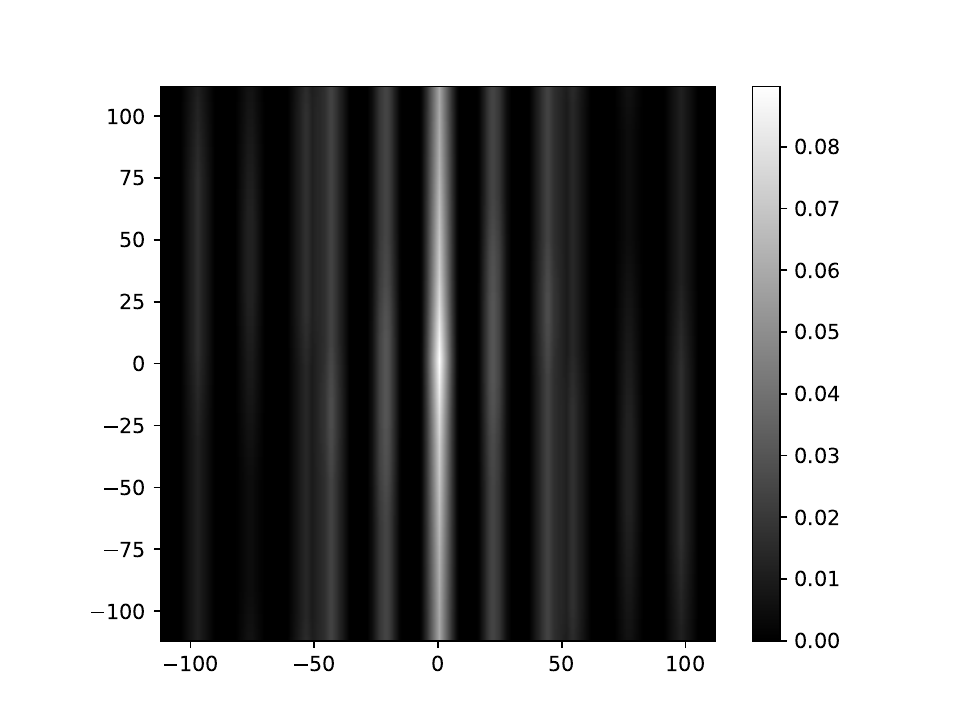} & \includegraphics[width=0.2\linewidth]{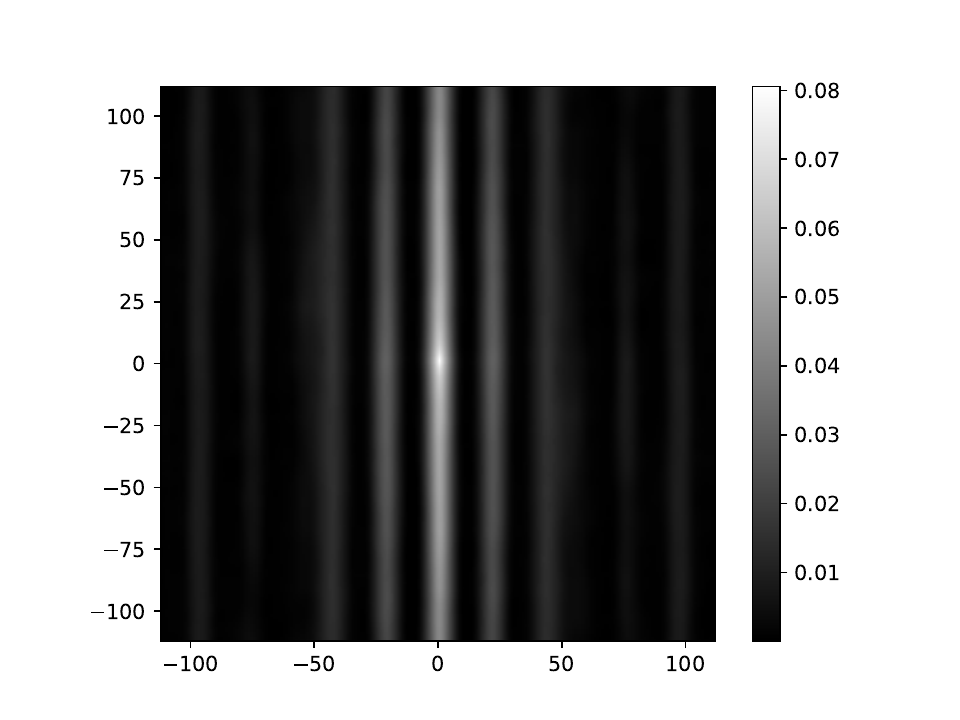} \\
\includegraphics[width=0.2\linewidth]{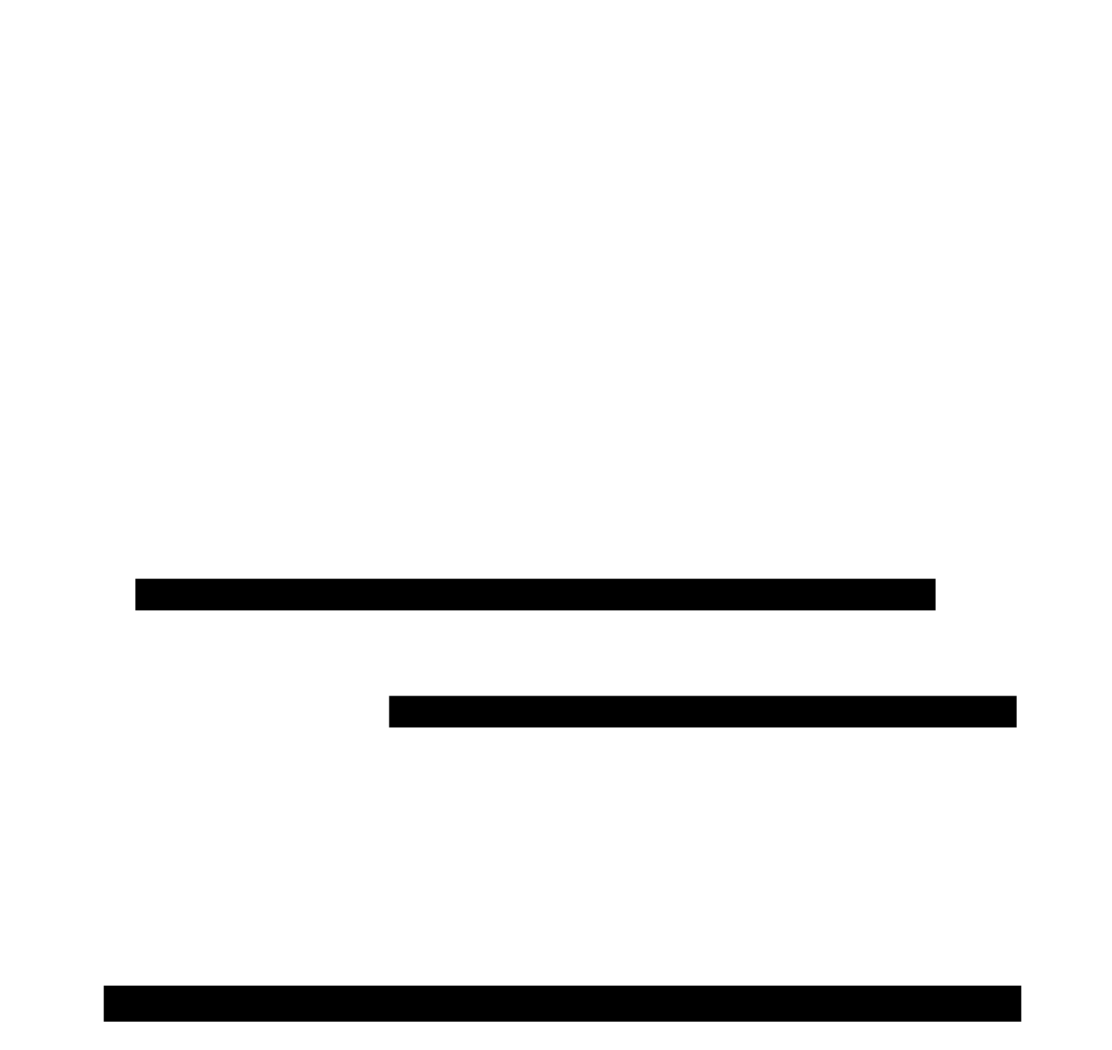} & \includegraphics[width=0.2\linewidth]{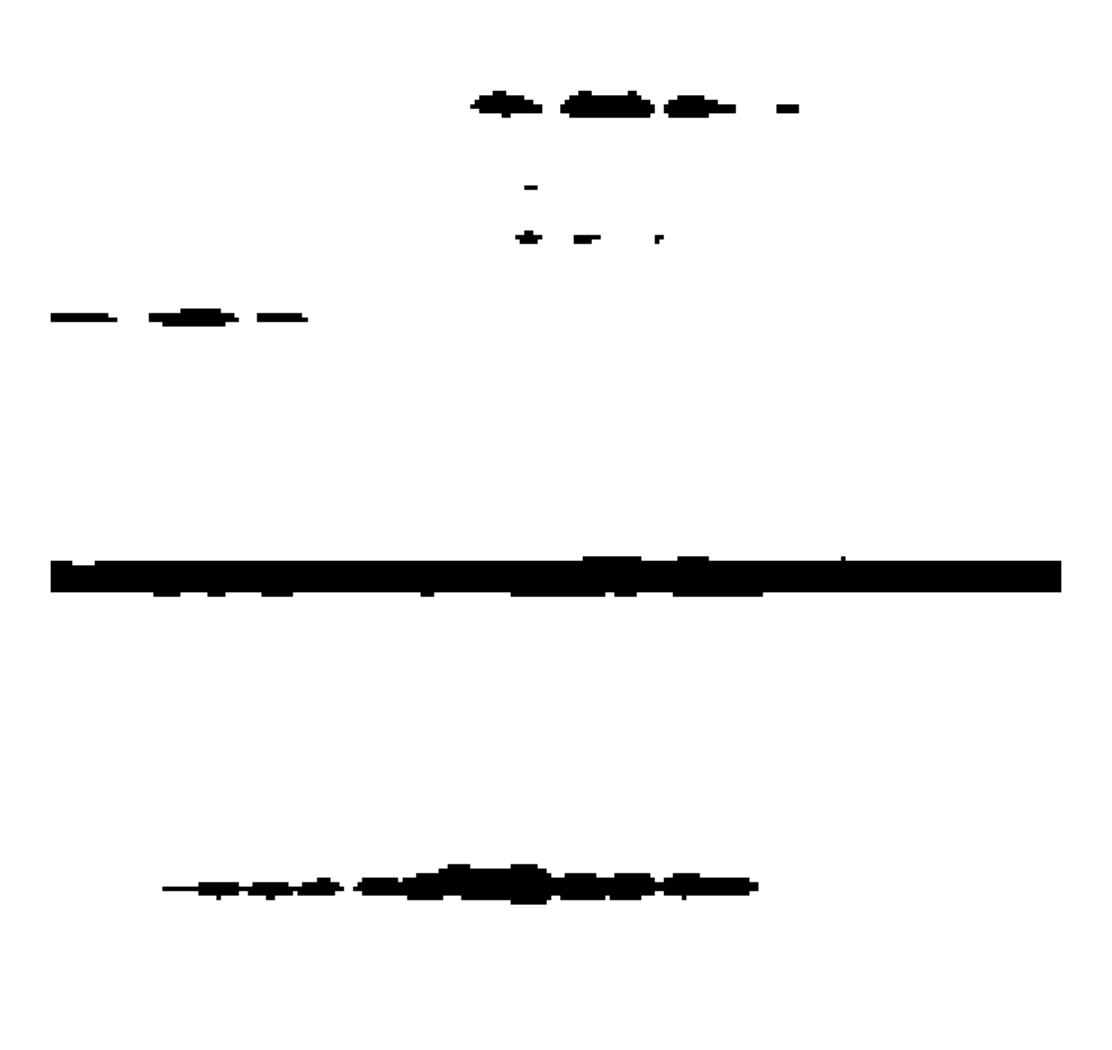} & \includegraphics[width=0.2\linewidth]{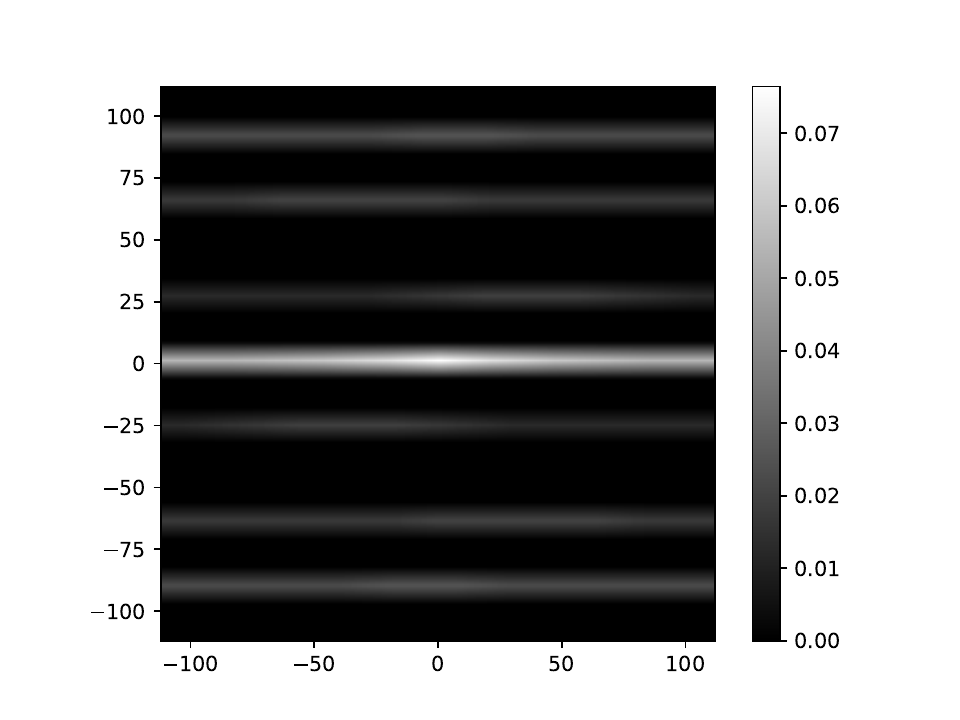} & \includegraphics[width=0.2\linewidth]{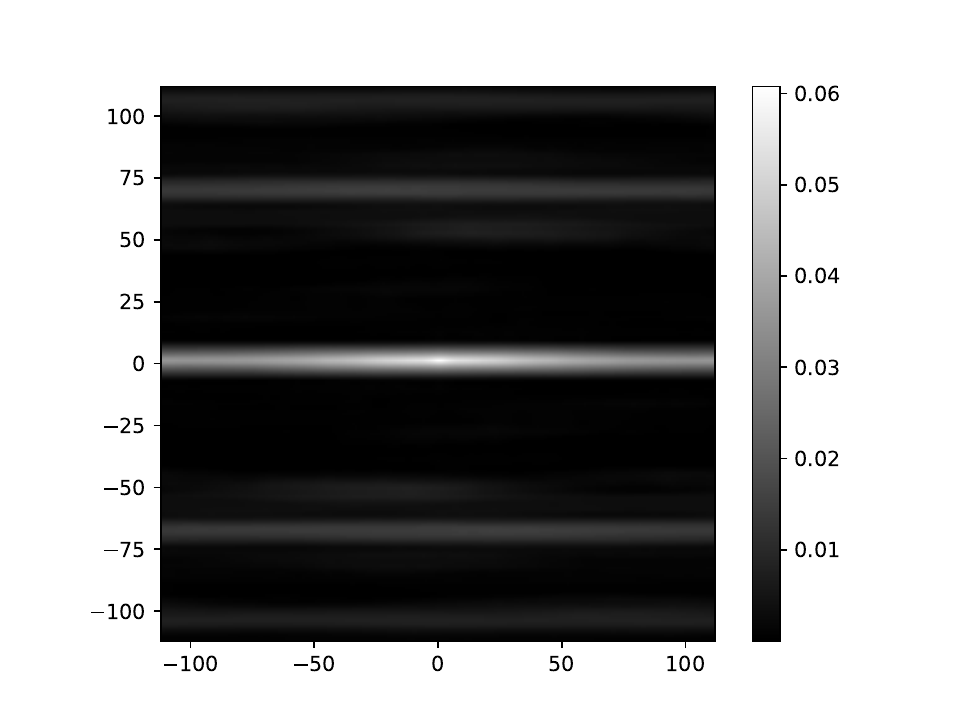} \\
\hline
\end{tabular}
\end{center}
\end{table}

\begin{table}[H]
\caption{Original, reconstructed, and spatial statistics difference}
\begin{center}
\begin{tabular}{cccc}
\hline
\multicolumn{2}{c}{\bf Data} & \multicolumn{2}{c}{\bf Auto-correlations} \\
\hline
{\bf Original} & {\bf Reconstructed} & {\bf Original} & {\bf Reconstructed} \\
\hline
\includegraphics[width=0.2\linewidth]{original_image_43.pdf} & \includegraphics[width=0.2\linewidth]{reconstructed_image_43.pdf} & \includegraphics[width=0.2\linewidth]{reconstructed_image_autocorrelation_43.pdf} & \includegraphics[width=0.2\linewidth]{original_image_autocorrelation_43.pdf} \\
\includegraphics[width=0.2\linewidth]{original_image_36.pdf} & \includegraphics[width=0.2\linewidth]{reconstructed_image_36.pdf} & \includegraphics[width=0.2\linewidth]{reconstructed_image_autocorrelation_36.pdf} & \includegraphics[width=0.2\linewidth]{original_image_autocorrelation_36.pdf} \\
\includegraphics[width=0.2\linewidth]{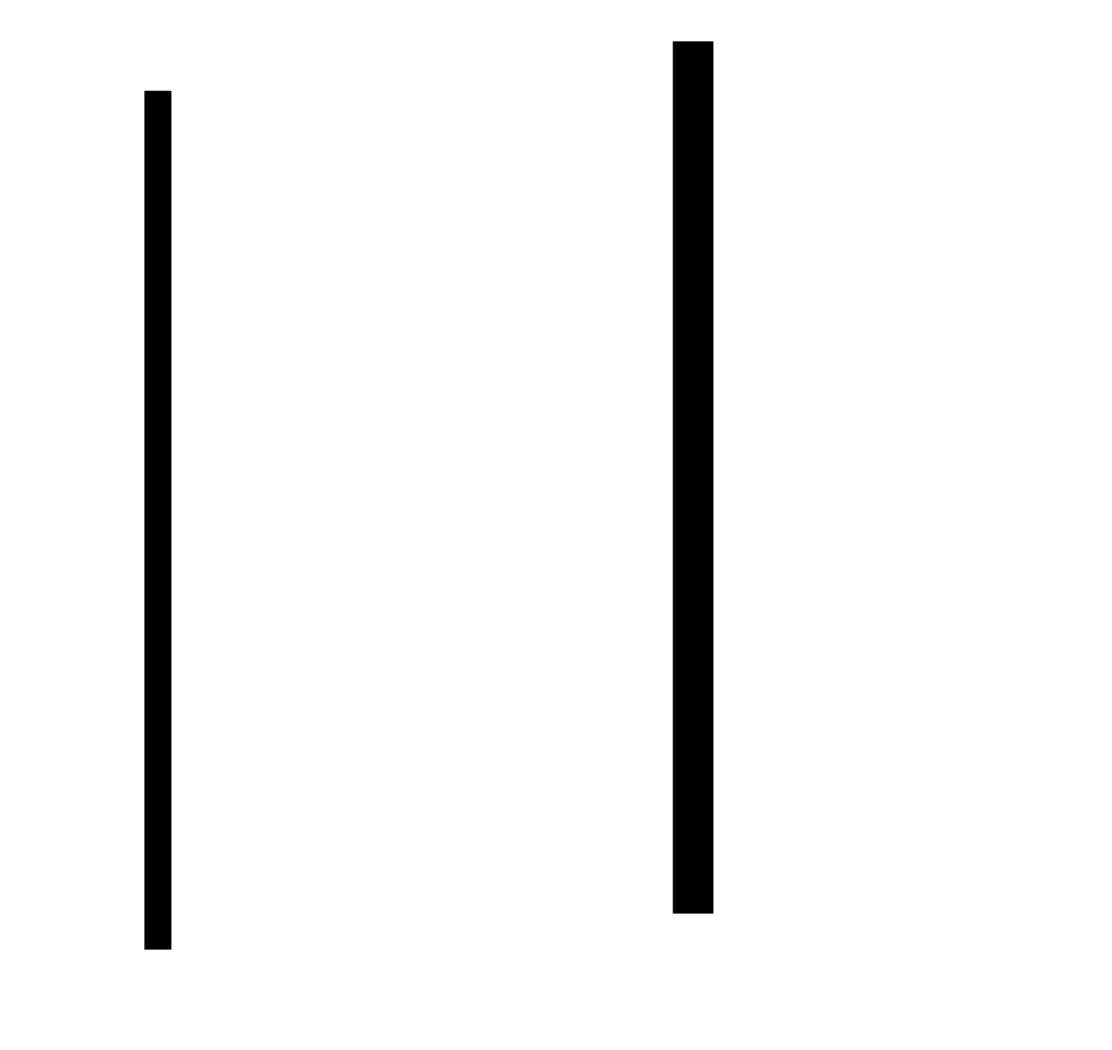} & \includegraphics[width=0.2\linewidth]{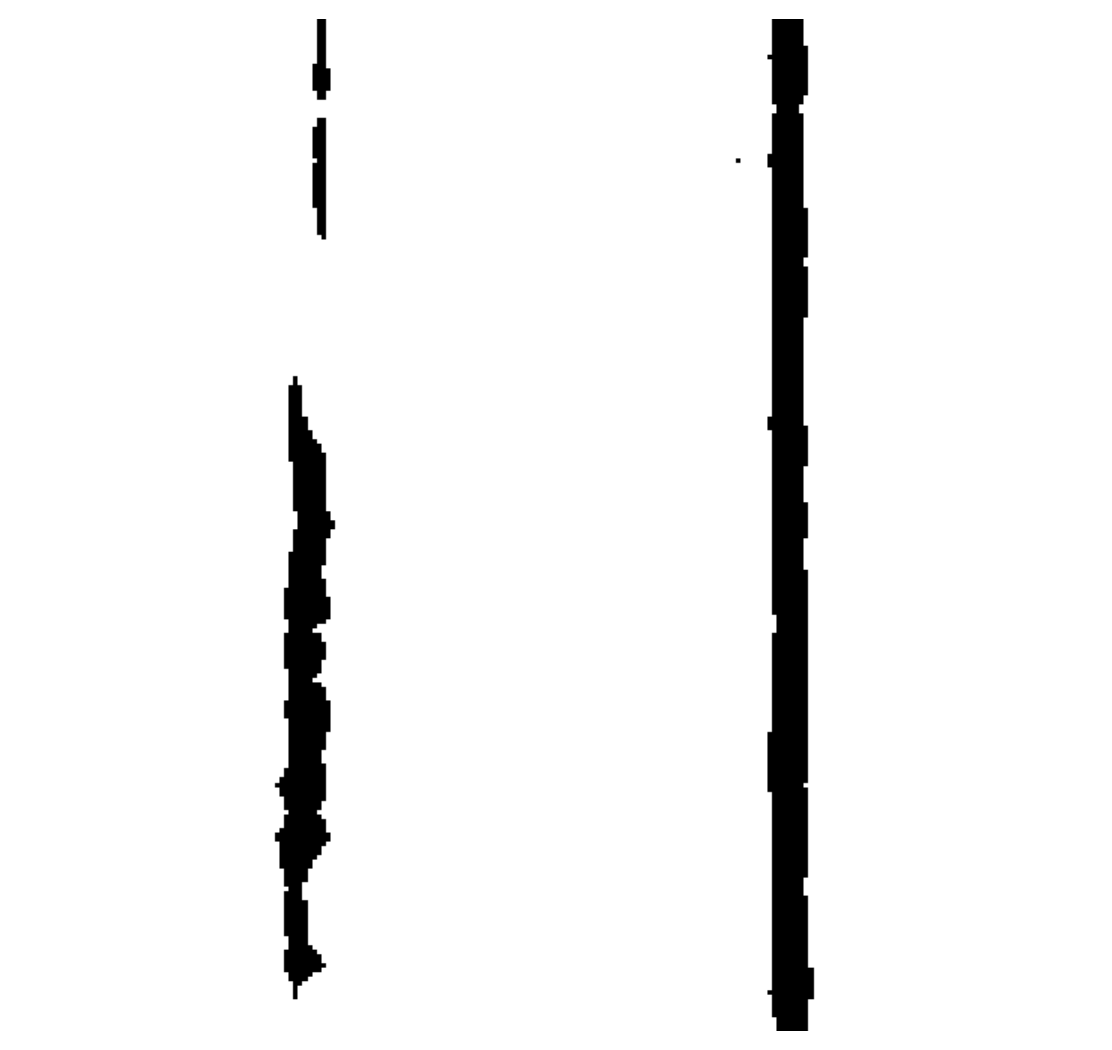} & \includegraphics[width=0.2\linewidth]{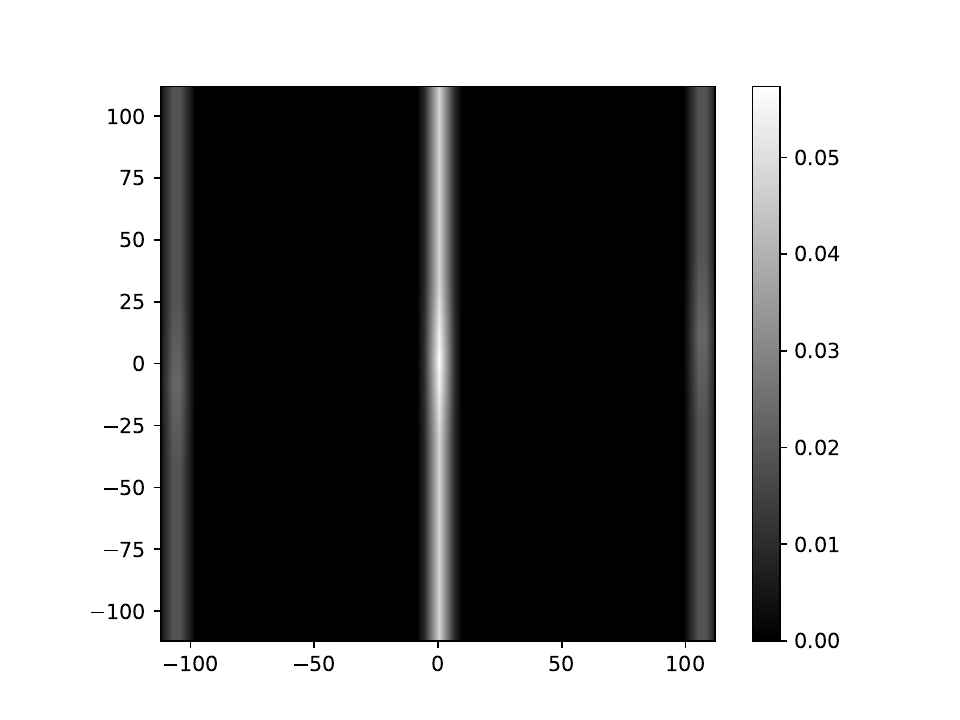} & \includegraphics[width=0.2\linewidth]{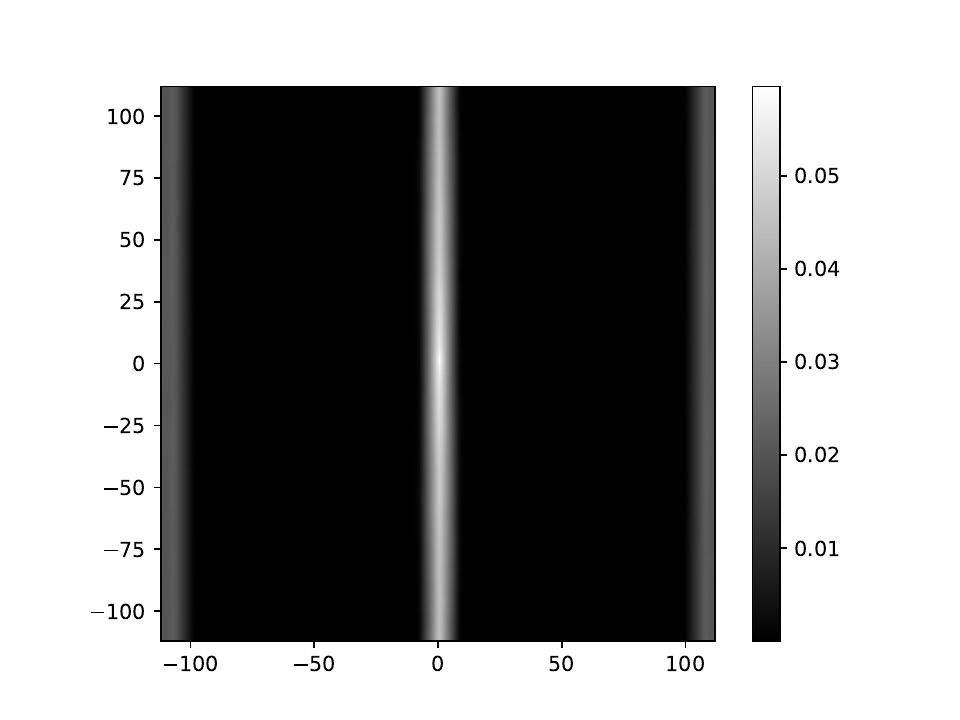} \\
\includegraphics[width=0.2\linewidth]{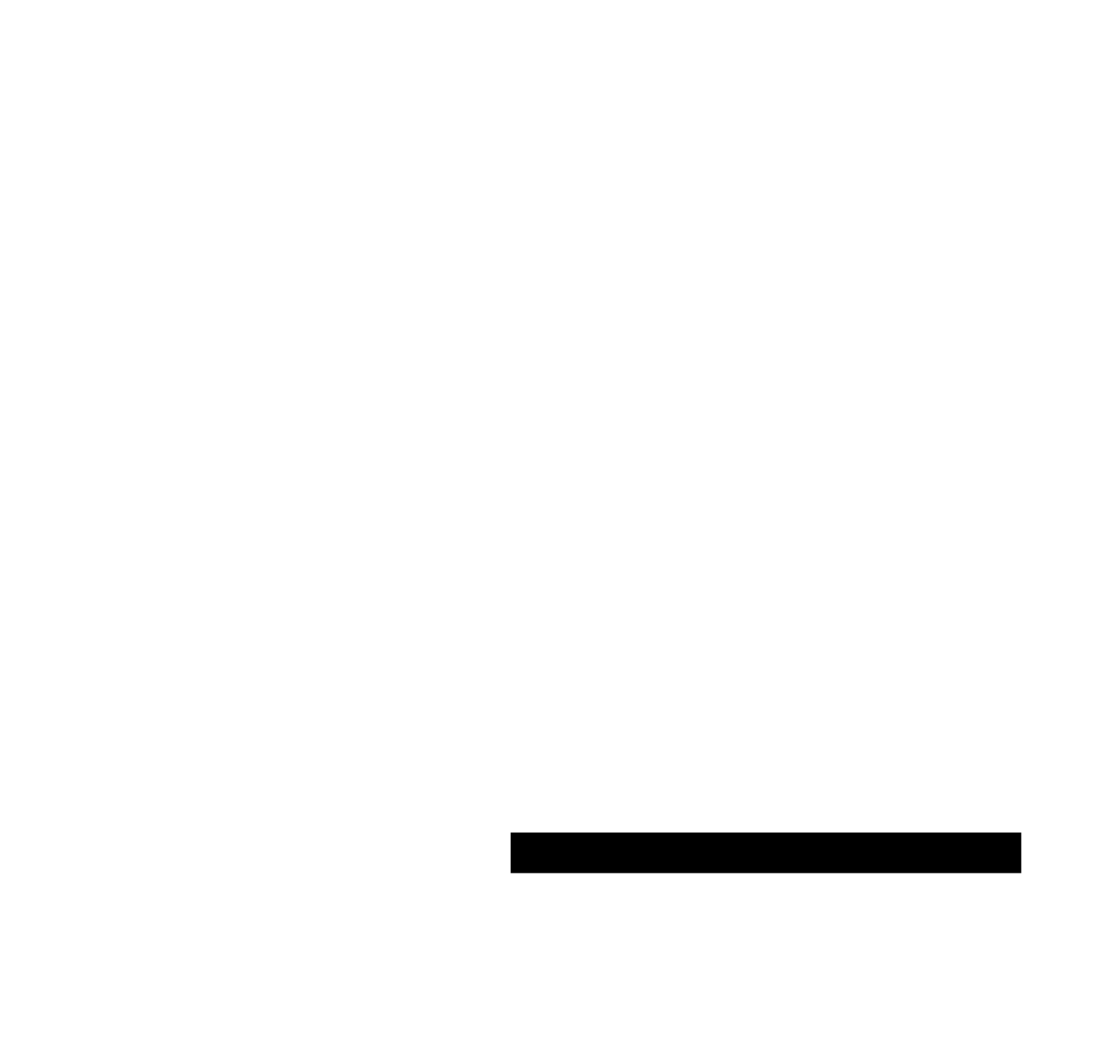} & \includegraphics[width=0.2\linewidth]{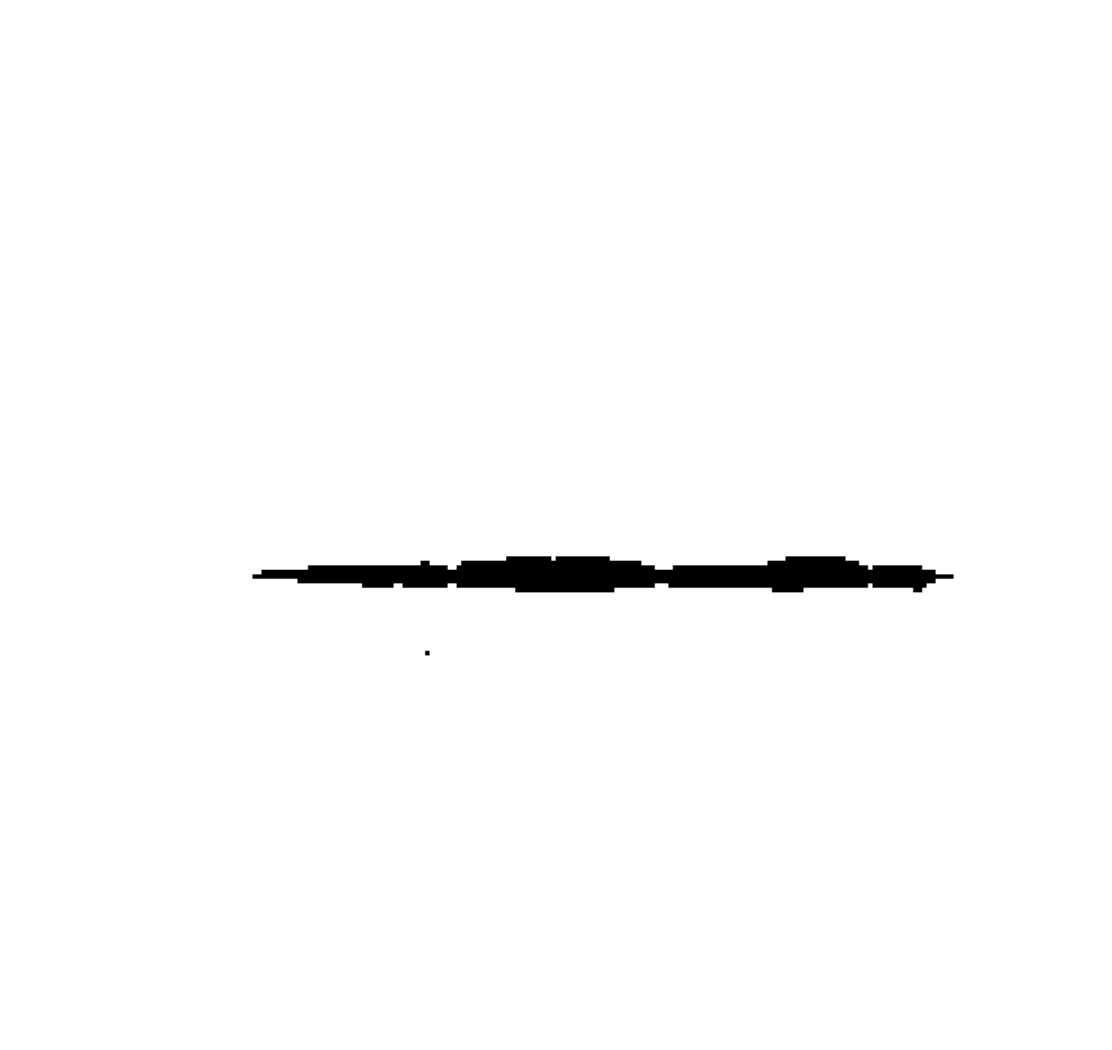} & \includegraphics[width=0.2\linewidth]{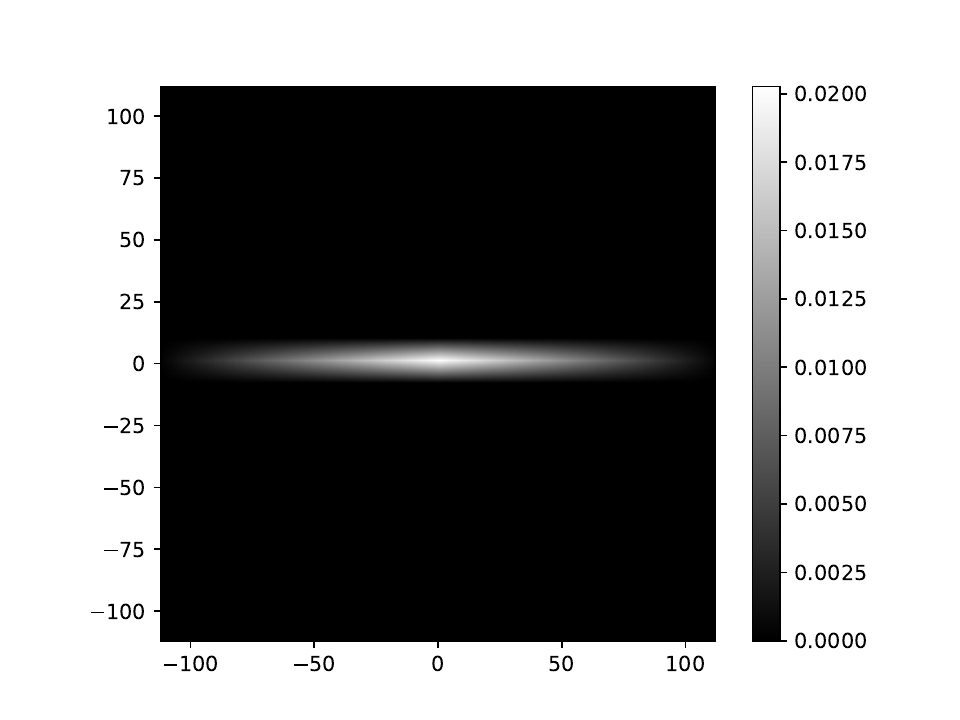} & \includegraphics[width=0.2\linewidth]{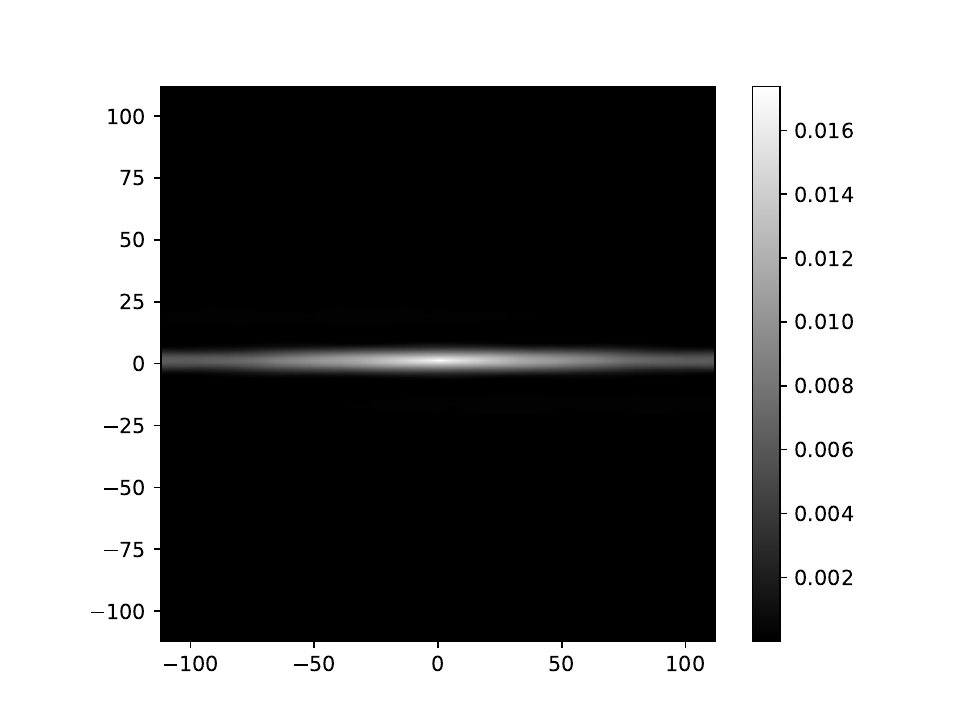} \\
\includegraphics[width=0.2\linewidth]{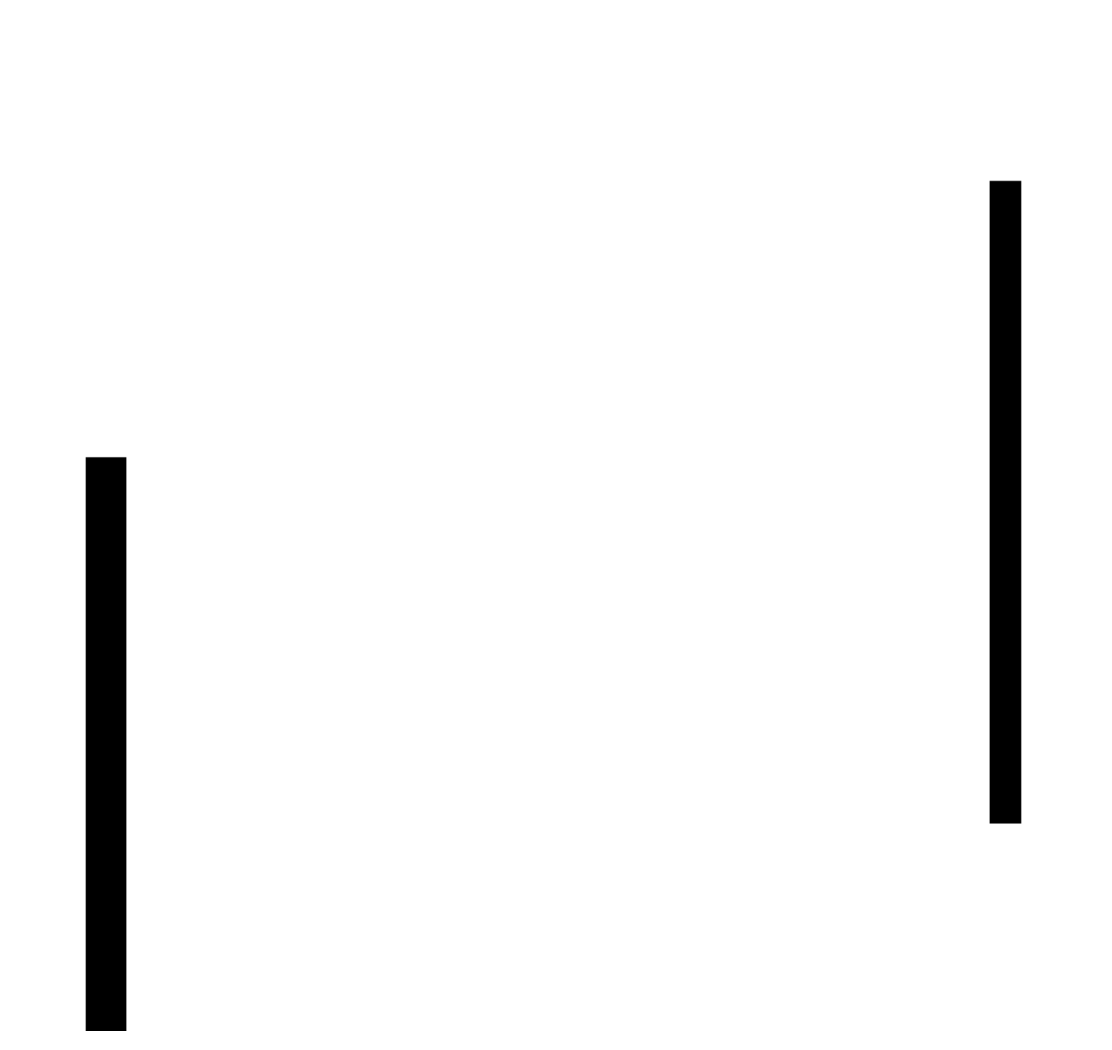} & \includegraphics[width=0.2\linewidth]{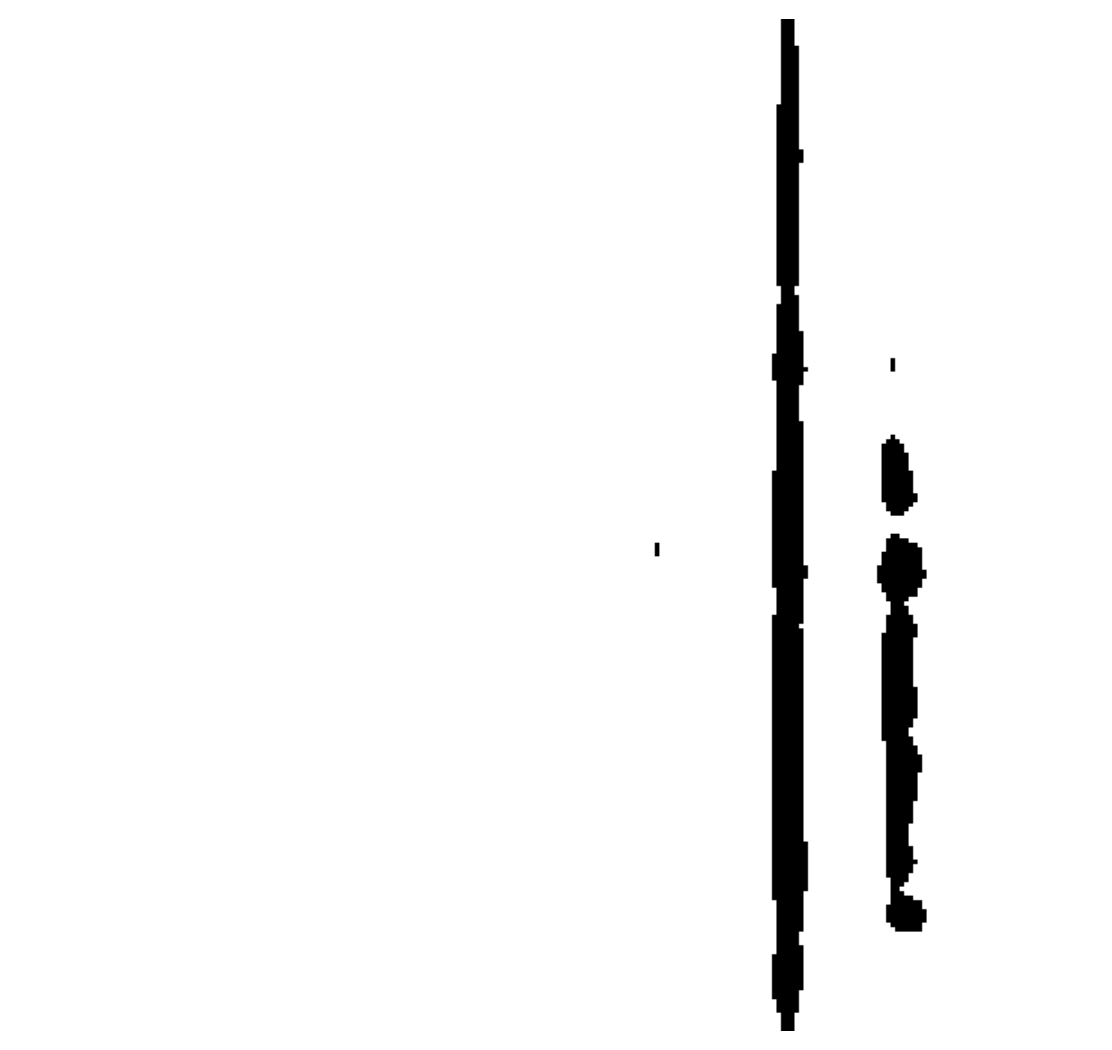} & \includegraphics[width=0.2\linewidth]{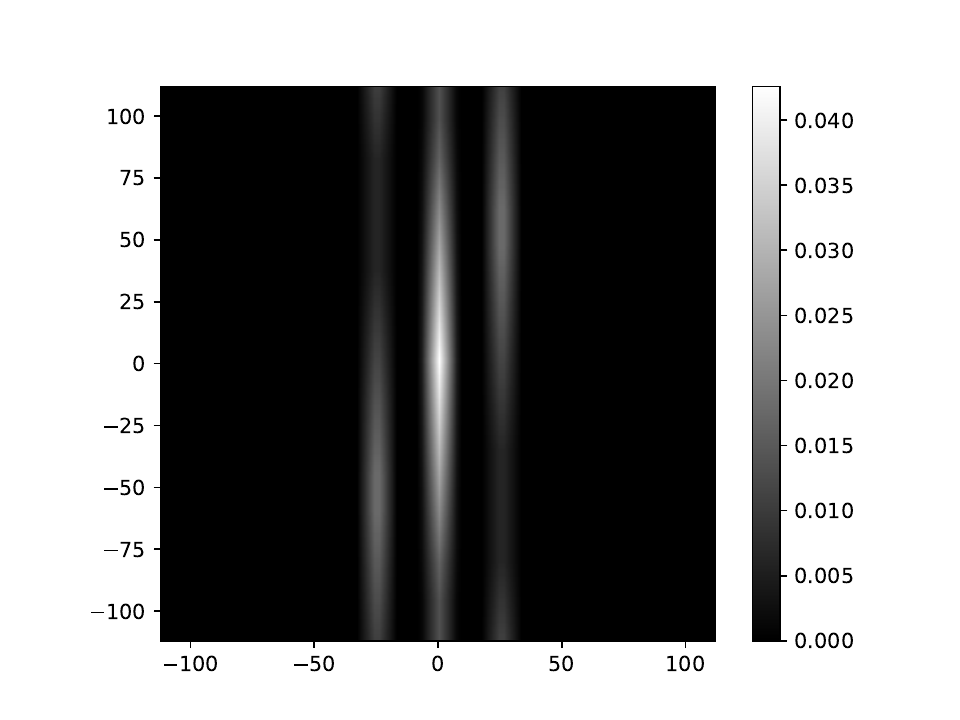} & \includegraphics[width=0.2\linewidth]{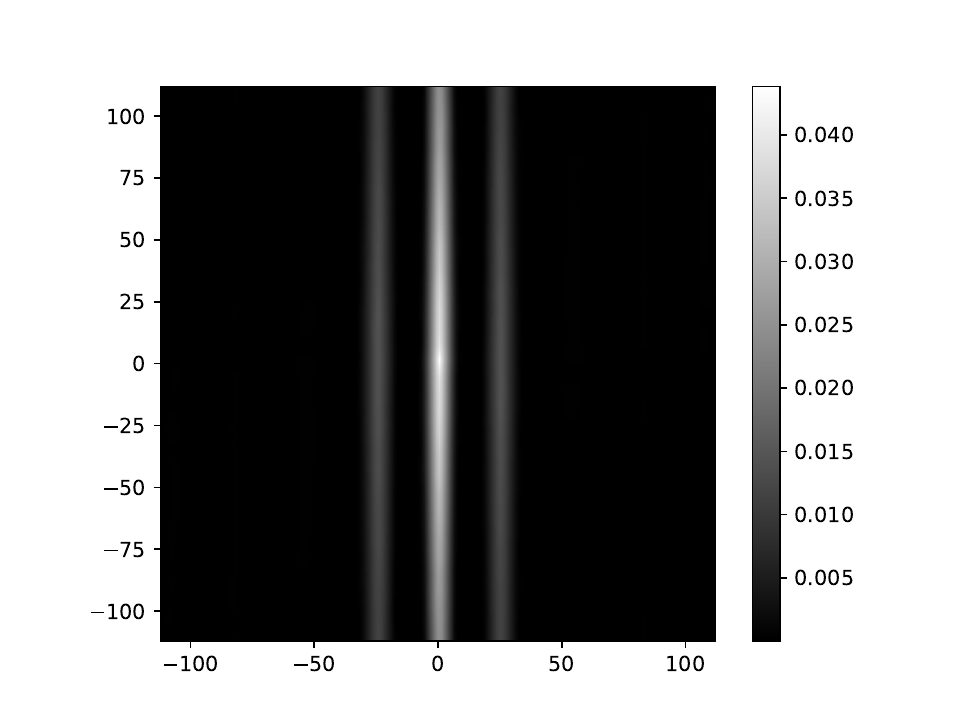} \\
\includegraphics[width=0.2\linewidth]{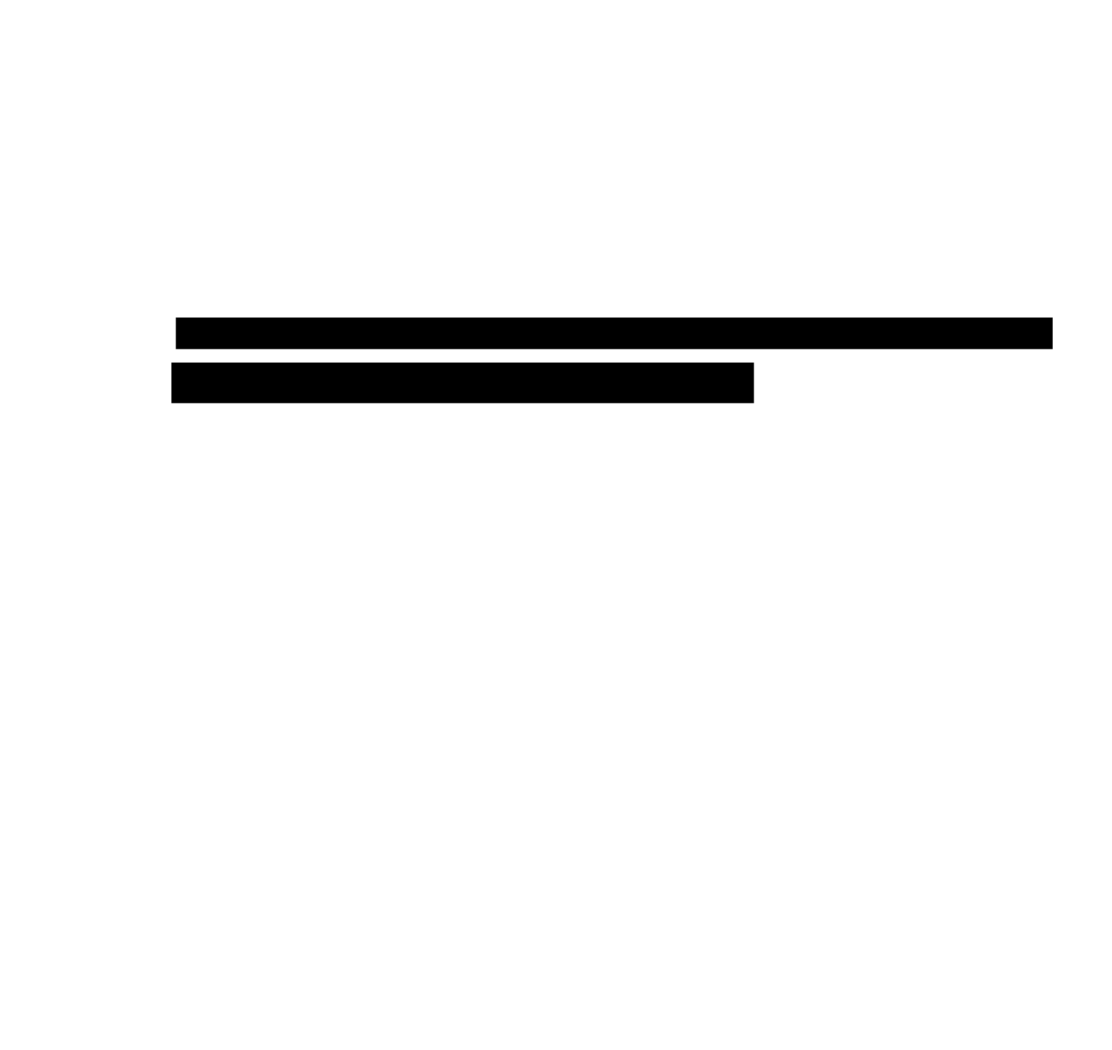} & \includegraphics[width=0.2\linewidth]{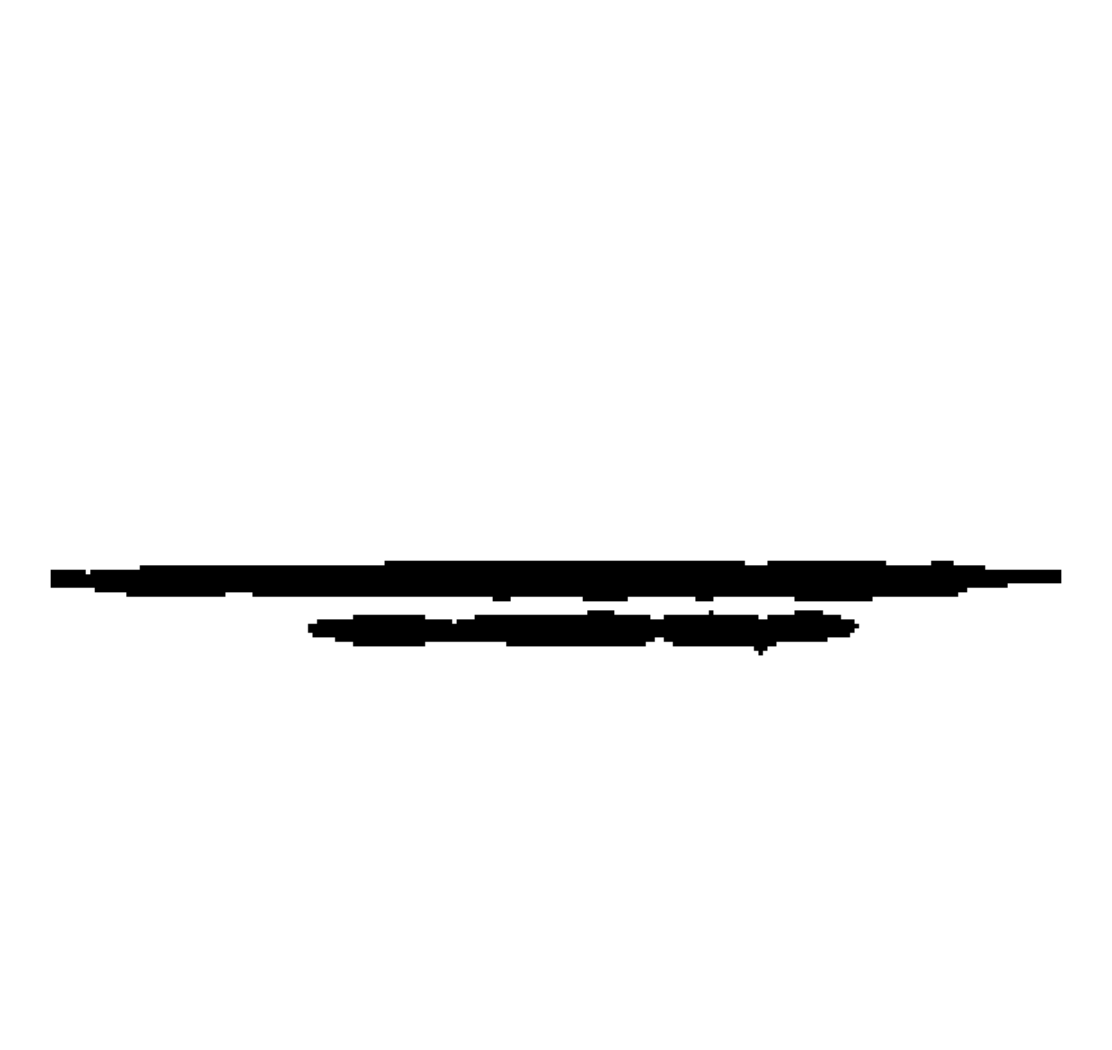} & \includegraphics[width=0.2\linewidth]{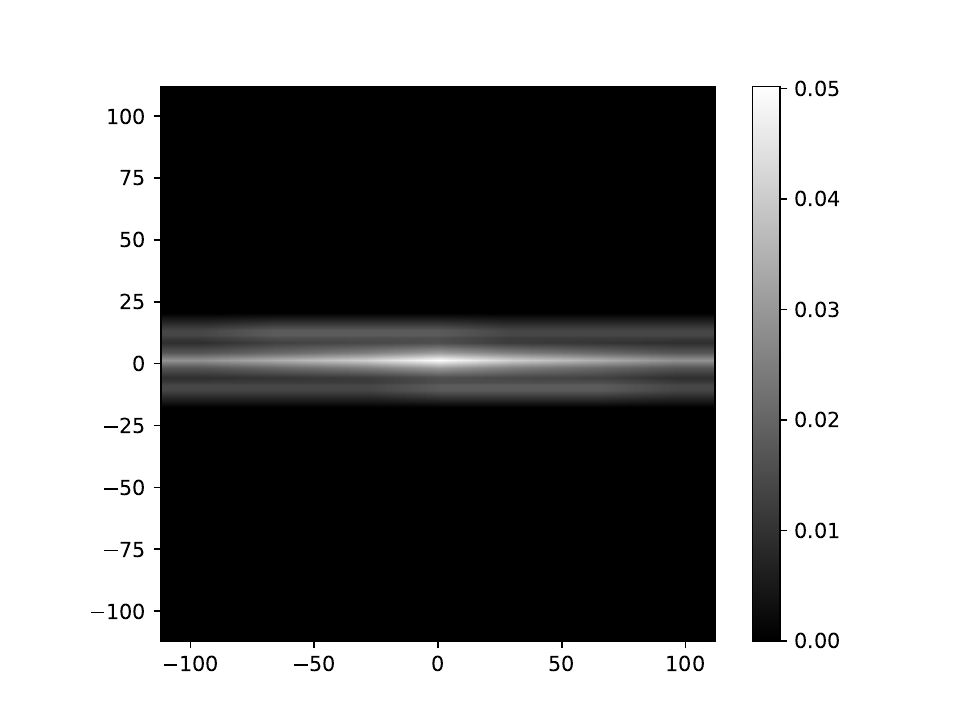} & \includegraphics[width=0.2\linewidth]{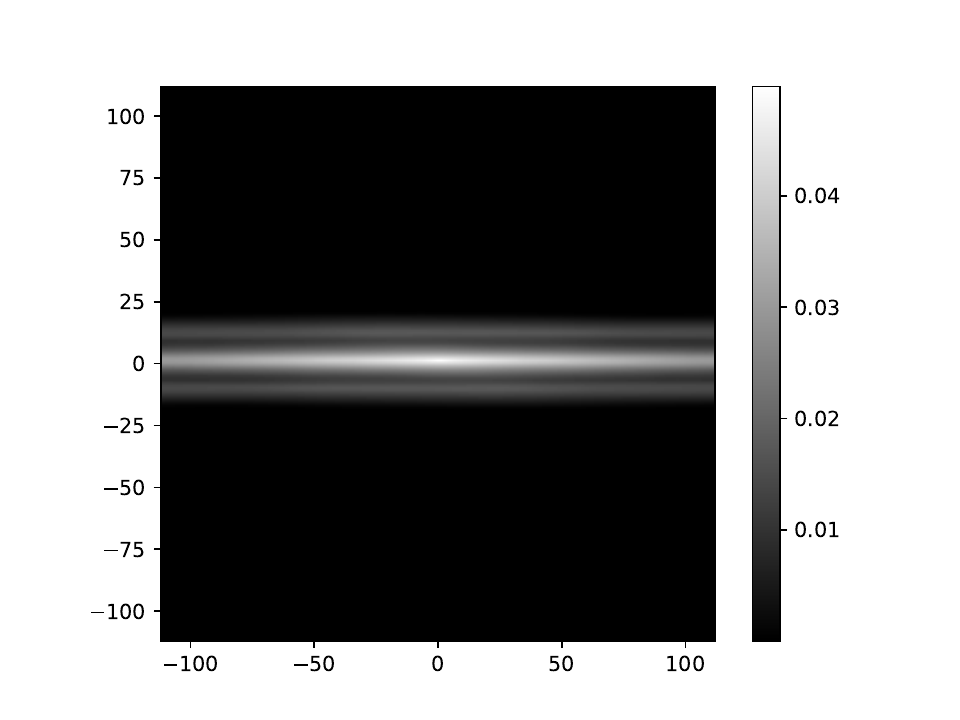} \\
\includegraphics[width=0.2\linewidth]{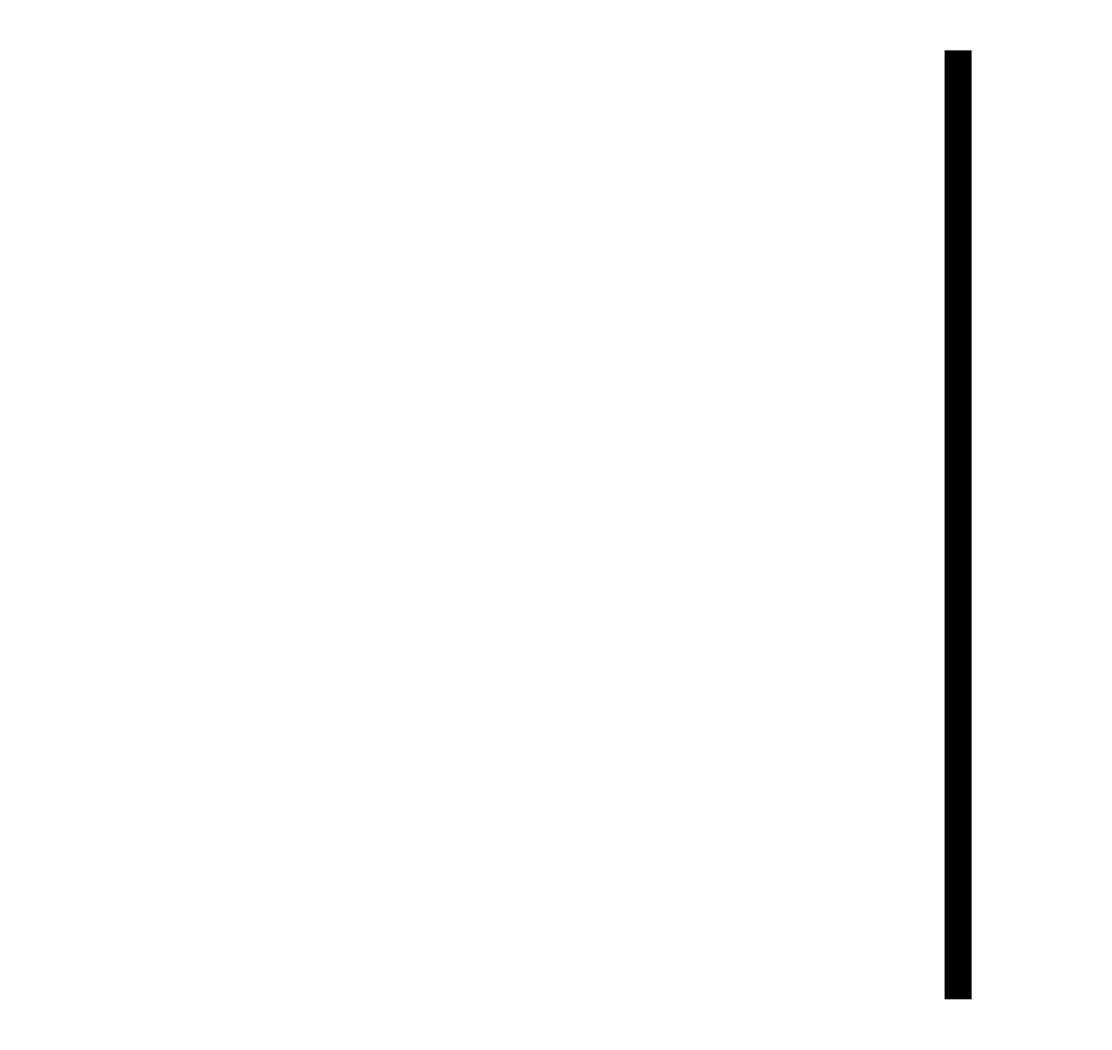} & \includegraphics[width=0.2\linewidth]{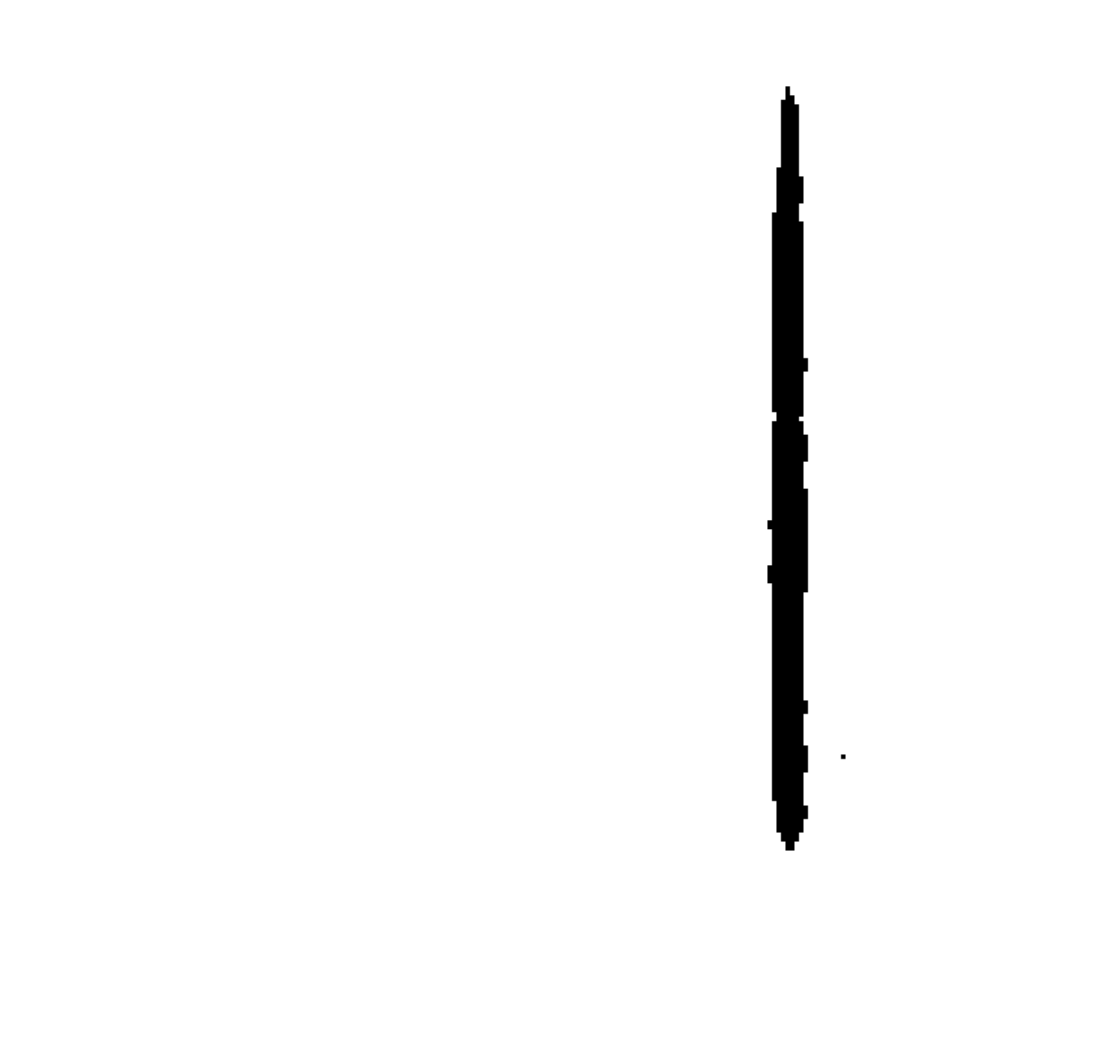} & \includegraphics[width=0.2\linewidth]{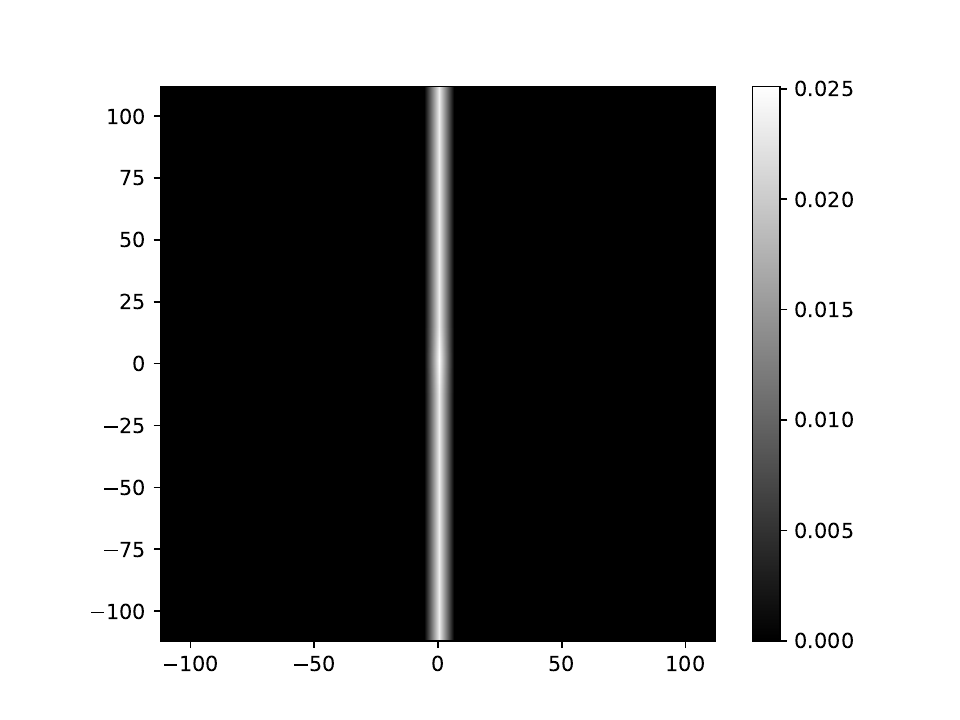} & \includegraphics[width=0.2\linewidth]{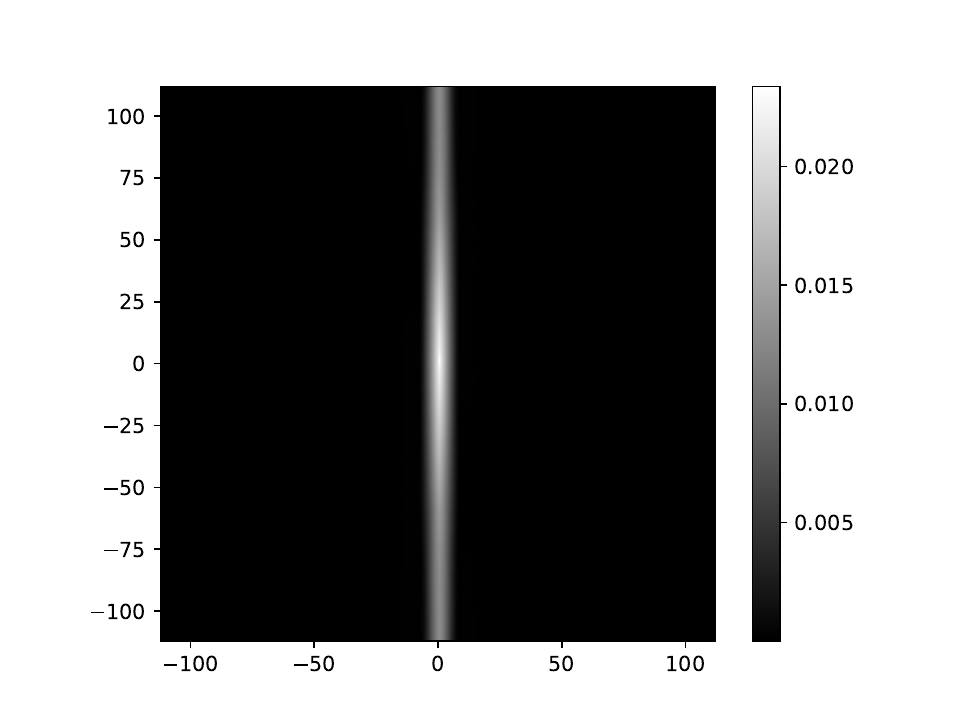} \\
\hline
\end{tabular}
\end{center}
\end{table}

\begin{table}[H]
\caption{Original, reconstructed, and spatial statistics difference}
\begin{center}
\begin{tabular}{cccc}
\hline
\multicolumn{2}{c}{\bf Data} & \multicolumn{2}{c}{\bf Auto-correlations} \\
\hline
{\bf Original} & {\bf Reconstructed} & {\bf Original} & {\bf Reconstructed} \\
\hline
\includegraphics[width=0.2\linewidth]{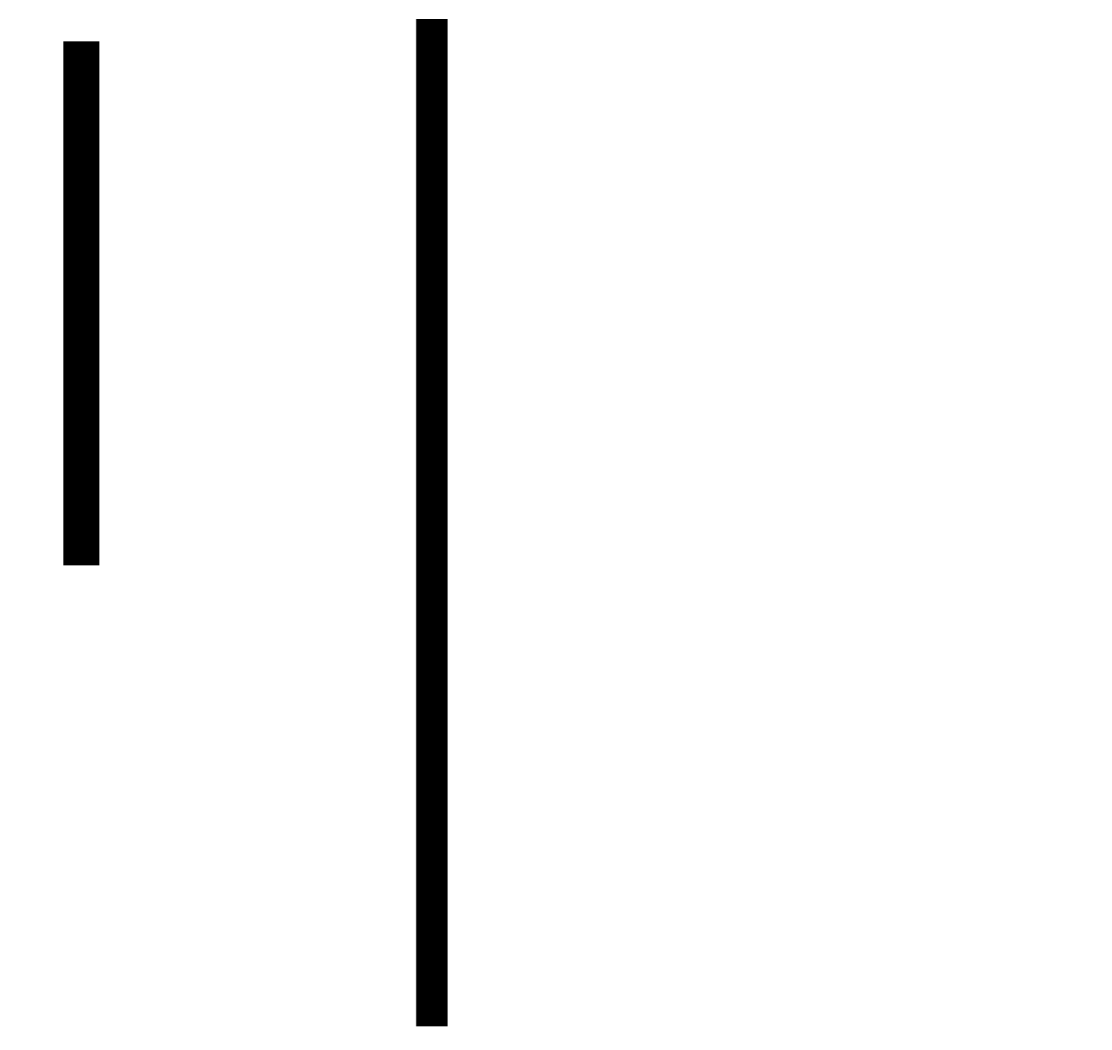} & \includegraphics[width=0.2\linewidth]{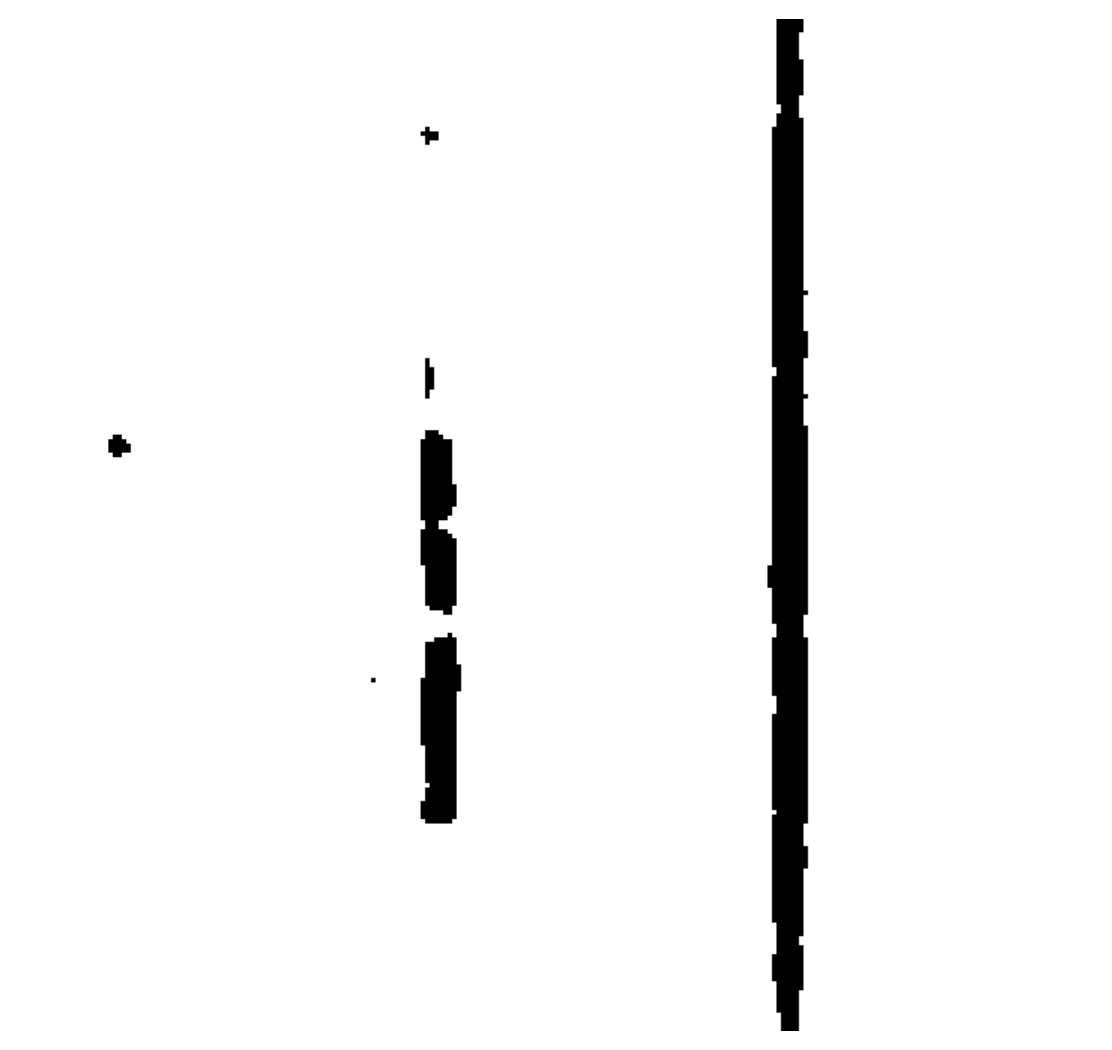} & \includegraphics[width=0.2\linewidth]{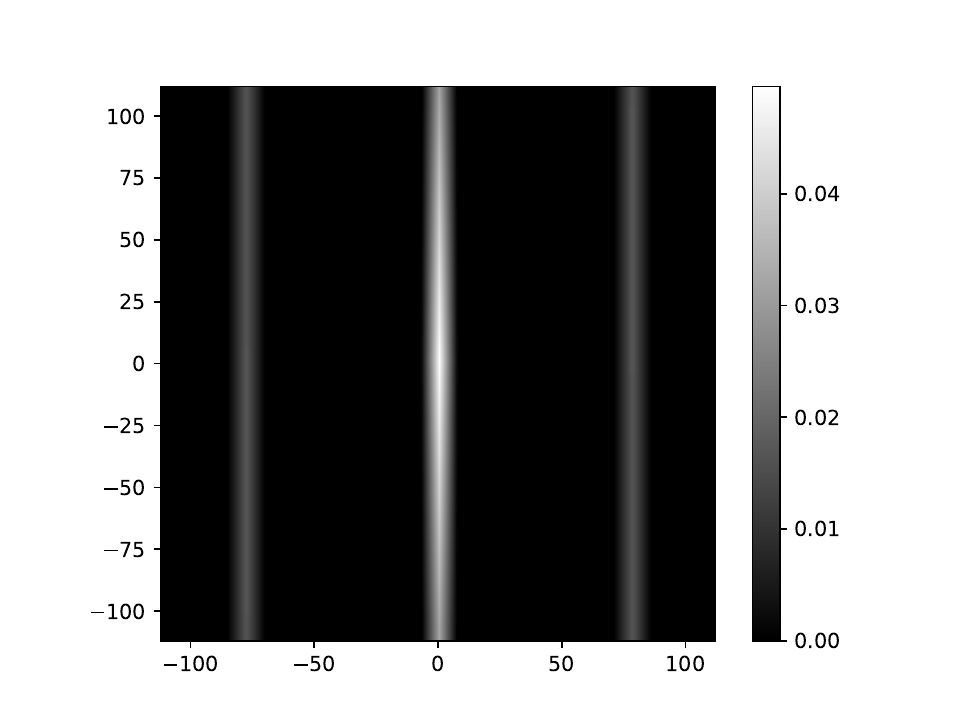} & \includegraphics[width=0.2\linewidth]{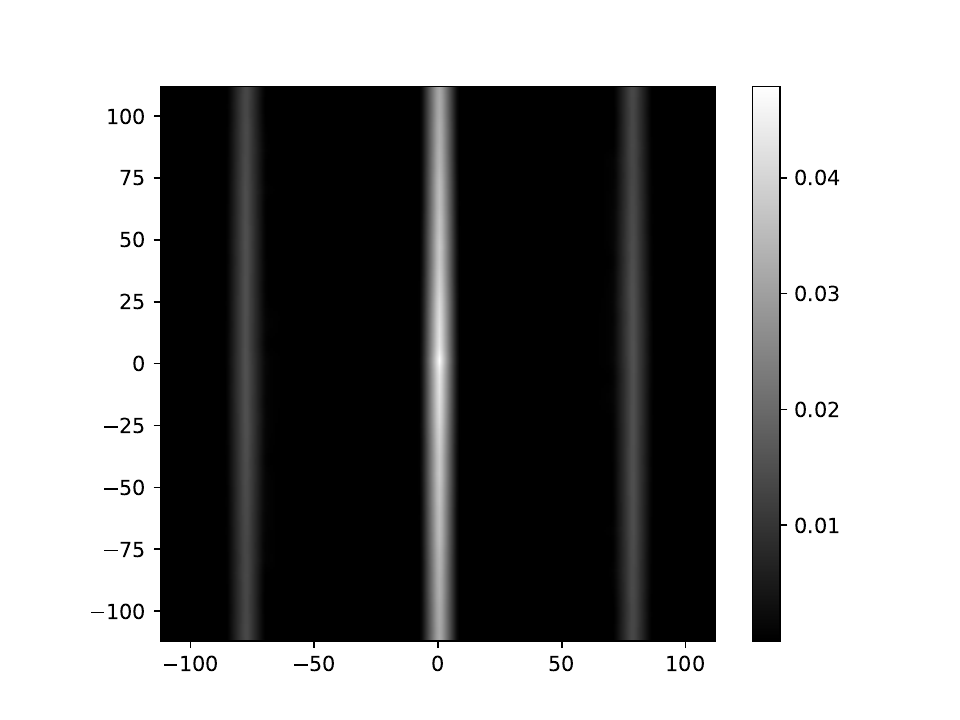} \\
\includegraphics[width=0.2\linewidth]{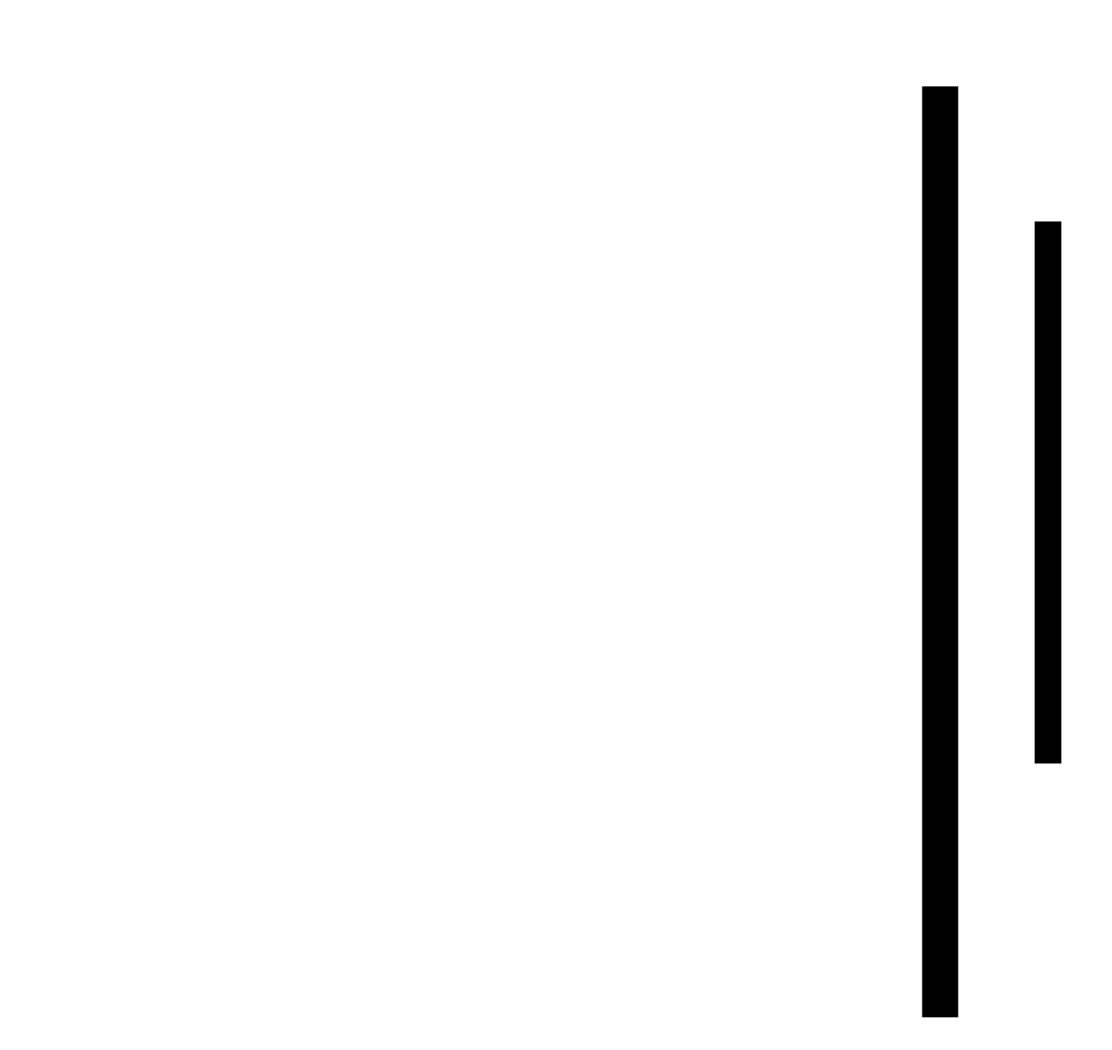} & \includegraphics[width=0.2\linewidth]{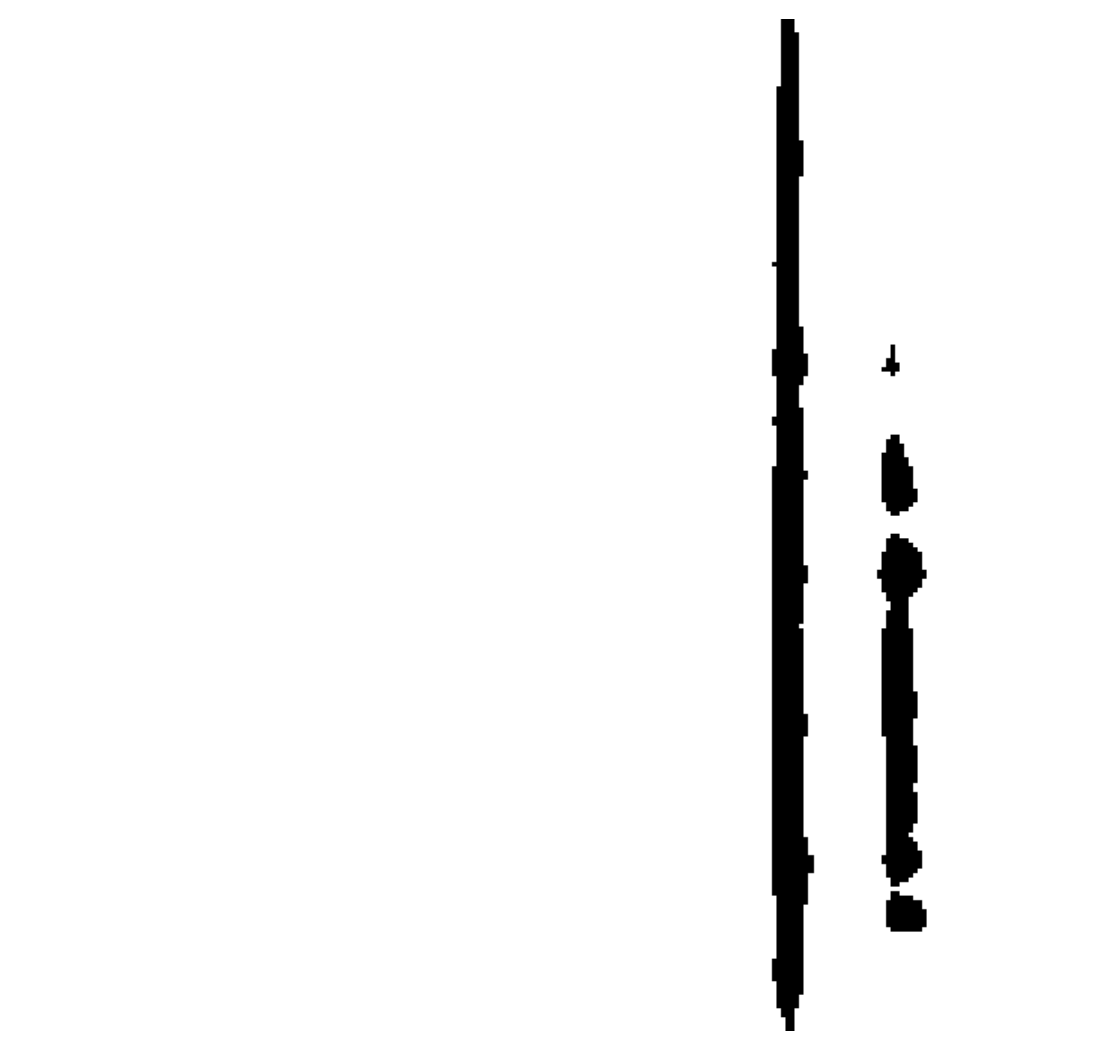} & \includegraphics[width=0.2\linewidth]{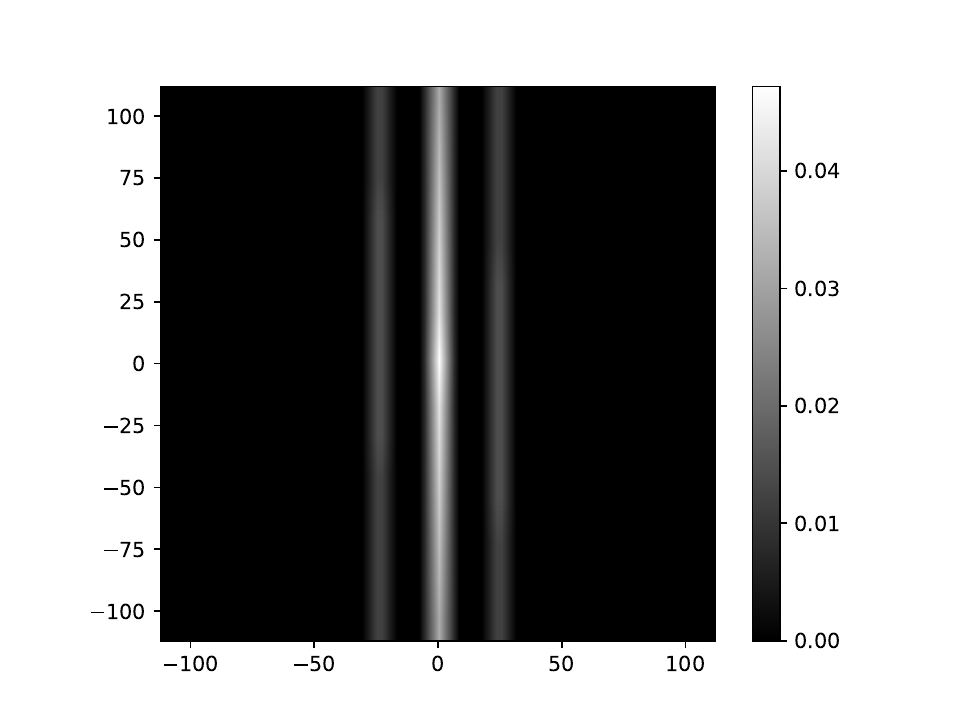} & \includegraphics[width=0.2\linewidth]{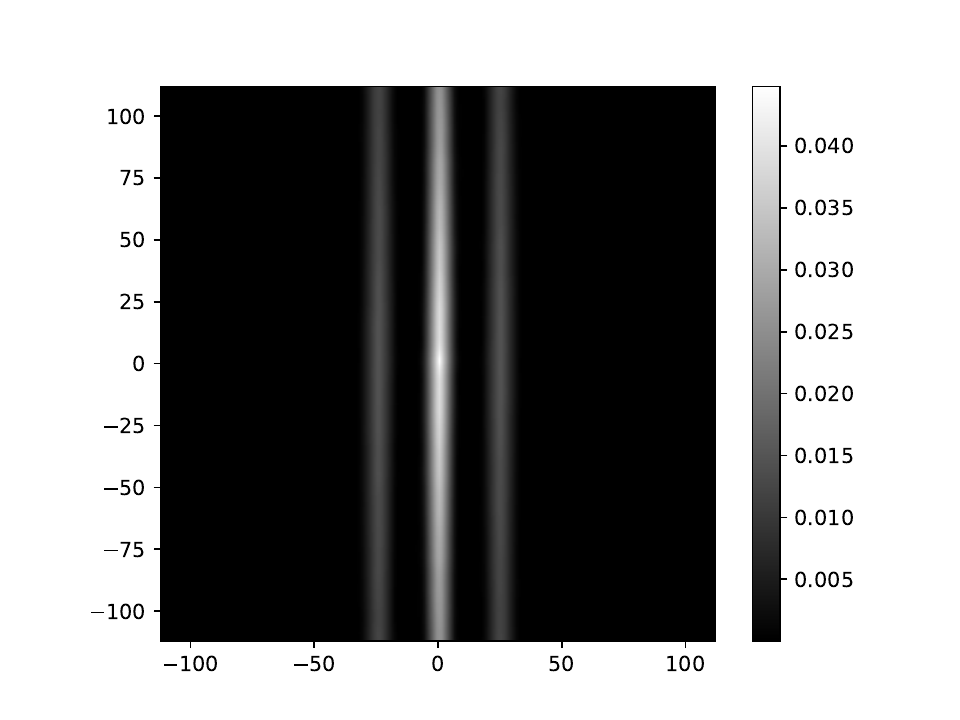} \\
\includegraphics[width=0.2\linewidth]{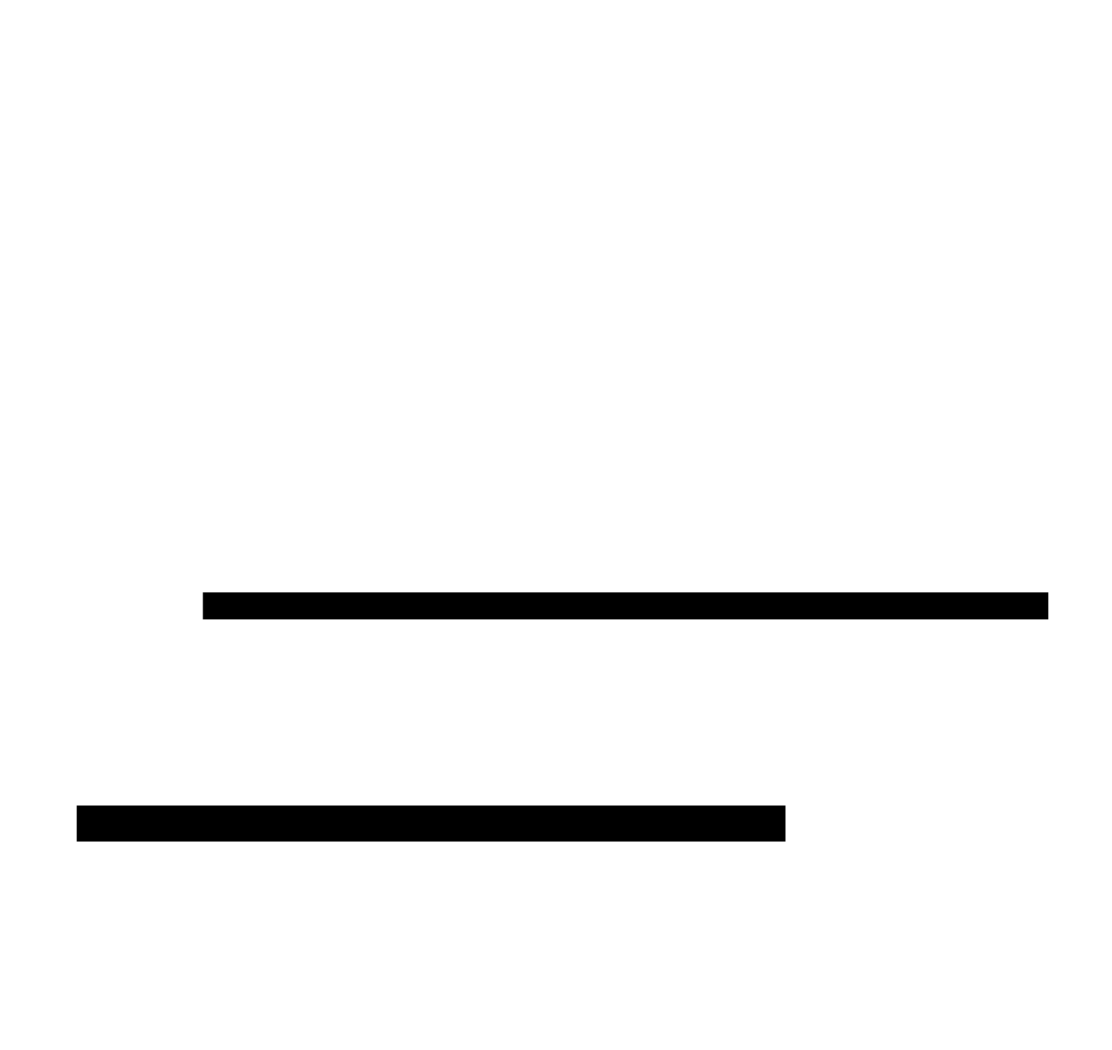} & \includegraphics[width=0.2\linewidth]{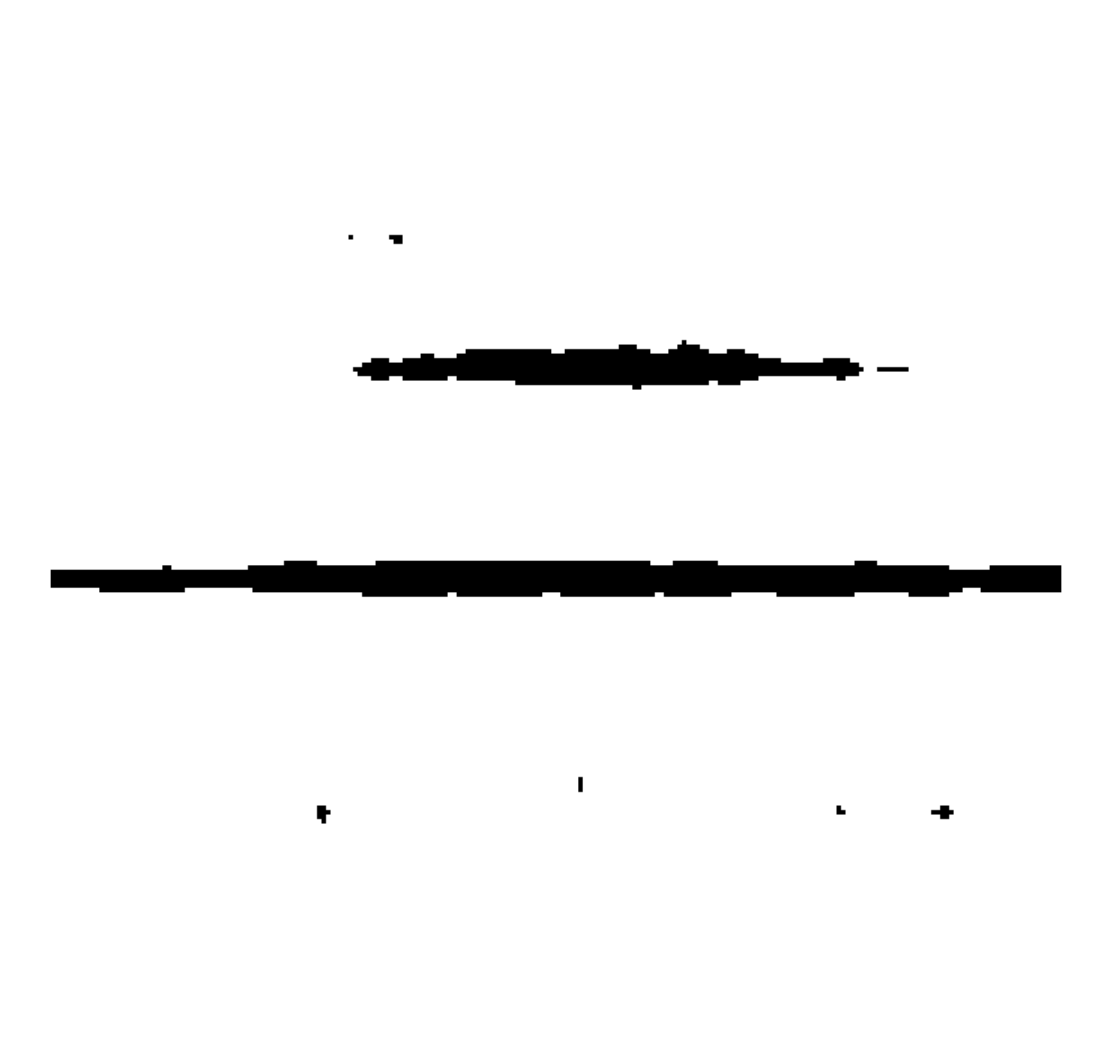} & \includegraphics[width=0.2\linewidth]{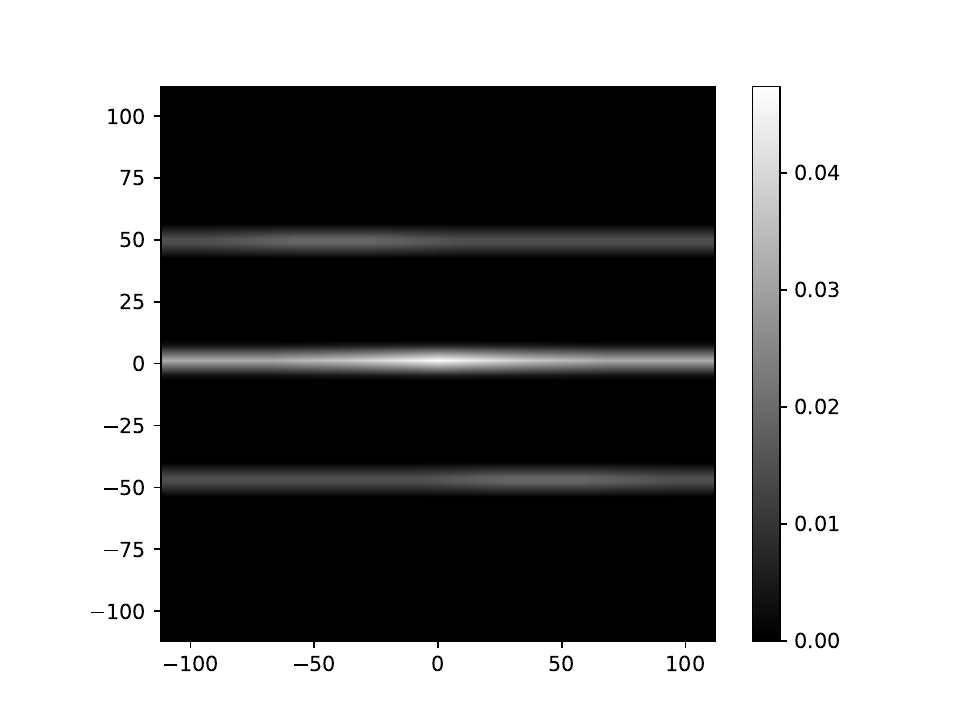} & \includegraphics[width=0.2\linewidth]{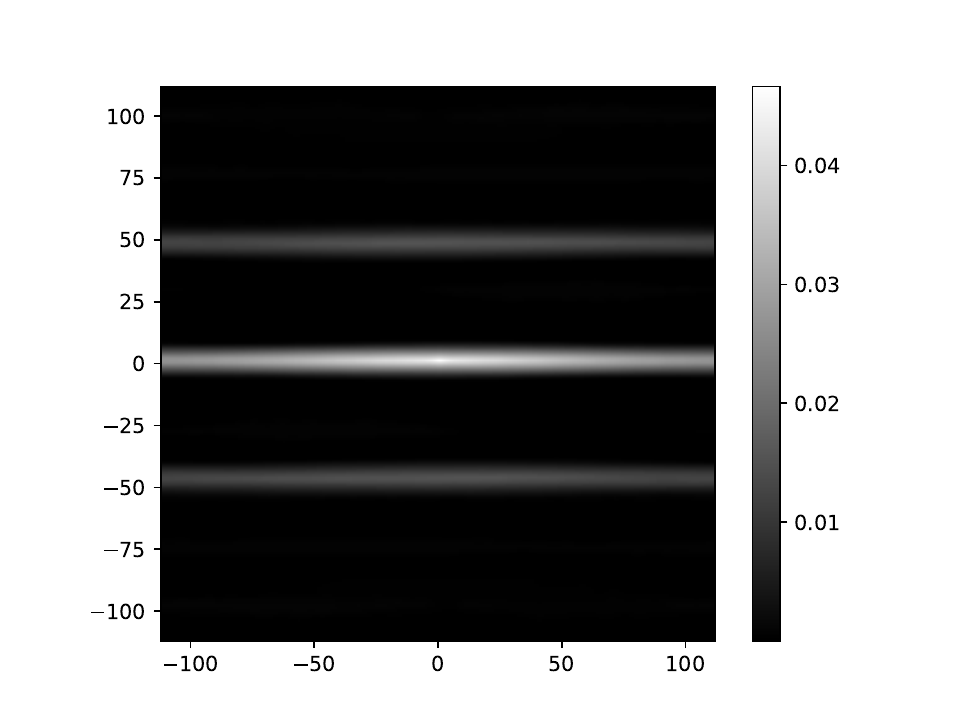} \\
\includegraphics[width=0.2\linewidth]{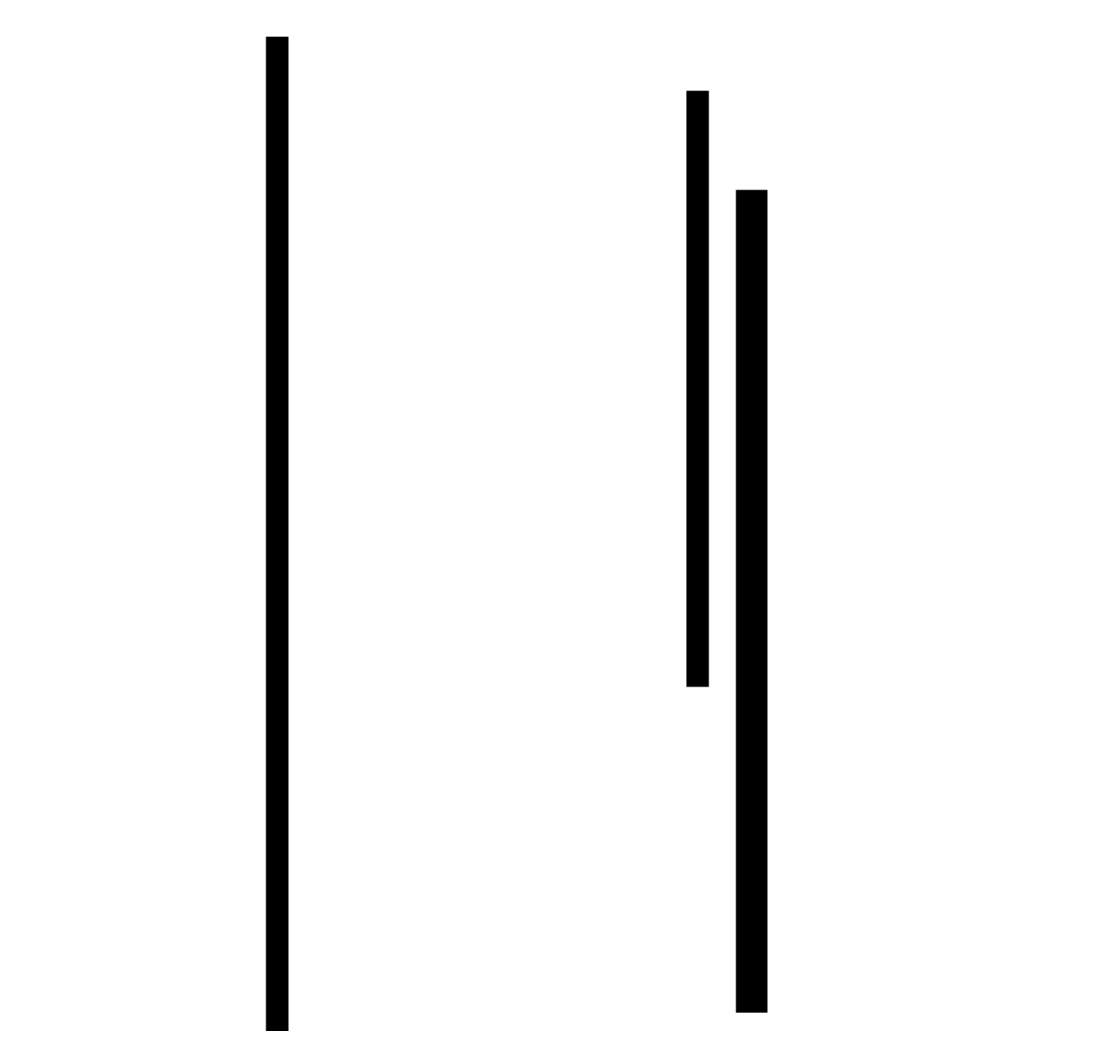} & \includegraphics[width=0.2\linewidth]{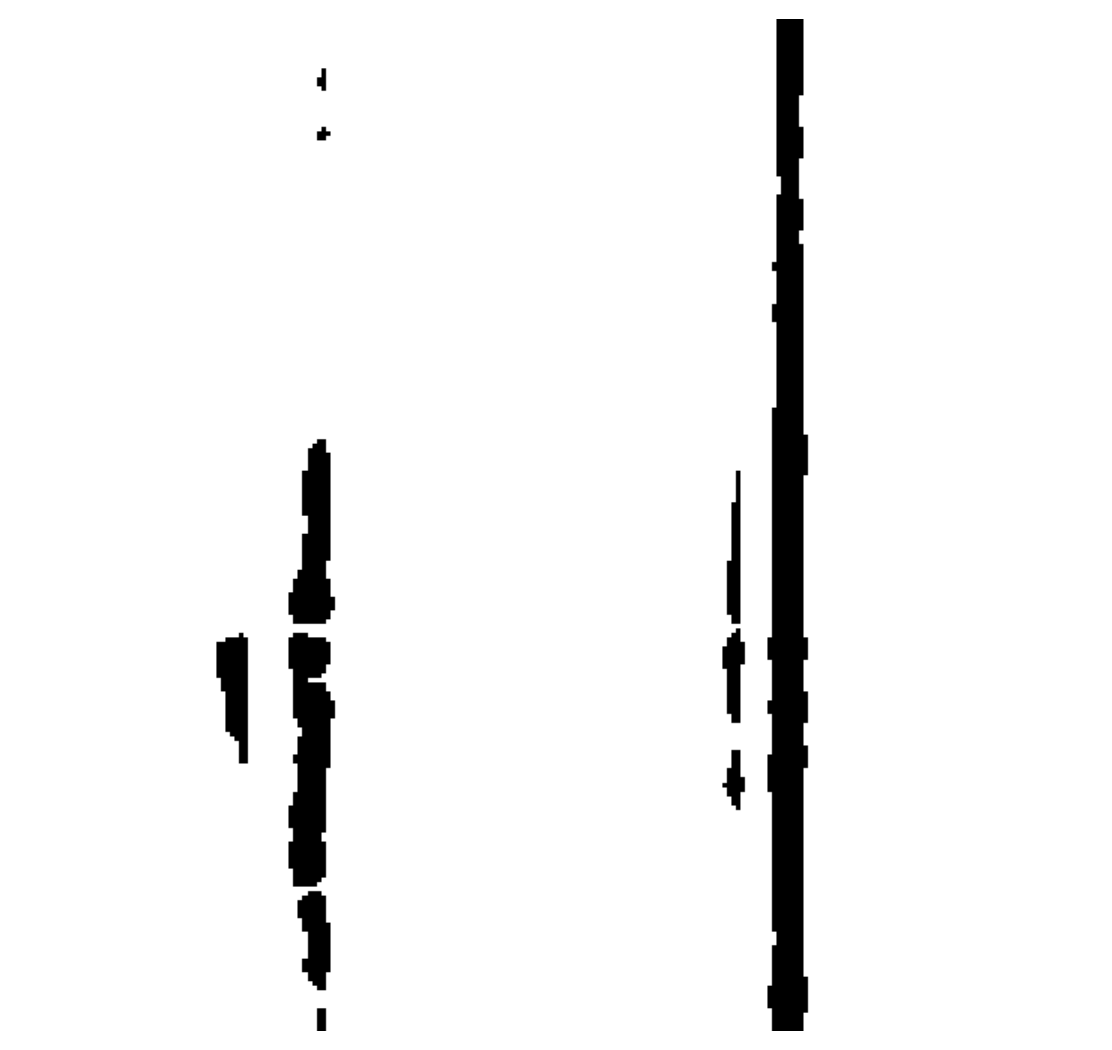} & \includegraphics[width=0.2\linewidth]{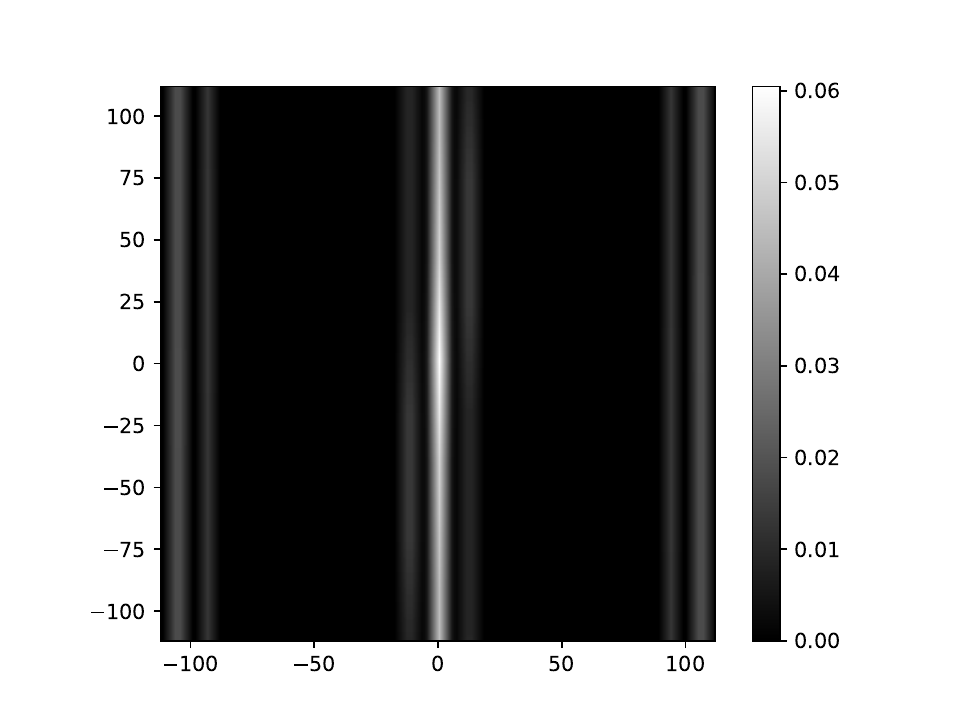} & \includegraphics[width=0.2\linewidth]{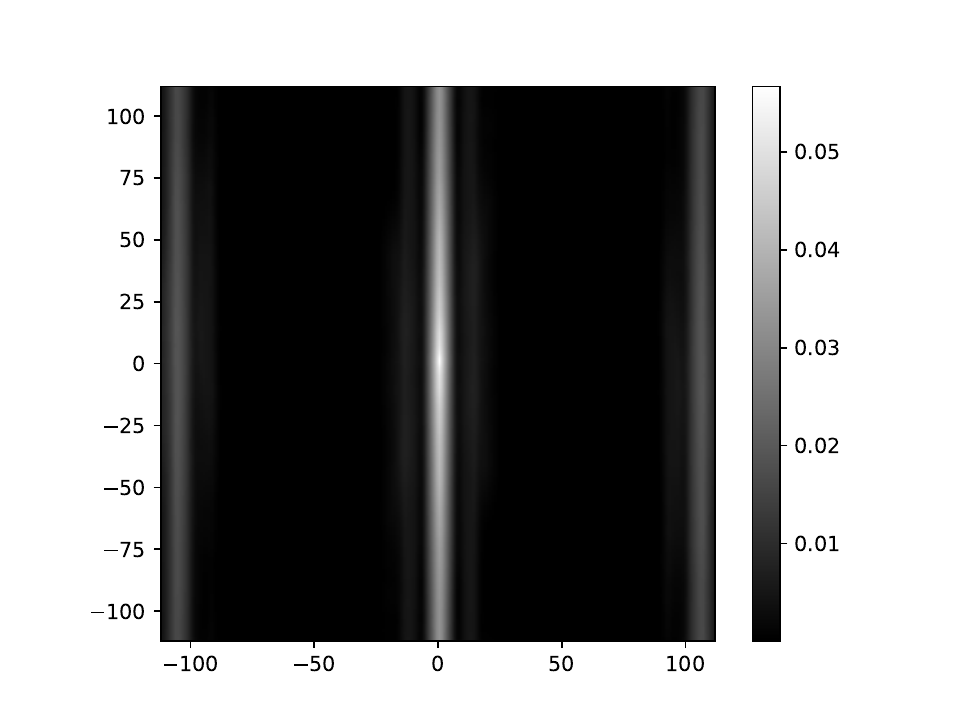} \\
\includegraphics[width=0.2\linewidth]{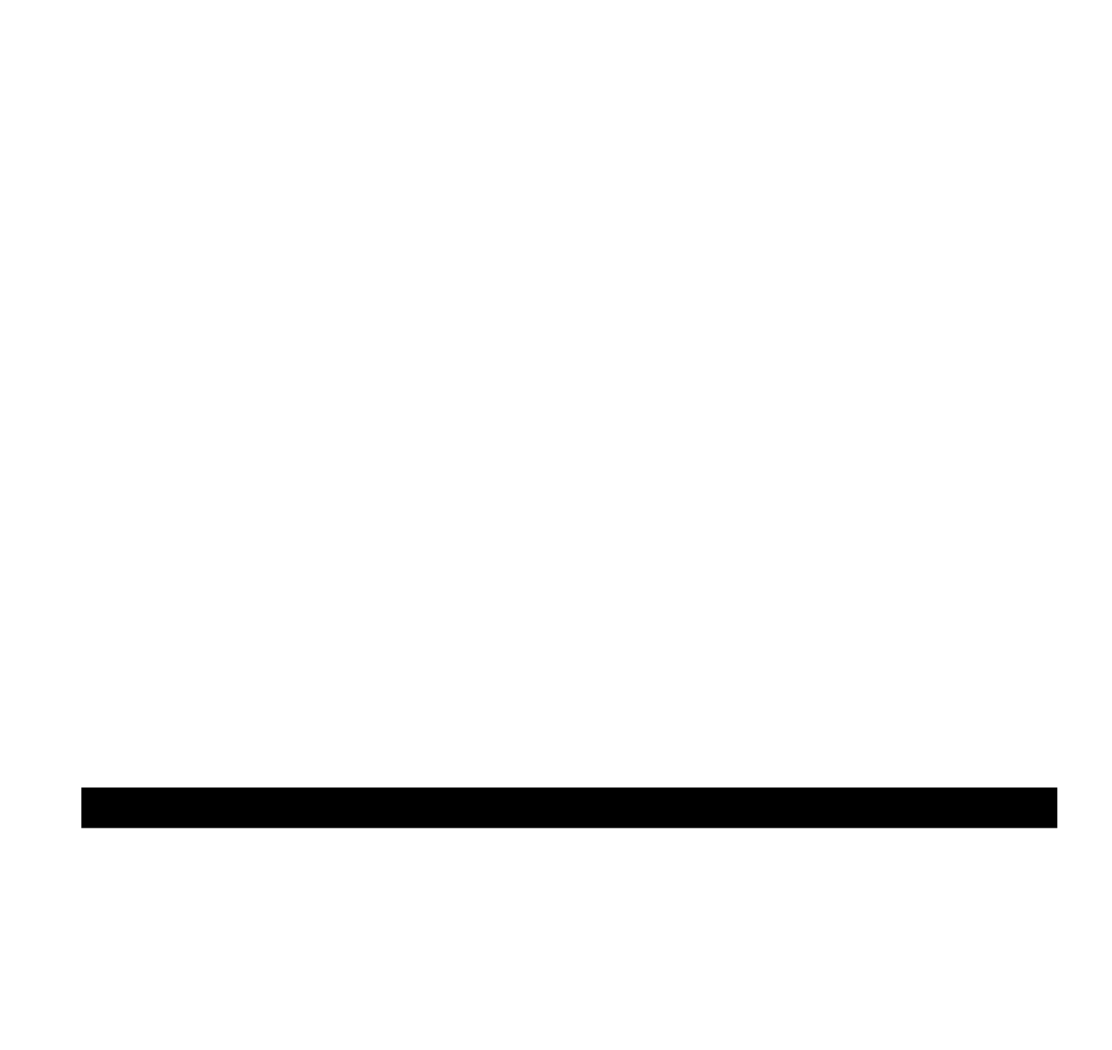} & \includegraphics[width=0.2\linewidth]{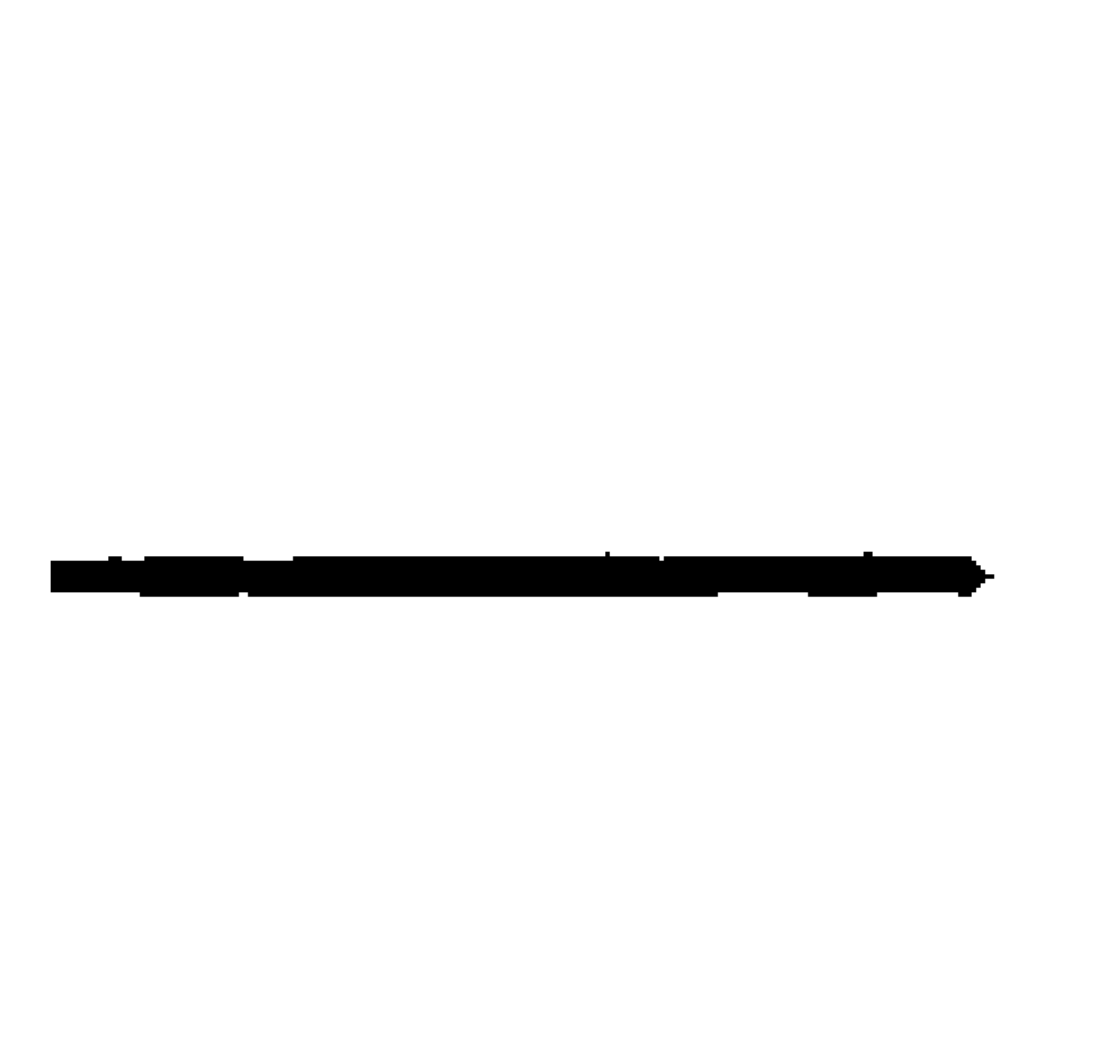} & \includegraphics[width=0.2\linewidth]{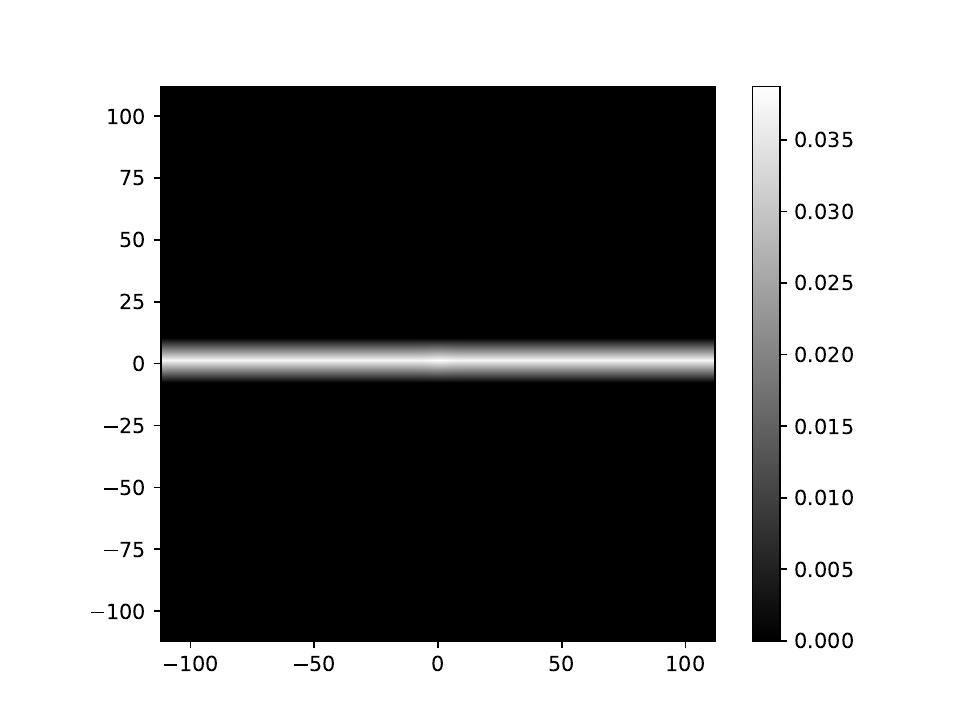} & \includegraphics[width=0.2\linewidth]{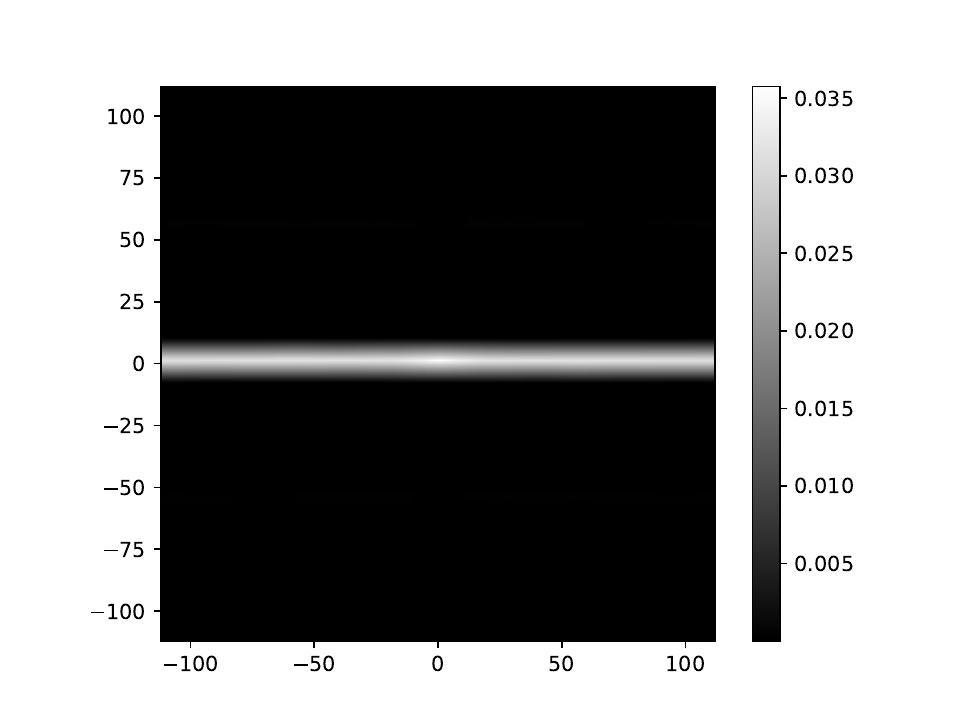} \\
\includegraphics[width=0.2\linewidth]{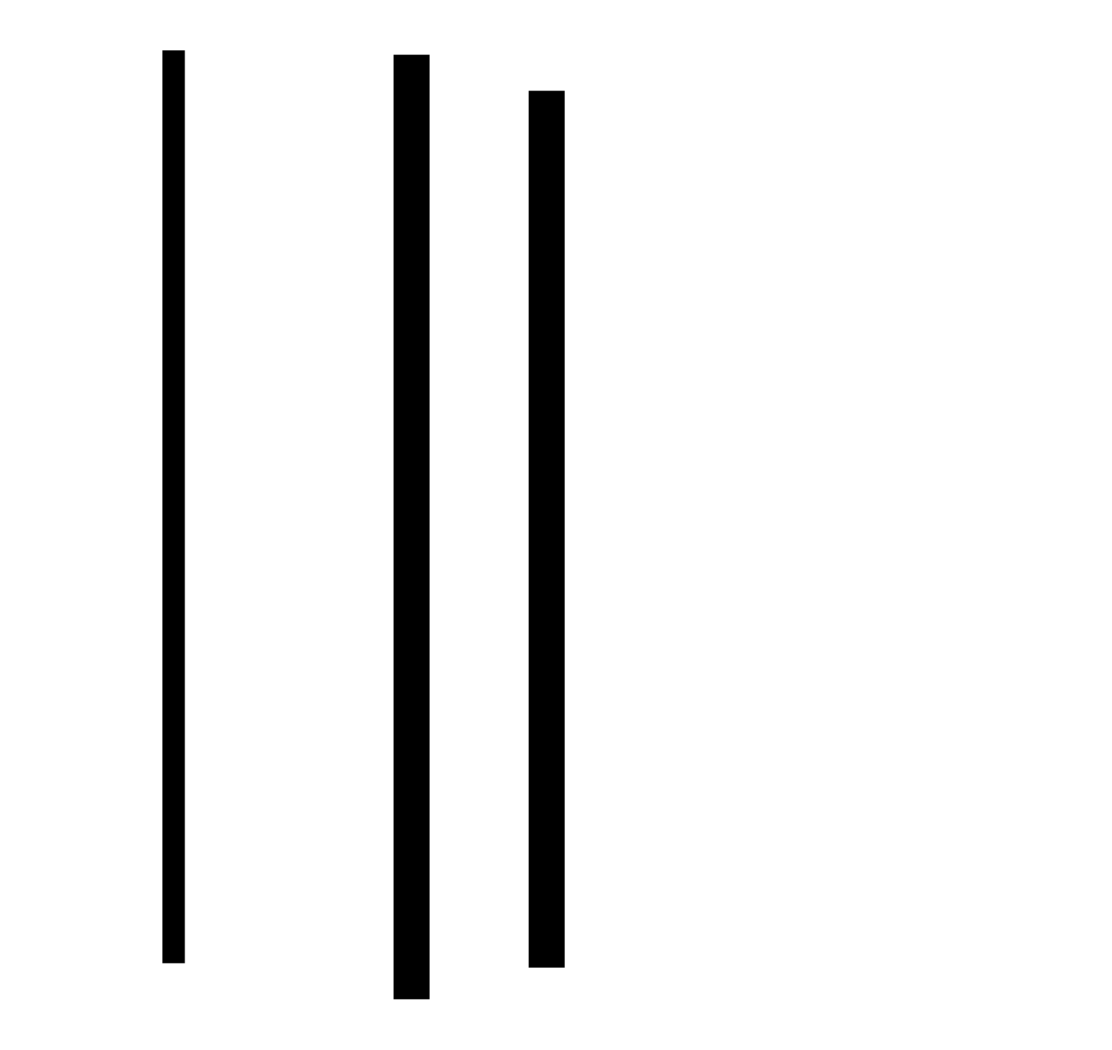} & \includegraphics[width=0.2\linewidth]{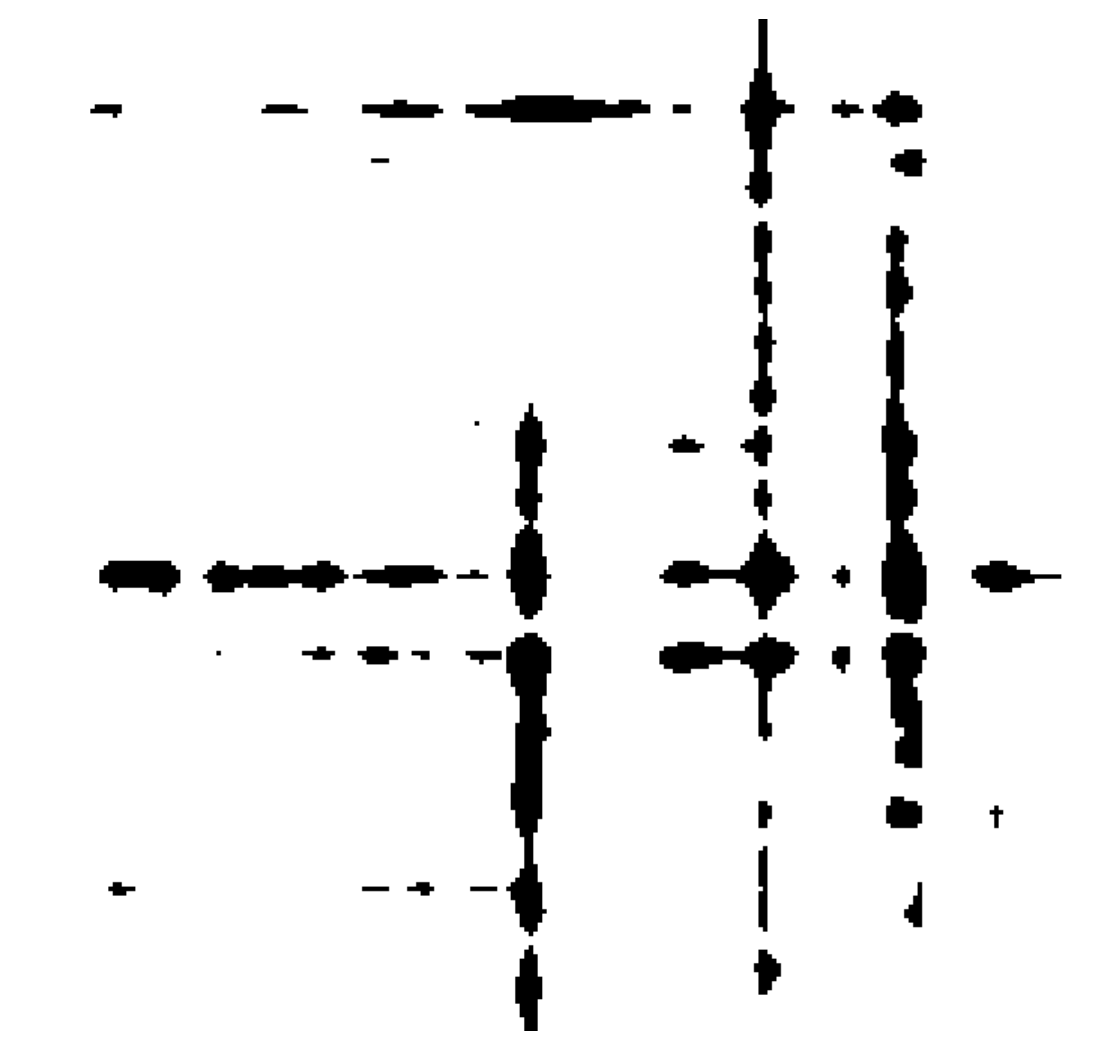} & \includegraphics[width=0.2\linewidth]{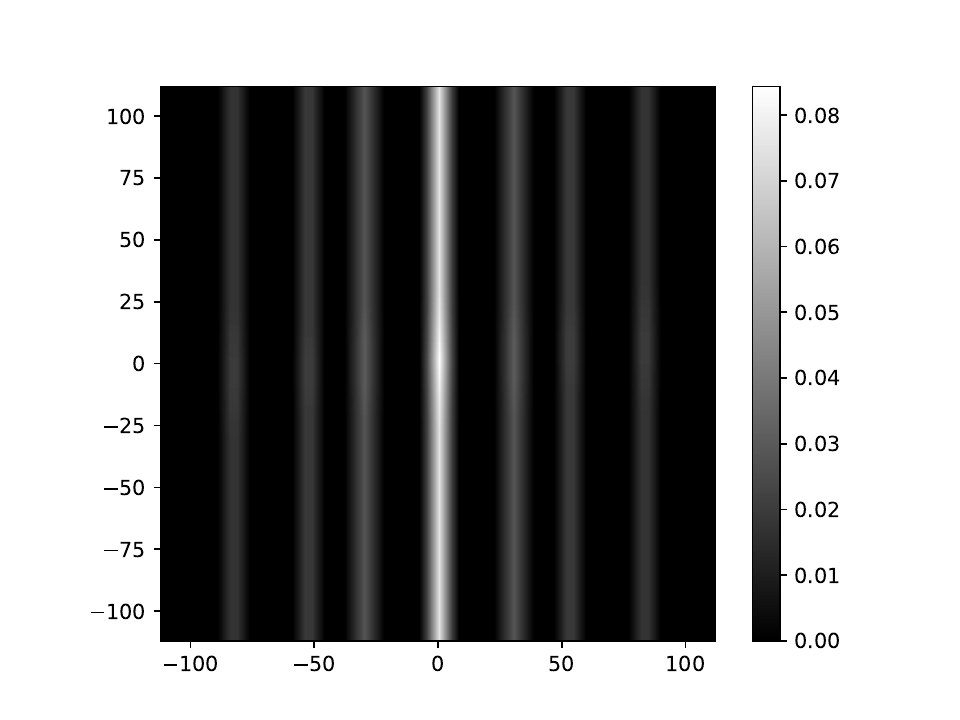} & \includegraphics[width=0.2\linewidth]{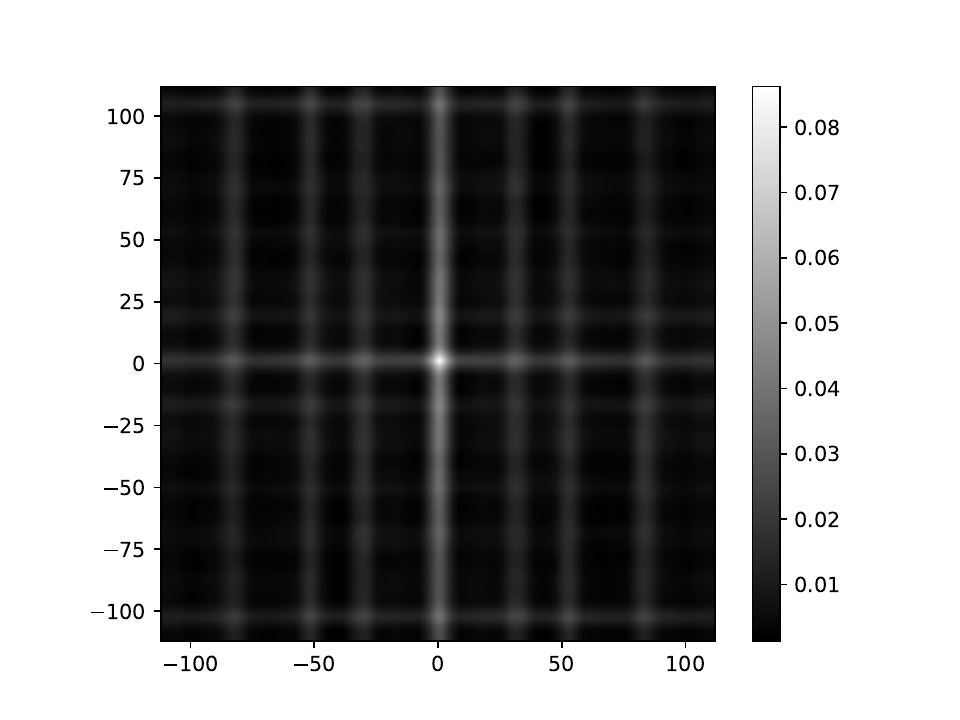} \\
\includegraphics[width=0.2\linewidth]{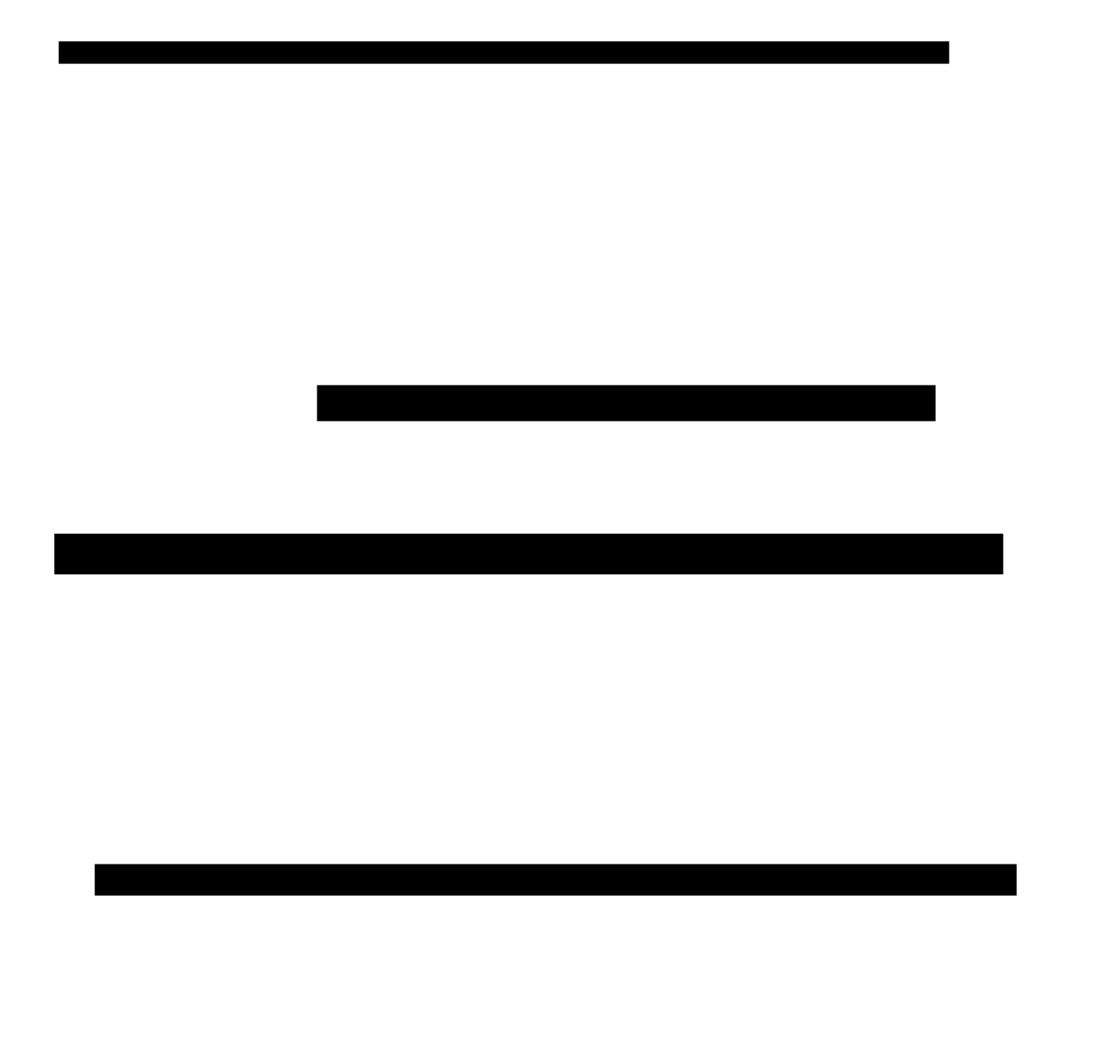} & \includegraphics[width=0.2\linewidth]{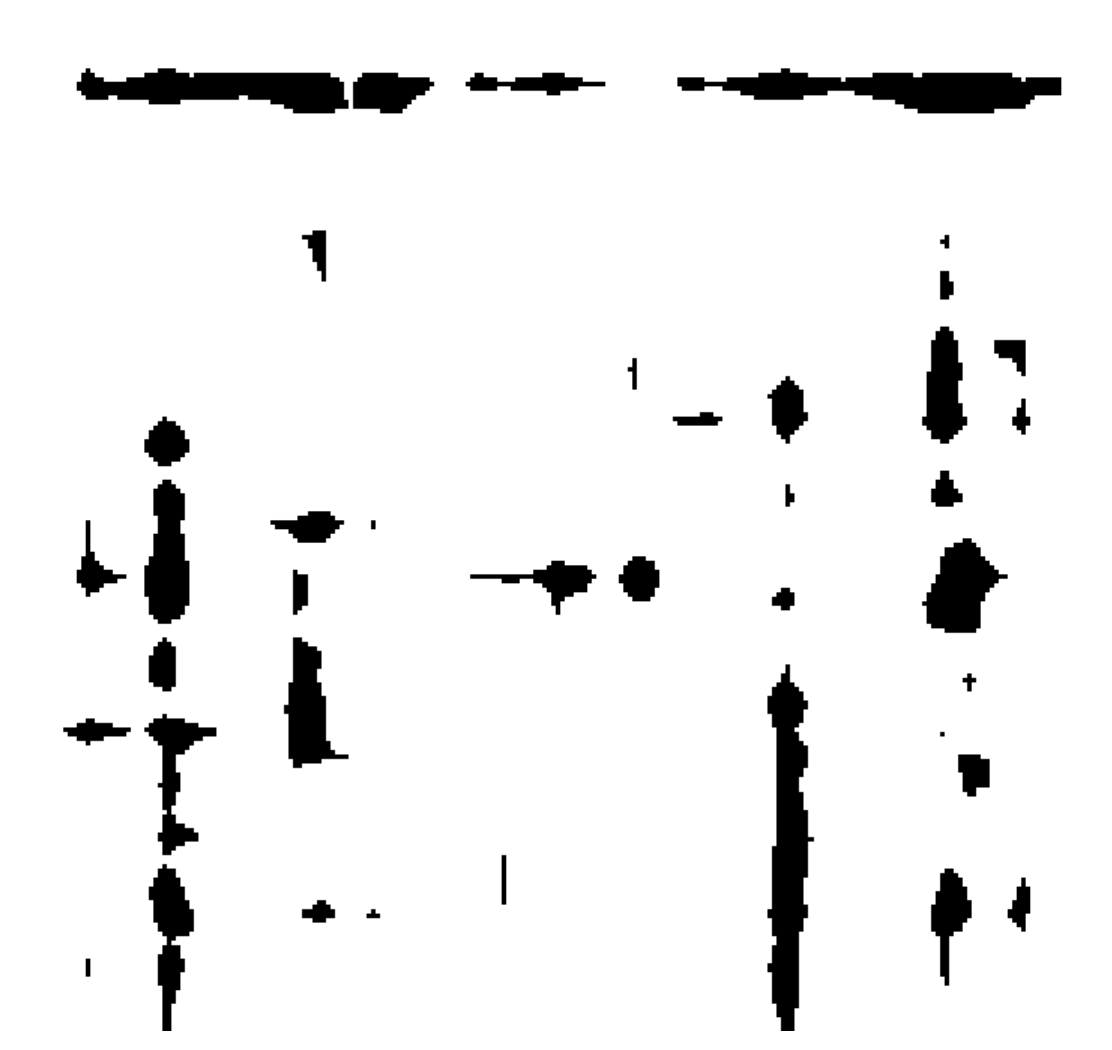} & \includegraphics[width=0.2\linewidth]{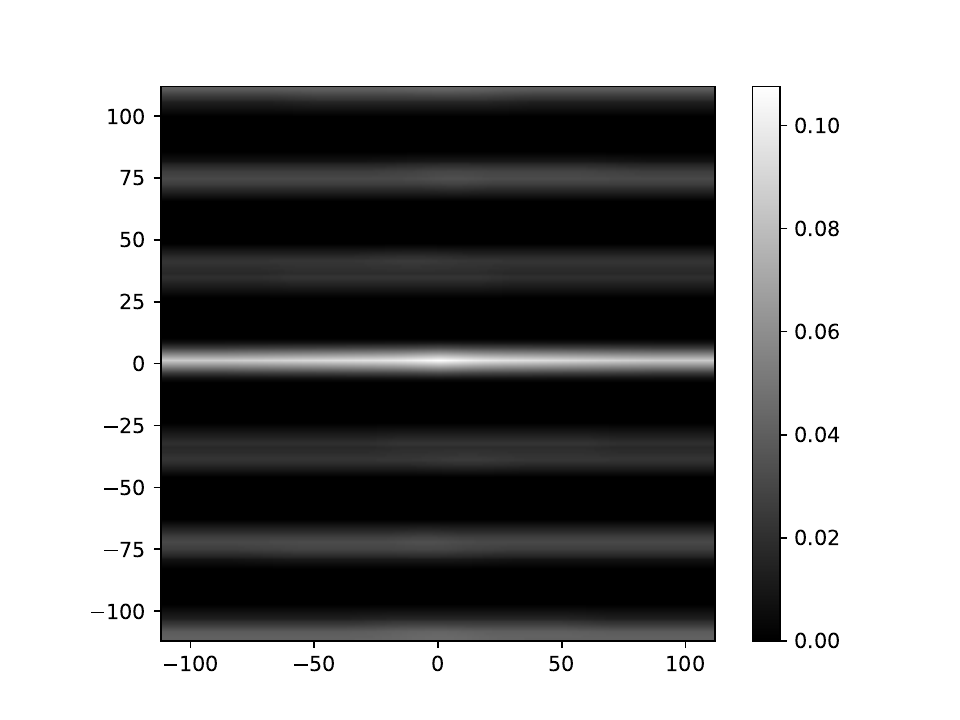} & \includegraphics[width=0.2\linewidth]{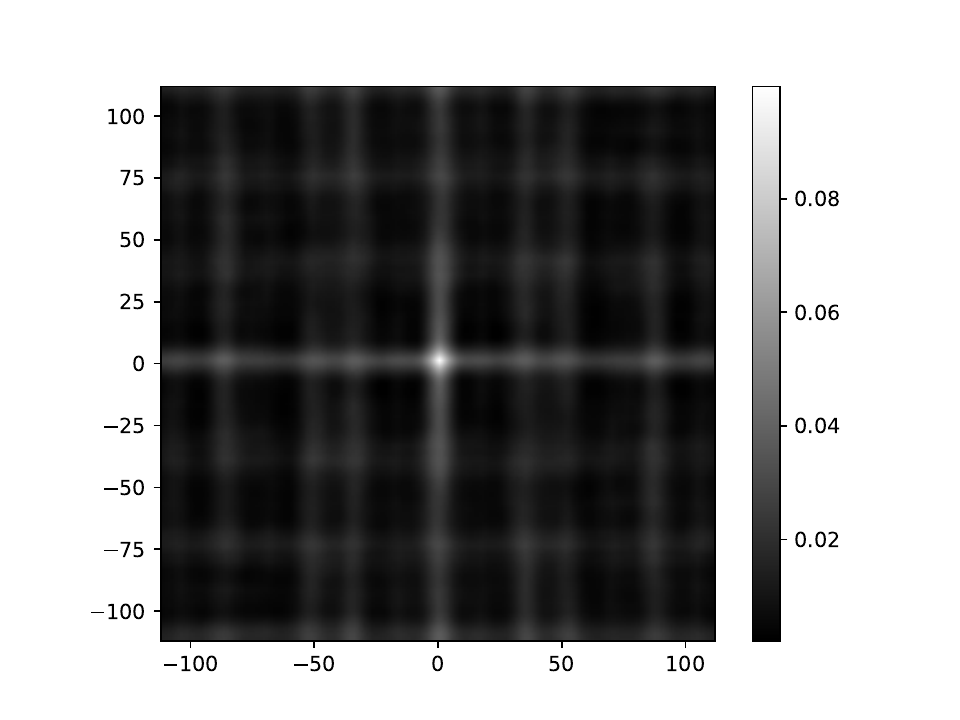} \\
\hline
\end{tabular}
\end{center}
\end{table}

\subsection{Reconstructions along with most similar images from the training set}
The following table contains illustrative instances of reconstructions, and the most similar images to them. The table shows the original image from the dataset, and the reconstructed image by our model in the two columns on the left. The third columns shows the most similar image to the reconstructed image from the training set based on MSE of the reconstruction and the sample from the training set. The fourth column shows the most similar image to the reconstructed image from the training set based on MSE of the spatial statistics of the reconstruction and the sample from the training set.
\begin{table}[H]
\caption{Original, reconstructed, and spatial statistics difference}
\begin{center}
\begin{tabular}{cccc}
\hline
\multicolumn{2}{c}{\bf Data} & \multicolumn{2}{c}{\bf Most similar images} \\
\hline
{\bf Original} & {\bf Reconstructed} & {\bf Based on image MSE} & {\bf Based on auto-corr. MSE} \\
\hline
\includegraphics[width=0.2\linewidth]{original_image_18.pdf} & \includegraphics[width=0.2\linewidth]{reconstructed_image_18.pdf} & \includegraphics[width=0.2\linewidth]{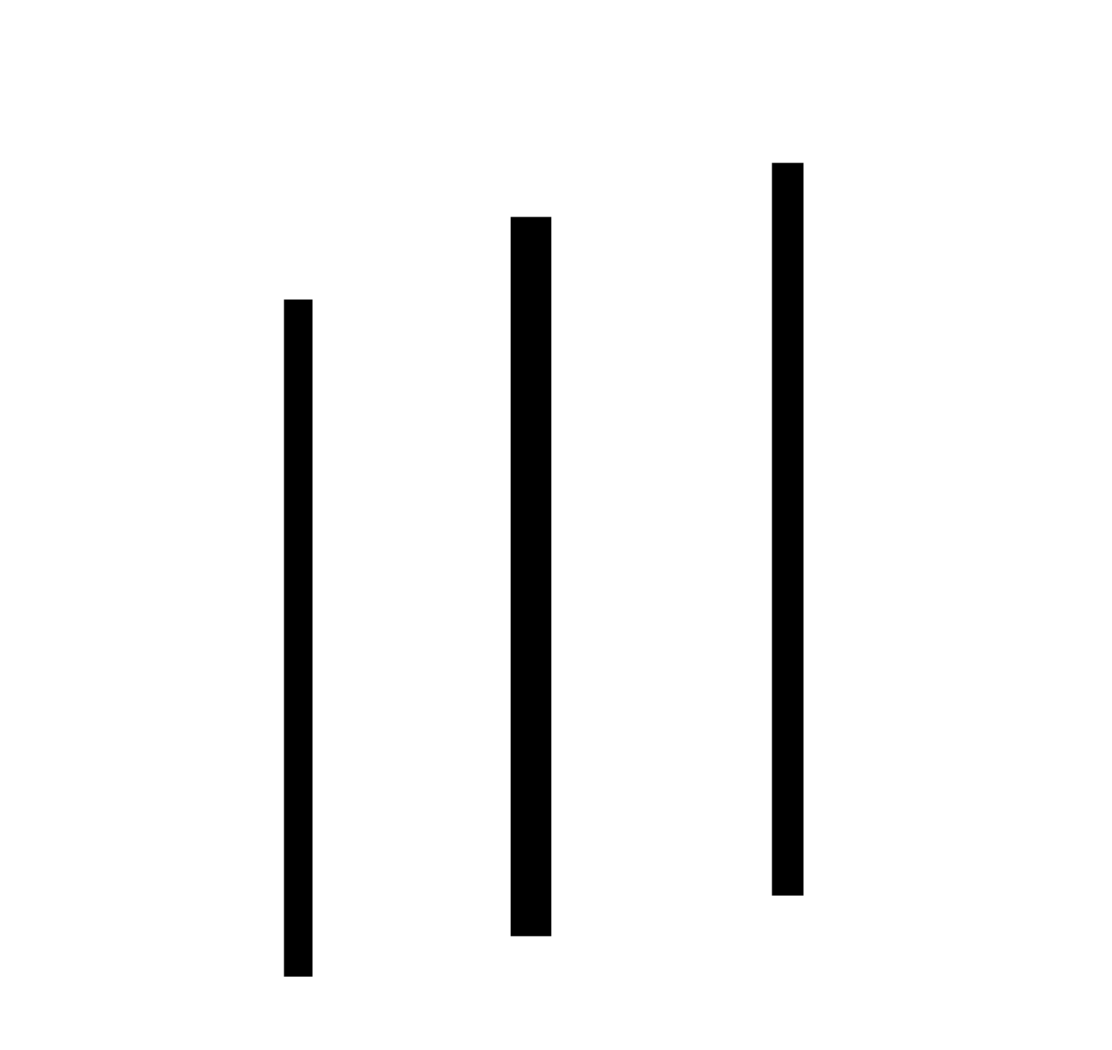} & \includegraphics[width=0.2\linewidth]{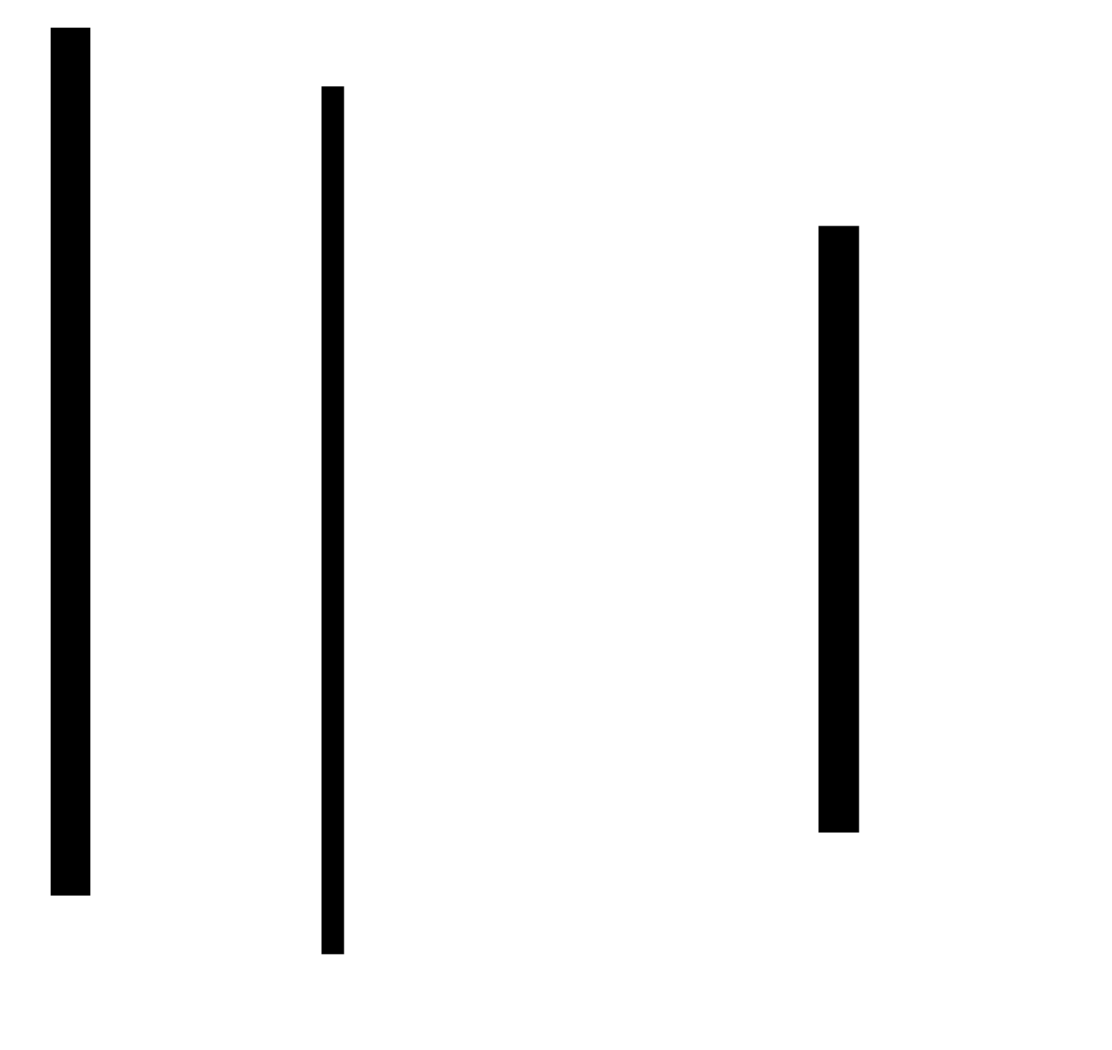} \\
\includegraphics[width=0.2\linewidth]{original_image_14.pdf} & \includegraphics[width=0.2\linewidth]{reconstructed_image_14.pdf} & \includegraphics[width=0.2\linewidth]{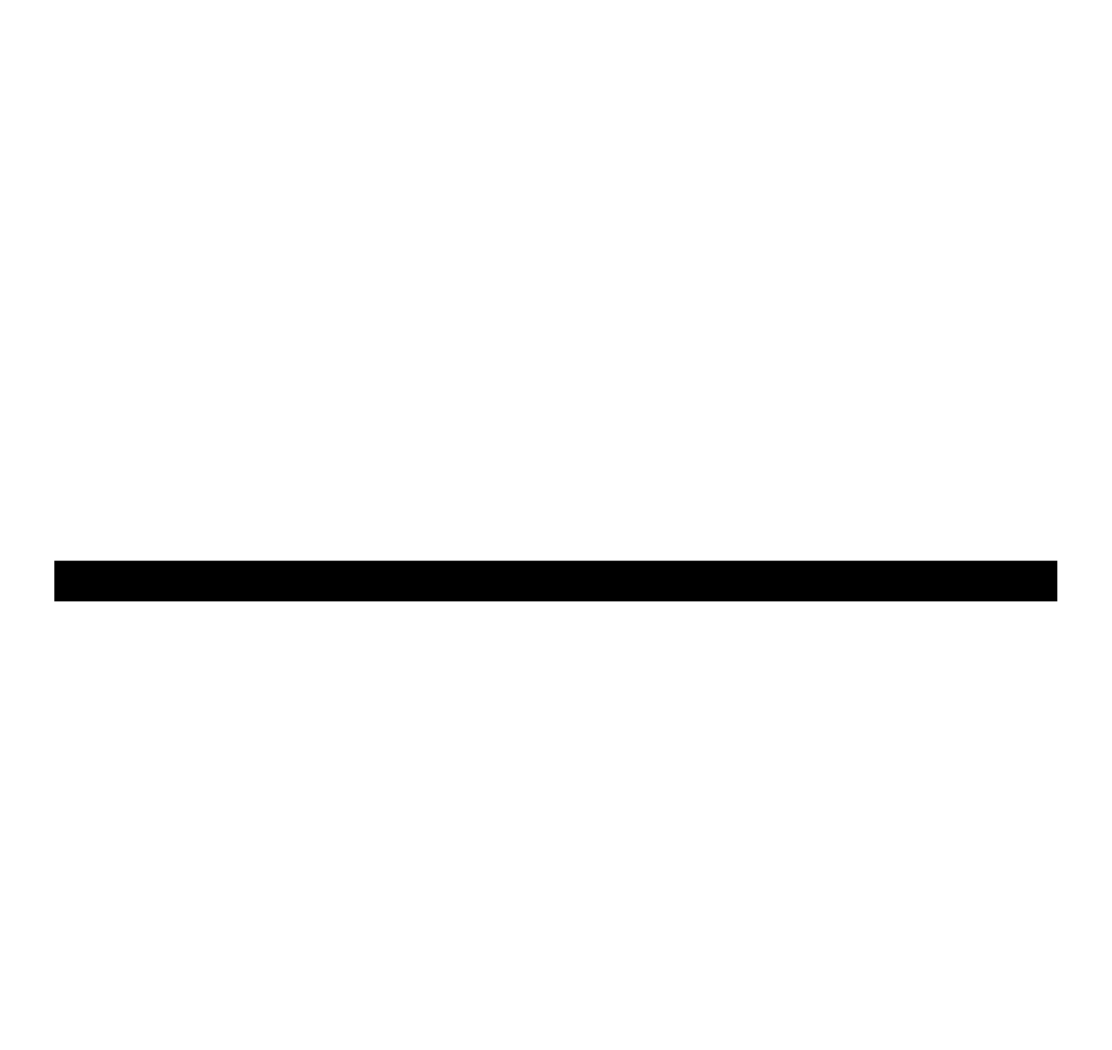} & \includegraphics[width=0.2\linewidth]{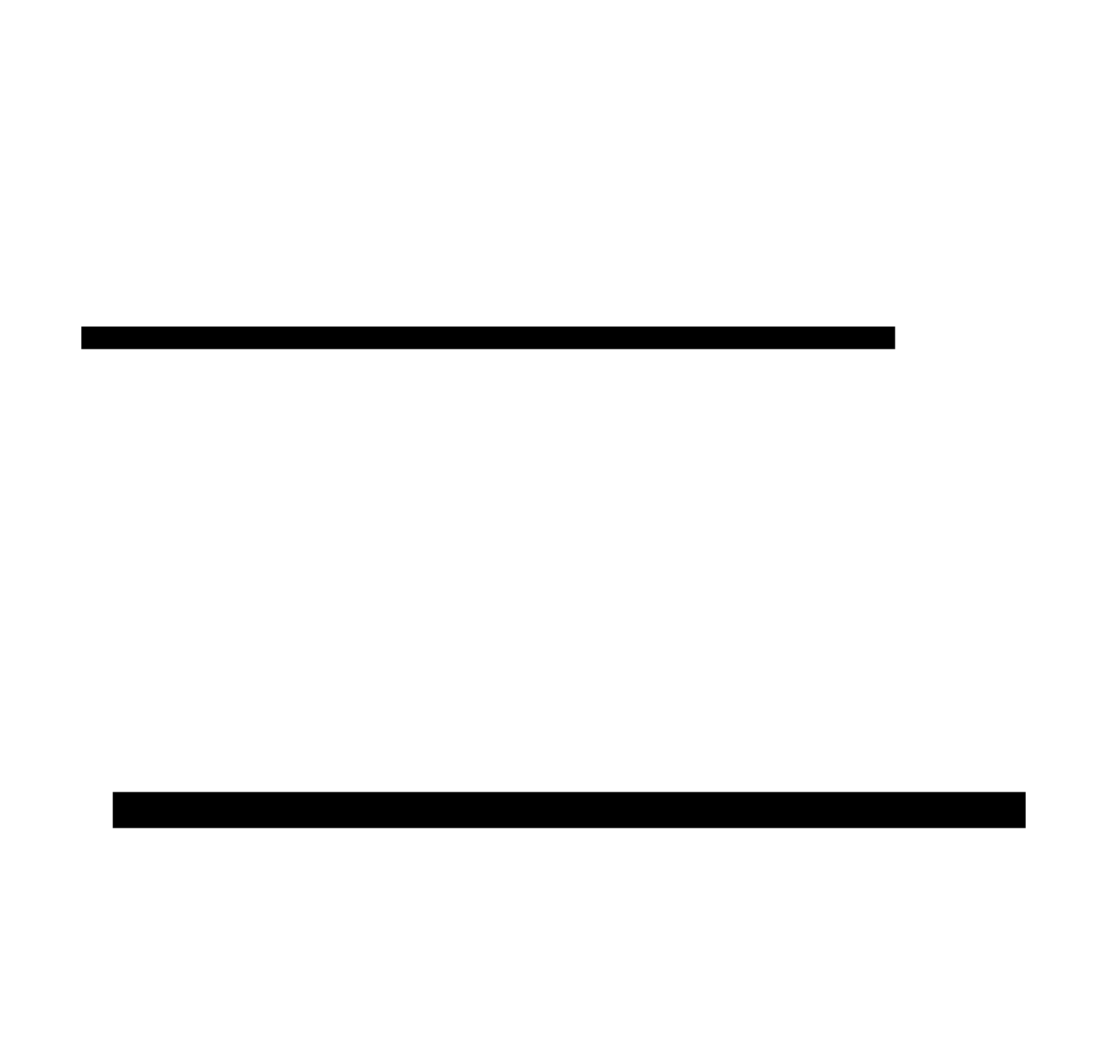} \\
\includegraphics[width=0.2\linewidth]{original_image_16.pdf} & \includegraphics[width=0.2\linewidth]{reconstructed_image_16.pdf} & \includegraphics[width=0.2\linewidth]{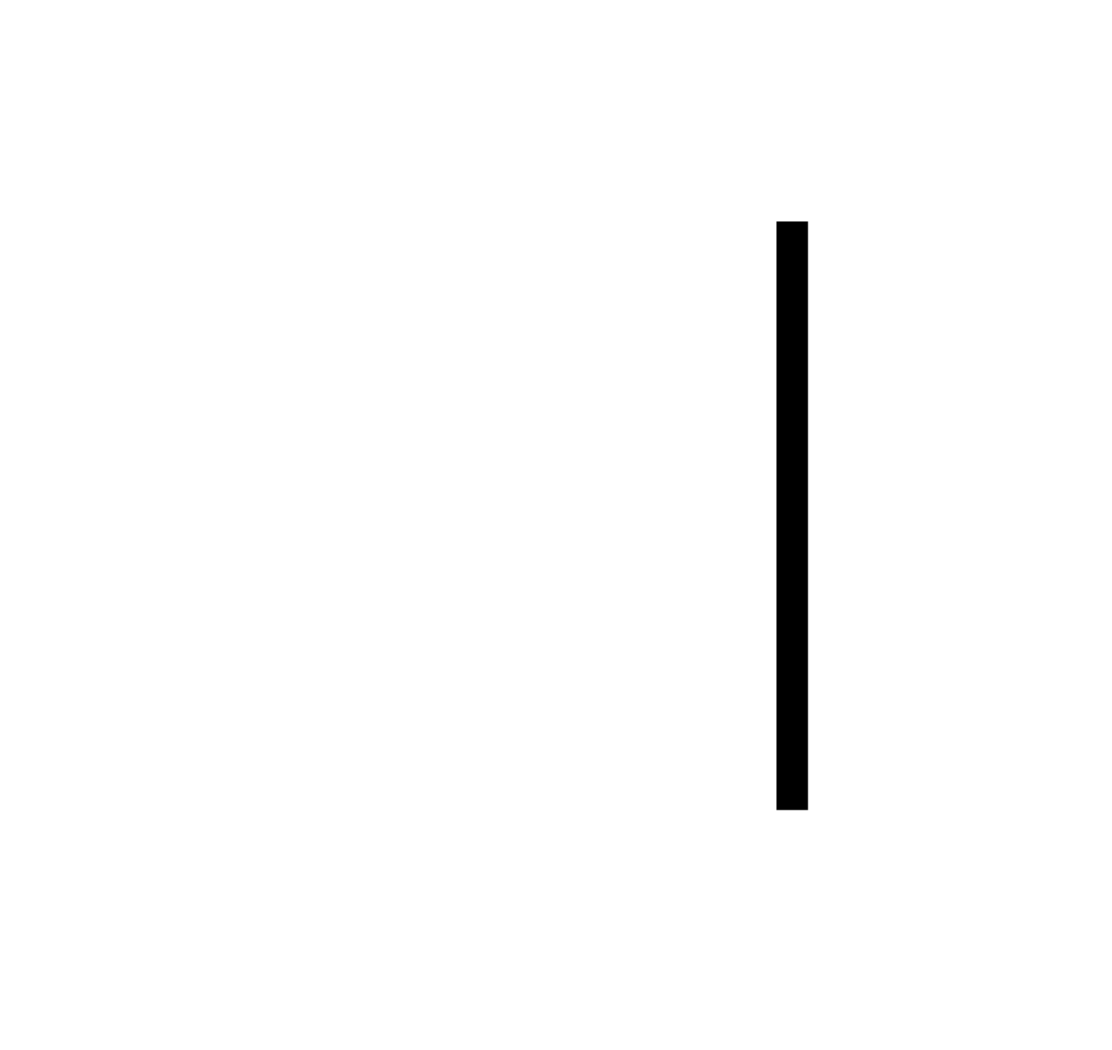} & \includegraphics[width=0.2\linewidth]{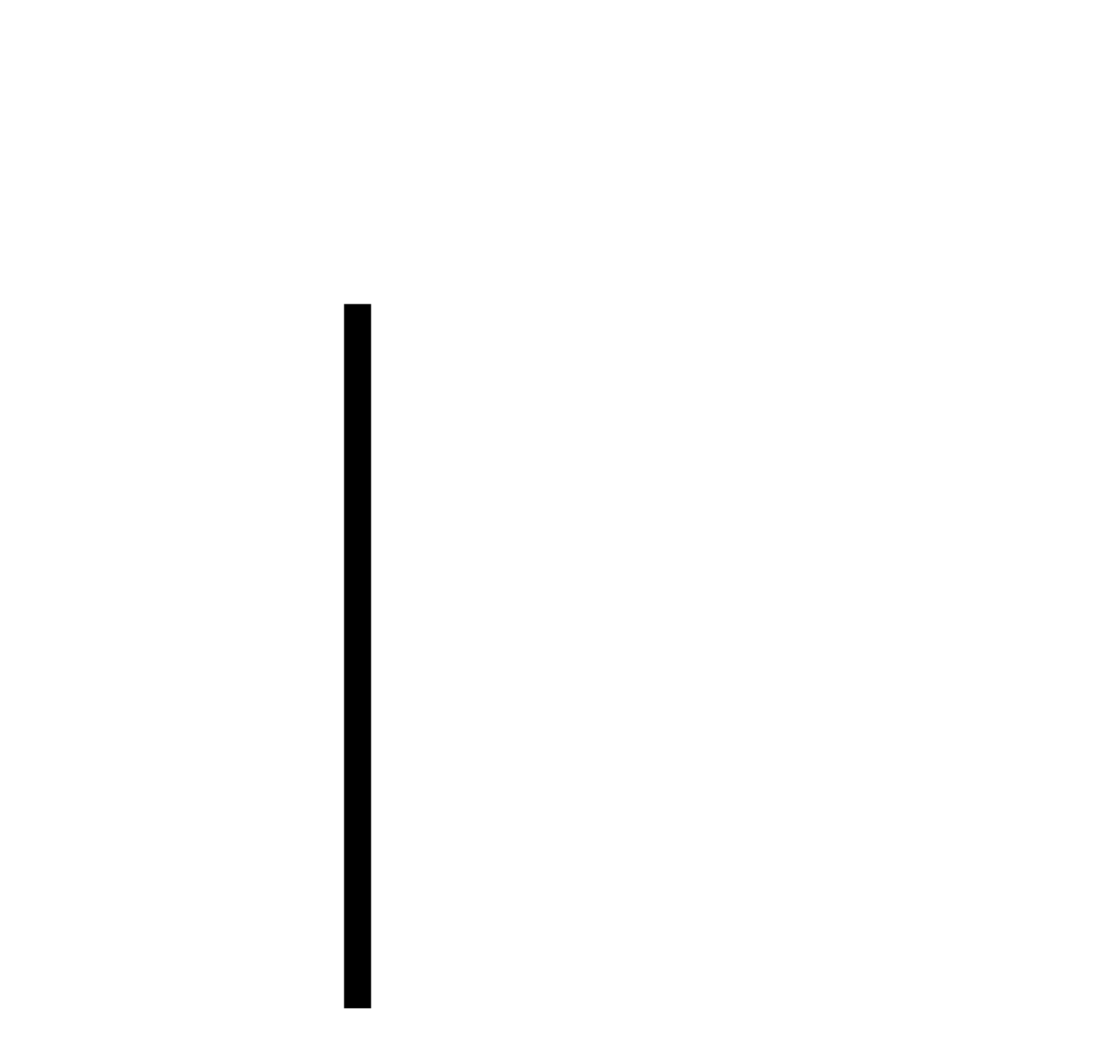} \\
\includegraphics[width=0.2\linewidth]{original_image_9.pdf} & \includegraphics[width=0.2\linewidth]{reconstructed_image_9.pdf} & \includegraphics[width=0.2\linewidth]{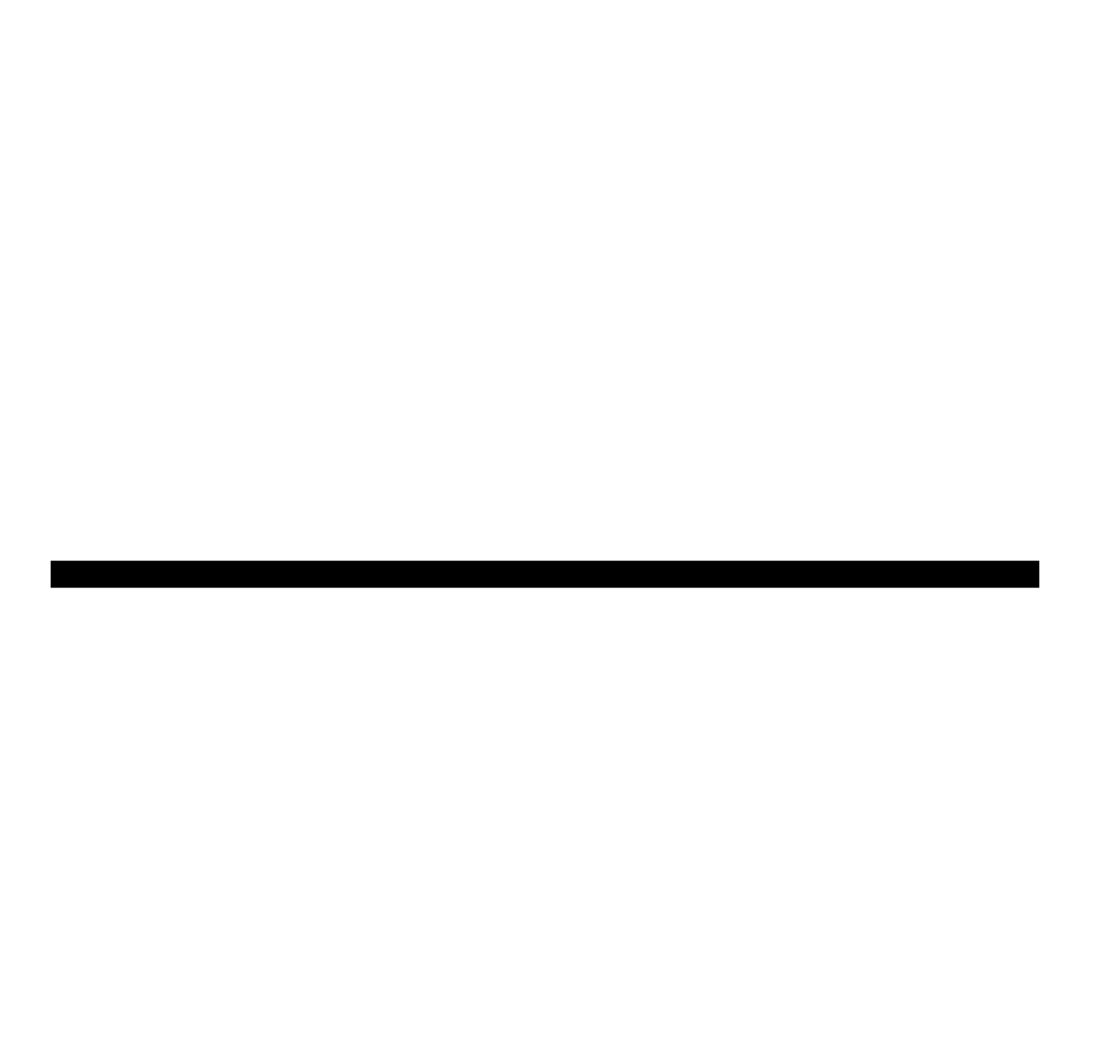} & \includegraphics[width=0.2\linewidth]{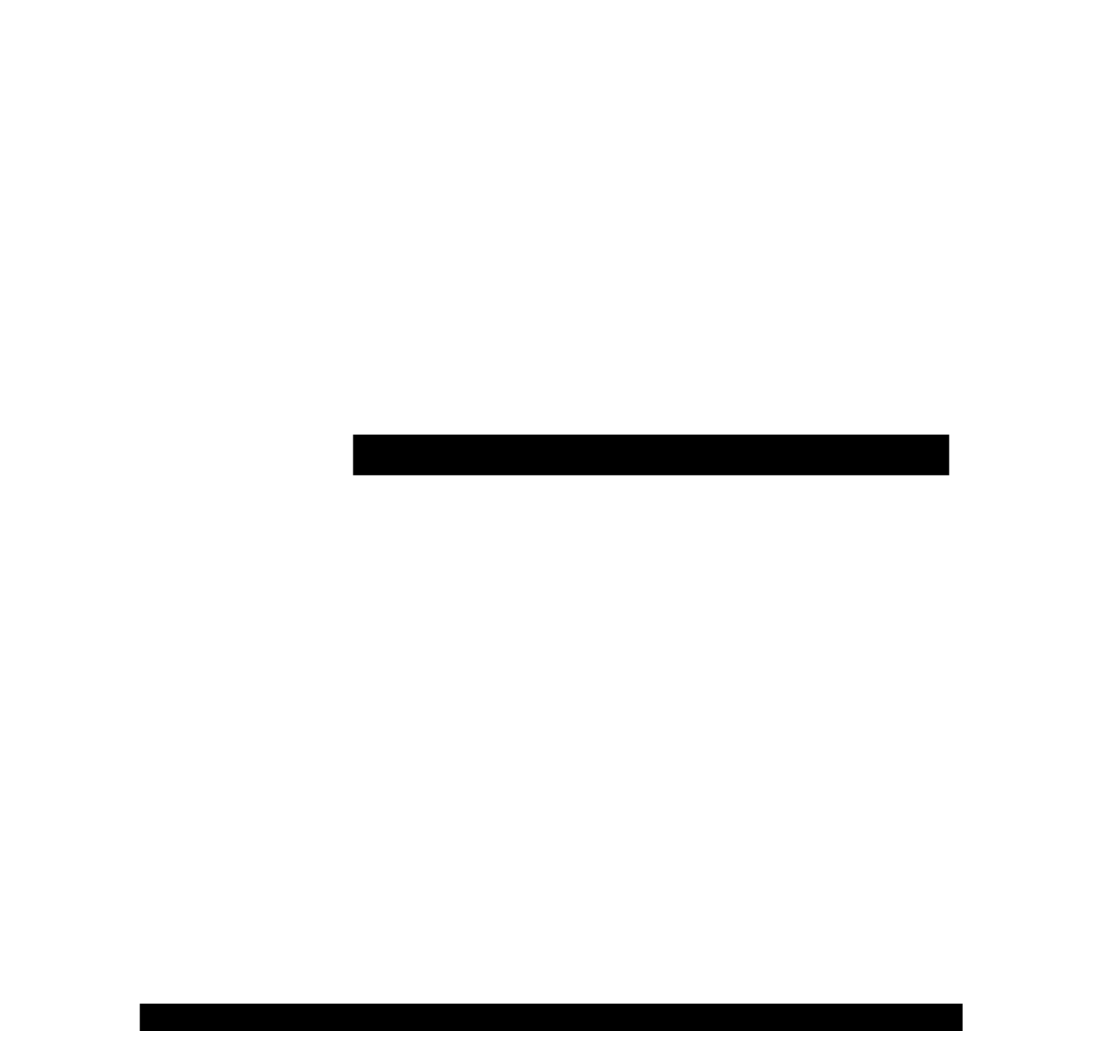} \\
\includegraphics[width=0.2\linewidth]{original_image_12.pdf} & \includegraphics[width=0.2\linewidth]{reconstructed_image_12.pdf} & \includegraphics[width=0.2\linewidth]{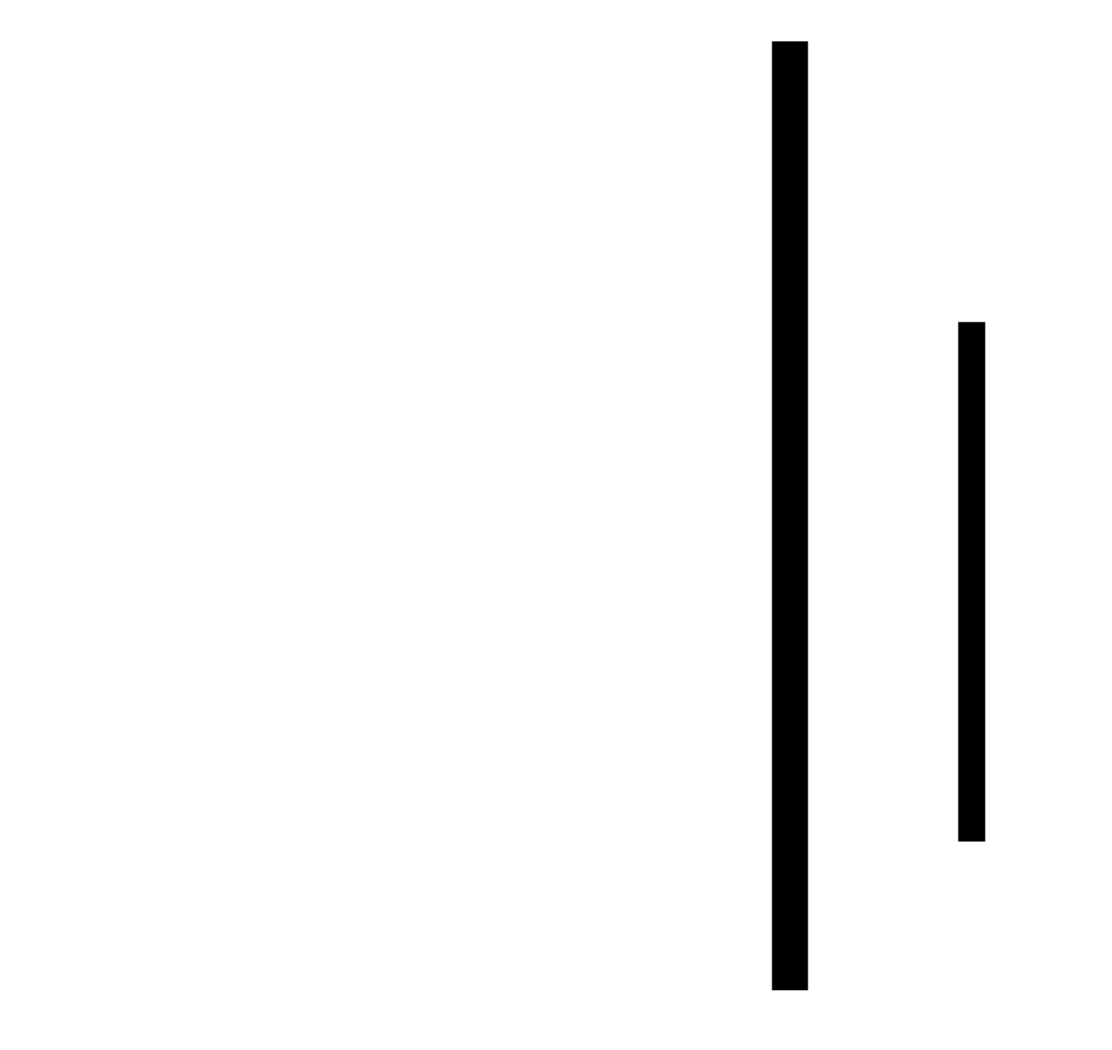} & \includegraphics[width=0.2\linewidth]{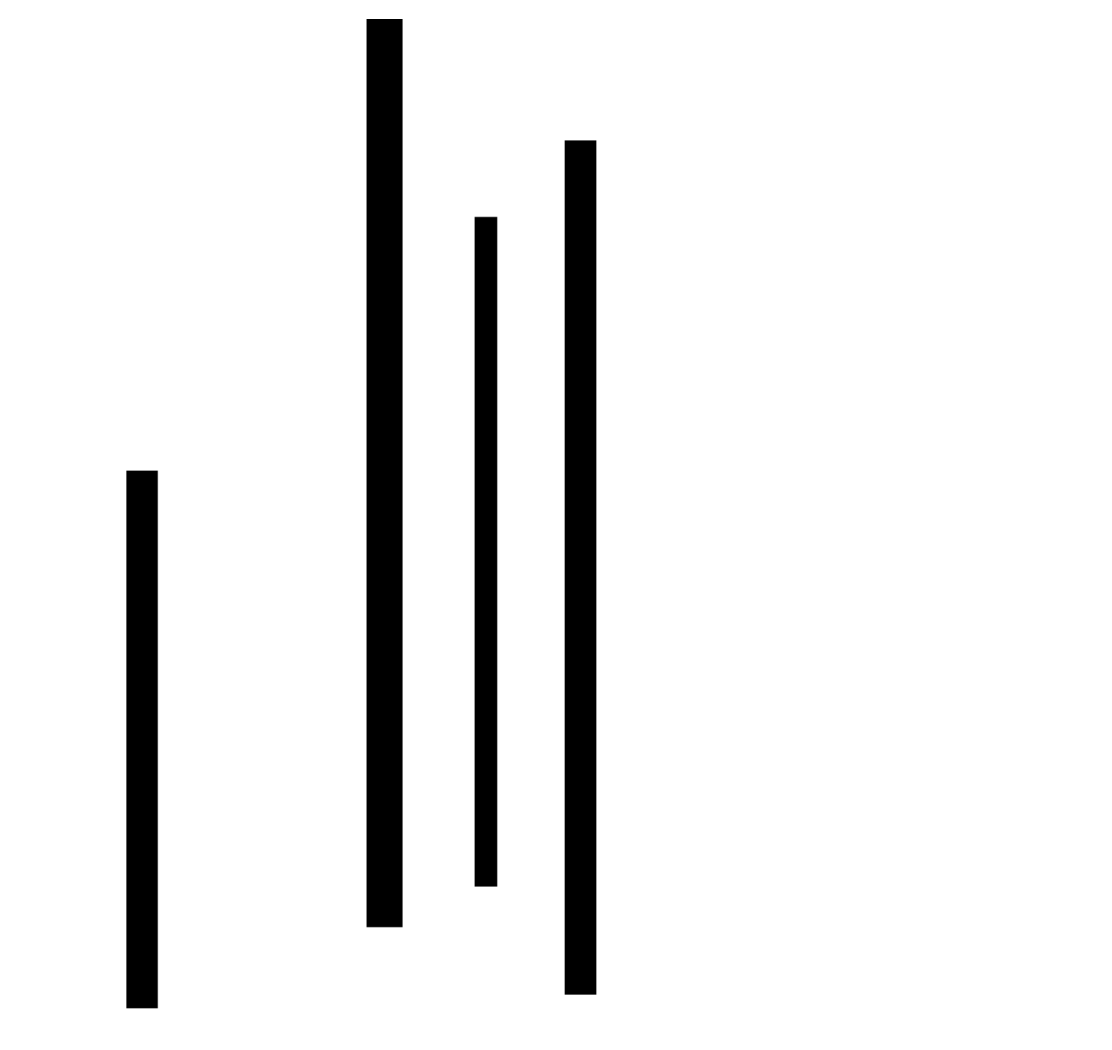} \\
\includegraphics[width=0.2\linewidth]{original_image_5.pdf} & \includegraphics[width=0.2\linewidth]{reconstructed_image_5.pdf} & \includegraphics[width=0.2\linewidth]{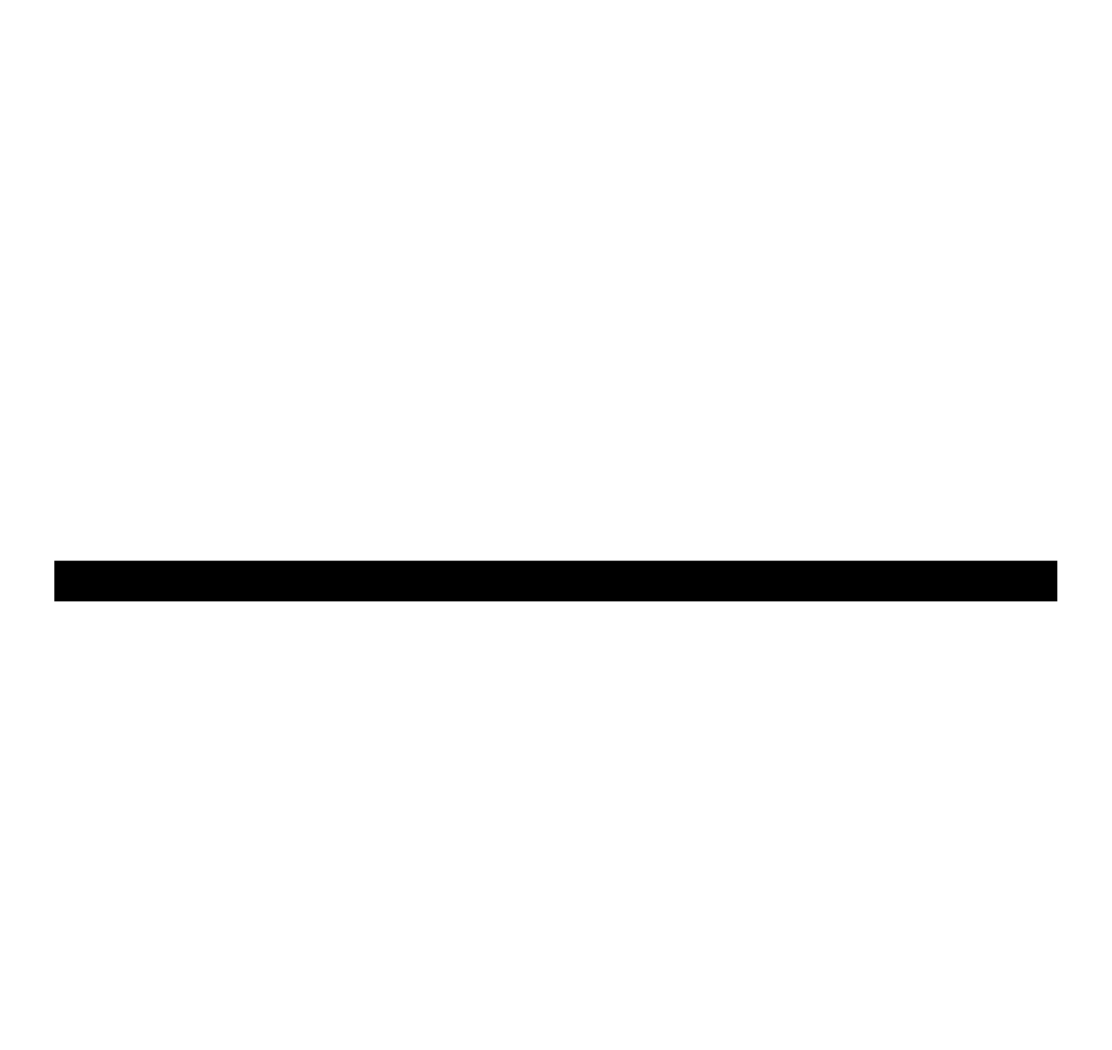} & \includegraphics[width=0.2\linewidth]{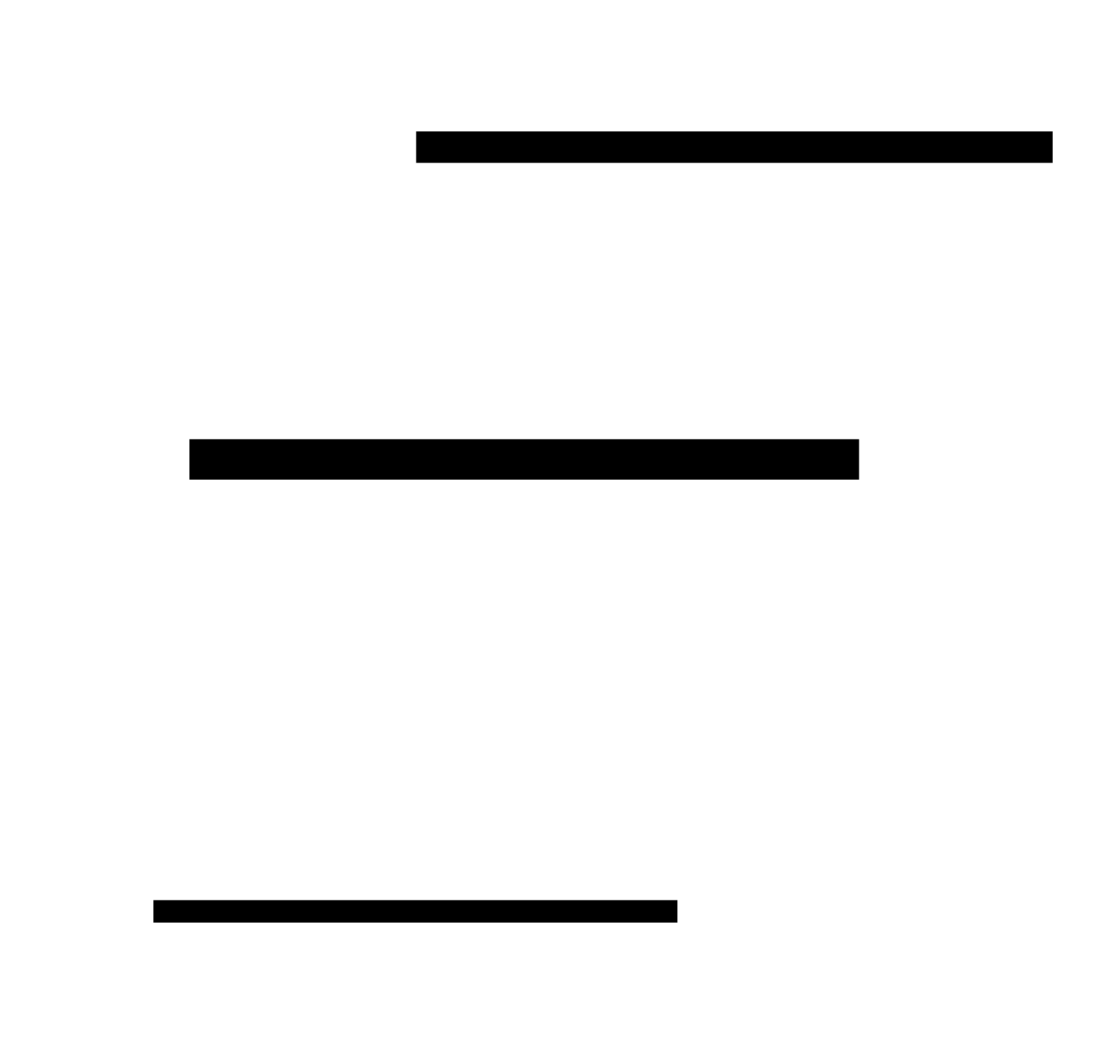} \\
\hline
\end{tabular}
\end{center}
\end{table}

\begin{table}[H]
\caption{Original, reconstructed, and spatial statistics difference}
\begin{center}
\begin{tabular}{cccc}
\hline
\multicolumn{2}{c}{\bf Data} & \multicolumn{2}{c}{\bf Most similar images} \\
\hline
{\bf Original} & {\bf Reconstructed} & {\bf Based on image MSE} & {\bf Based on auto-corr. MSE} \\
\hline
\includegraphics[width=0.2\linewidth]{original_image_43.pdf} & \includegraphics[width=0.2\linewidth]{reconstructed_image_43.pdf} & \includegraphics[width=0.2\linewidth]{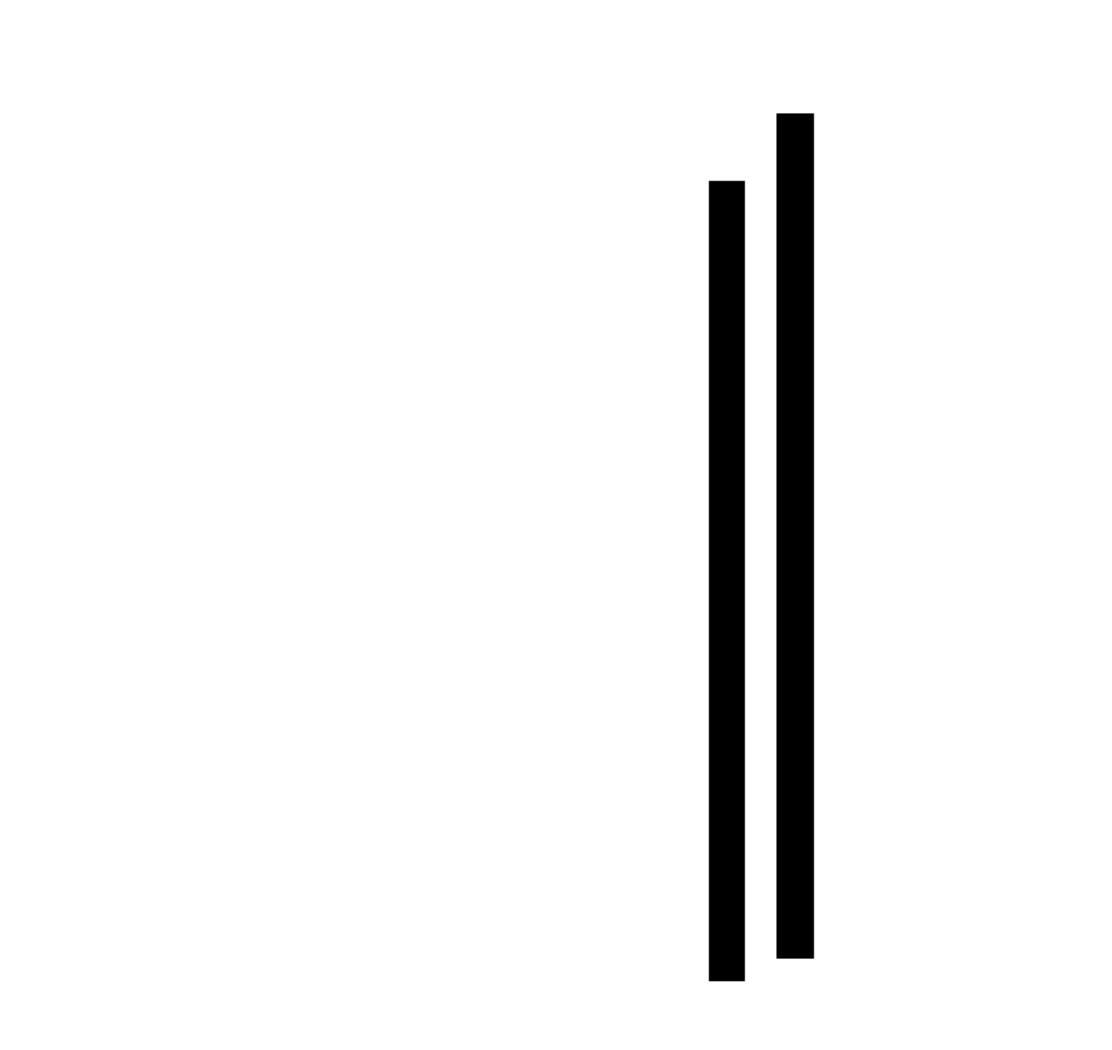} & \includegraphics[width=0.2\linewidth]{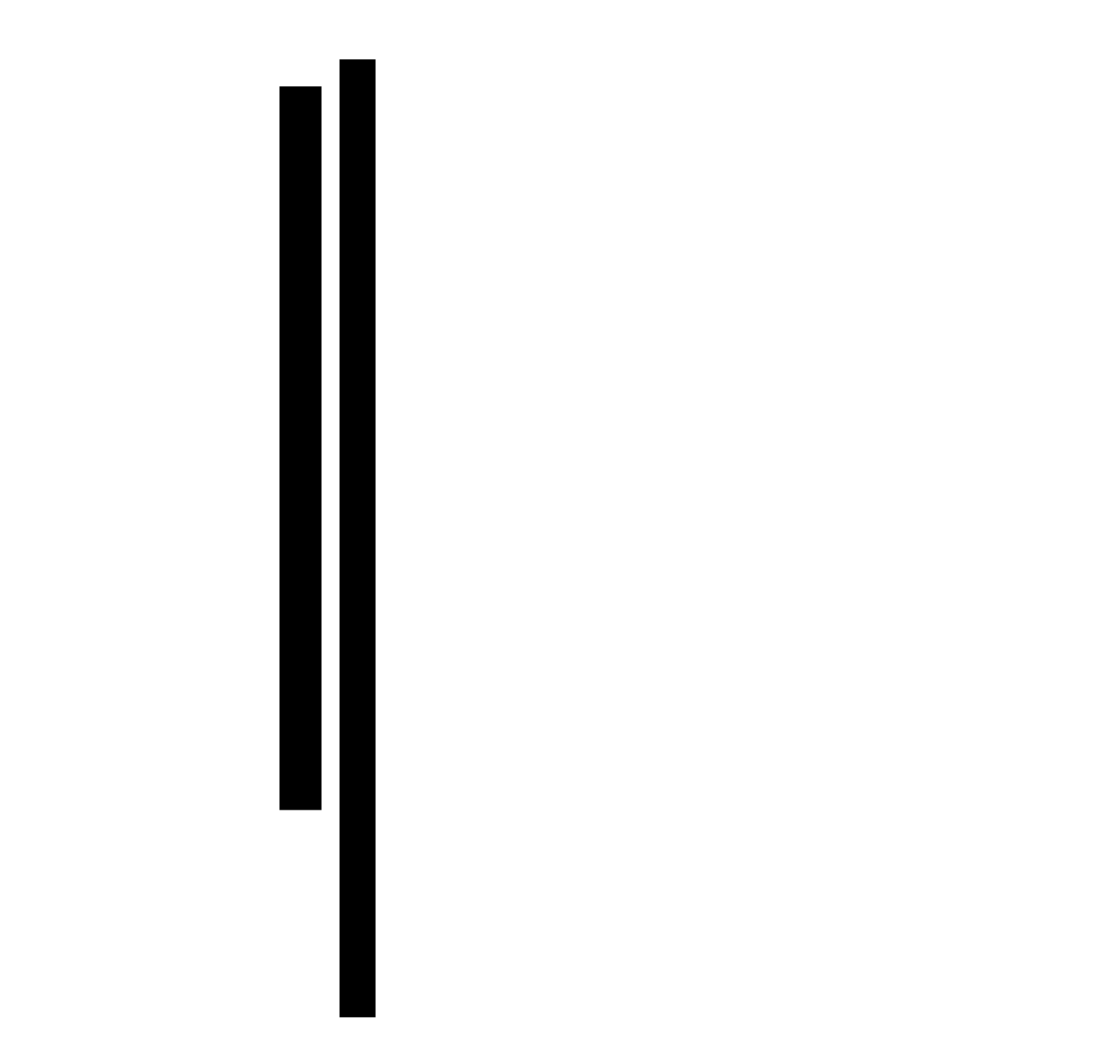} \\
\includegraphics[width=0.2\linewidth]{original_image_36.pdf} & \includegraphics[width=0.2\linewidth]{reconstructed_image_36.pdf} & \includegraphics[width=0.2\linewidth]{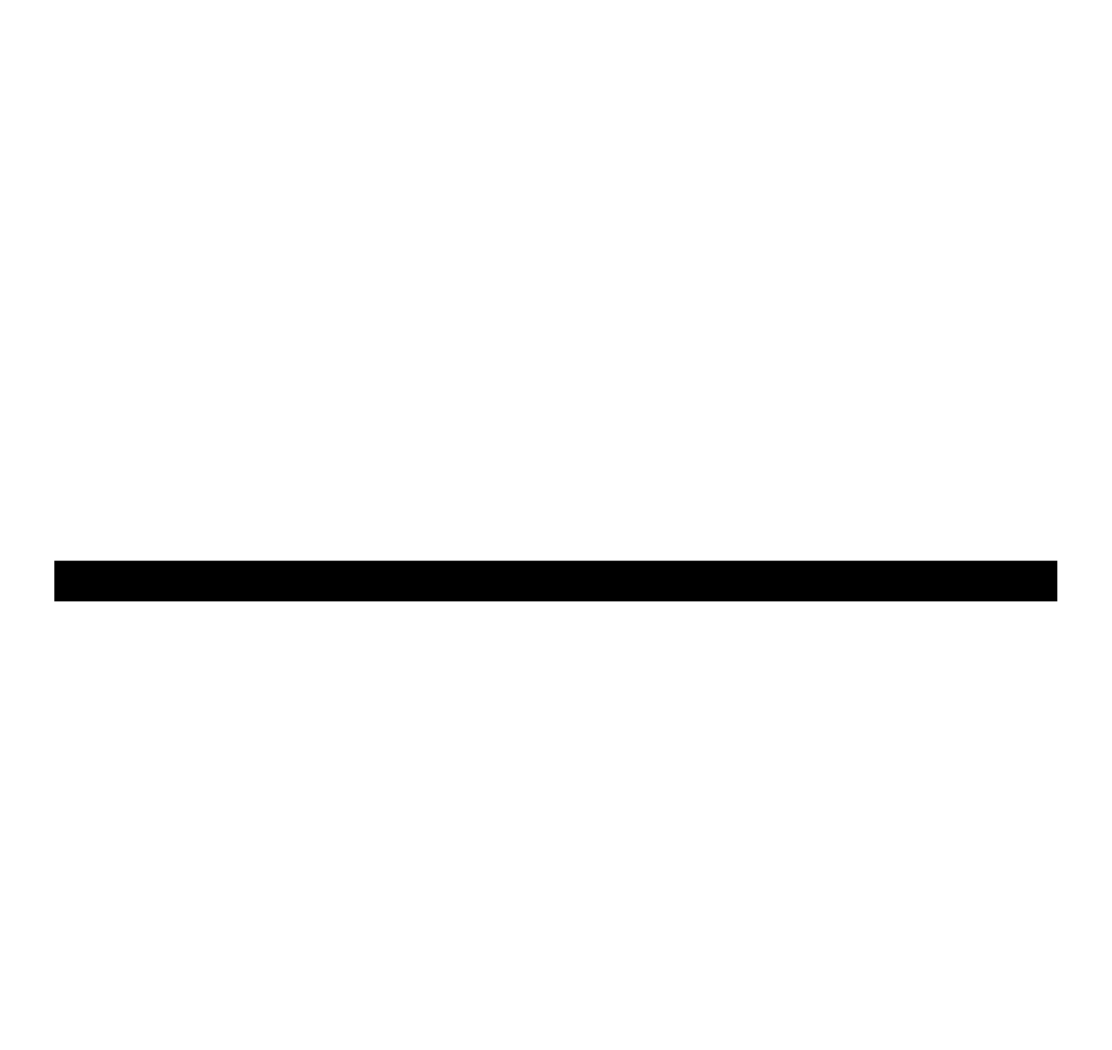} & \includegraphics[width=0.2\linewidth]{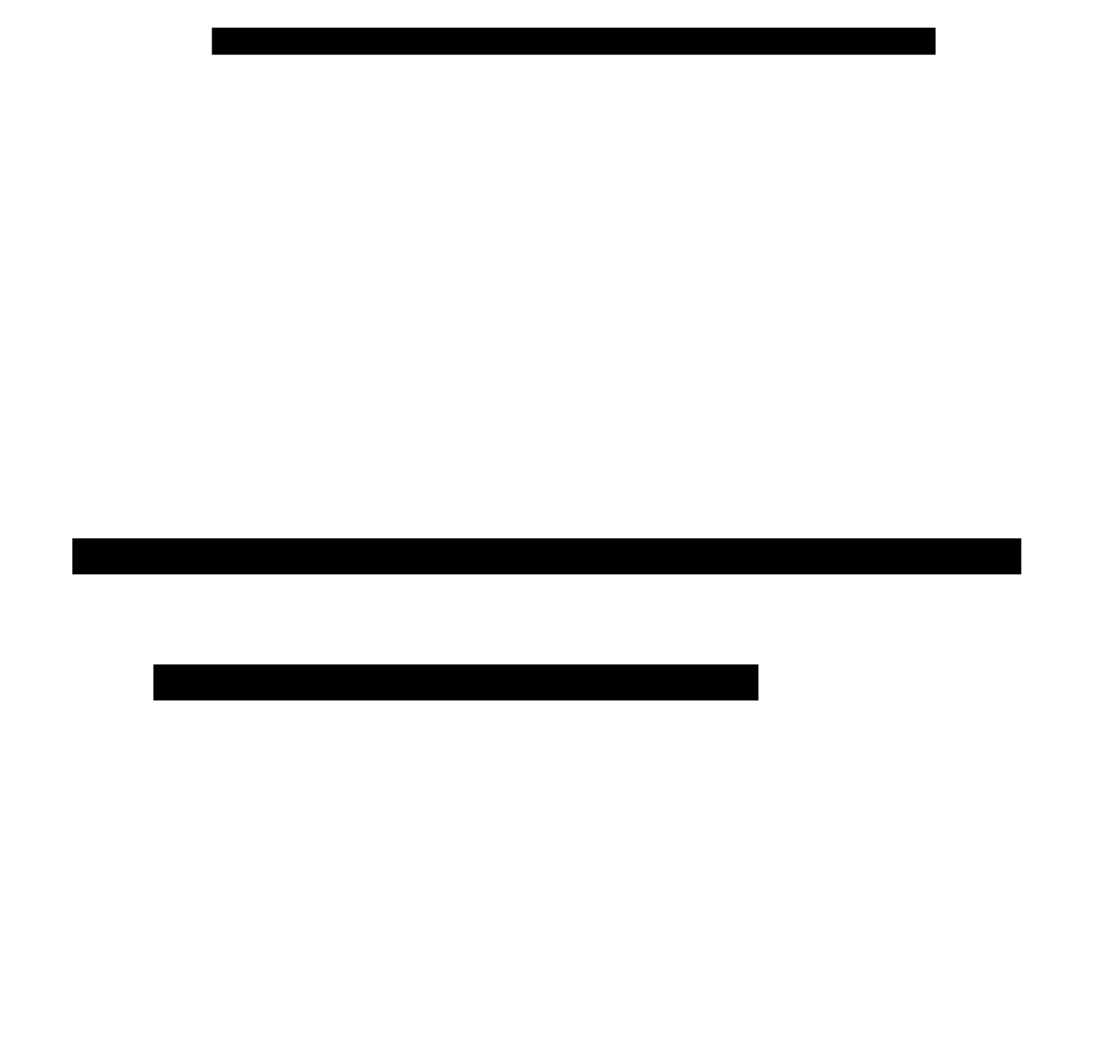} \\
\includegraphics[width=0.2\linewidth]{original_image_39.pdf} & \includegraphics[width=0.2\linewidth]{reconstructed_image_39.pdf} & \includegraphics[width=0.2\linewidth]{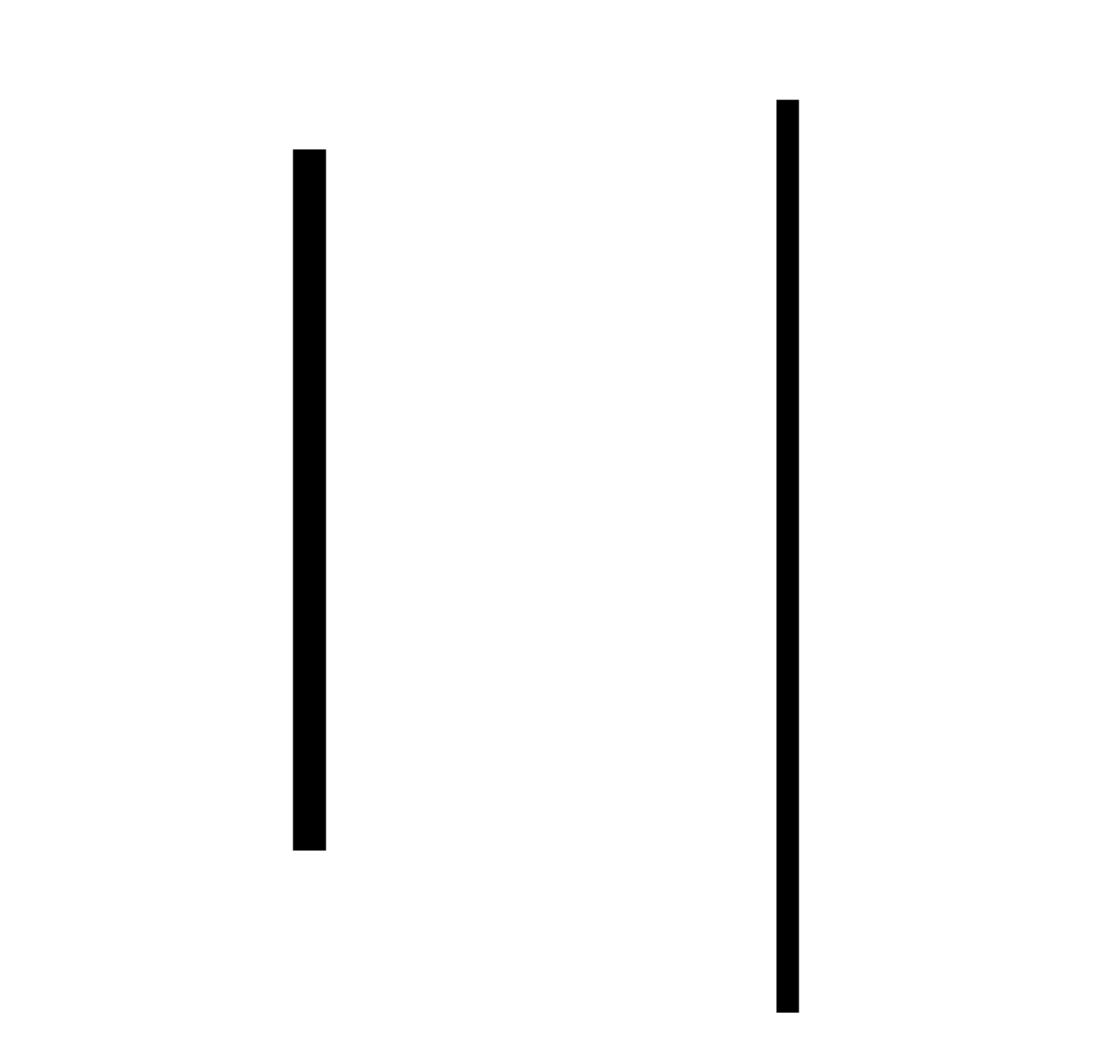} & \includegraphics[width=0.2\linewidth]{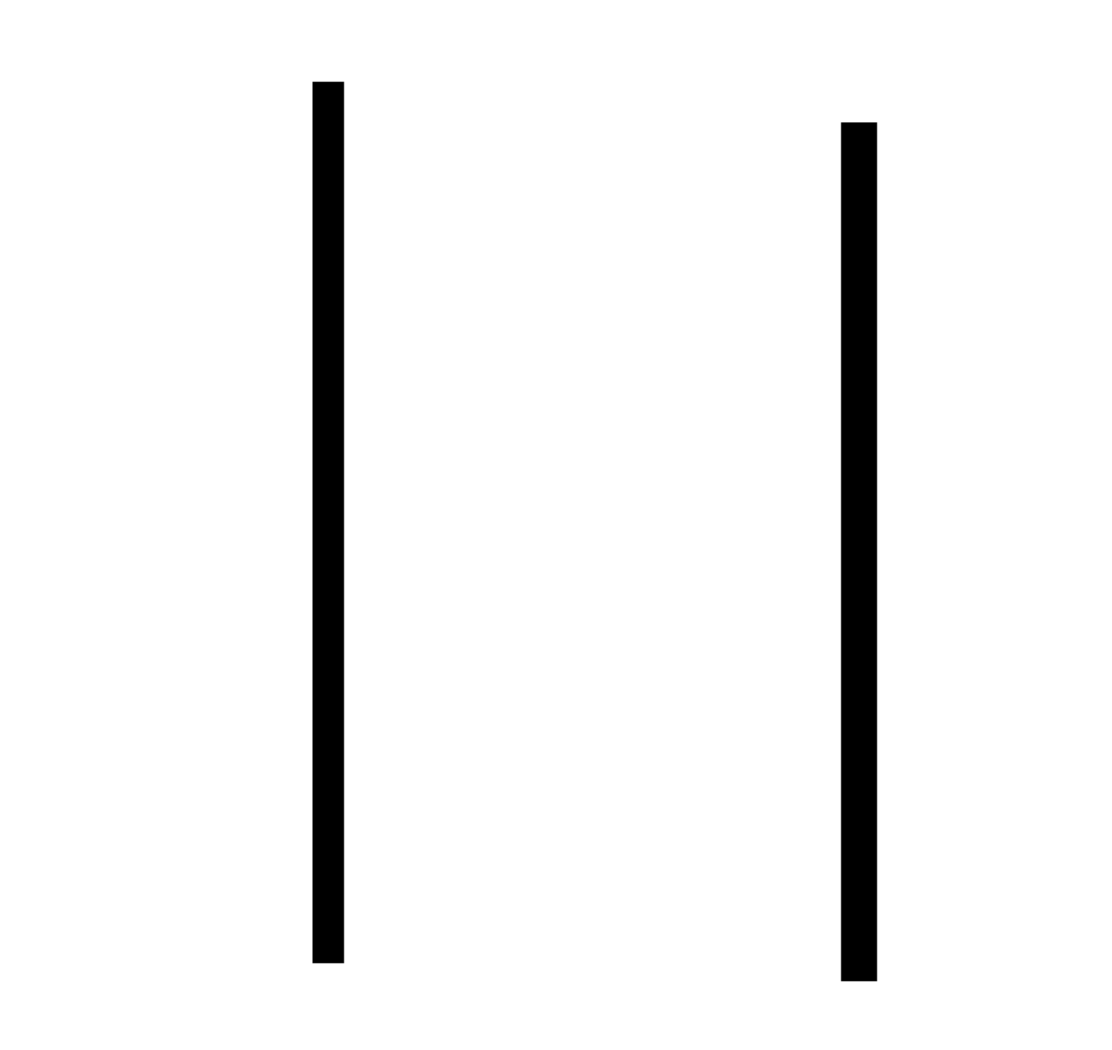} \\
\includegraphics[width=0.2\linewidth]{original_image_31.pdf} & \includegraphics[width=0.2\linewidth]{reconstructed_image_31.pdf} & \includegraphics[width=0.2\linewidth]{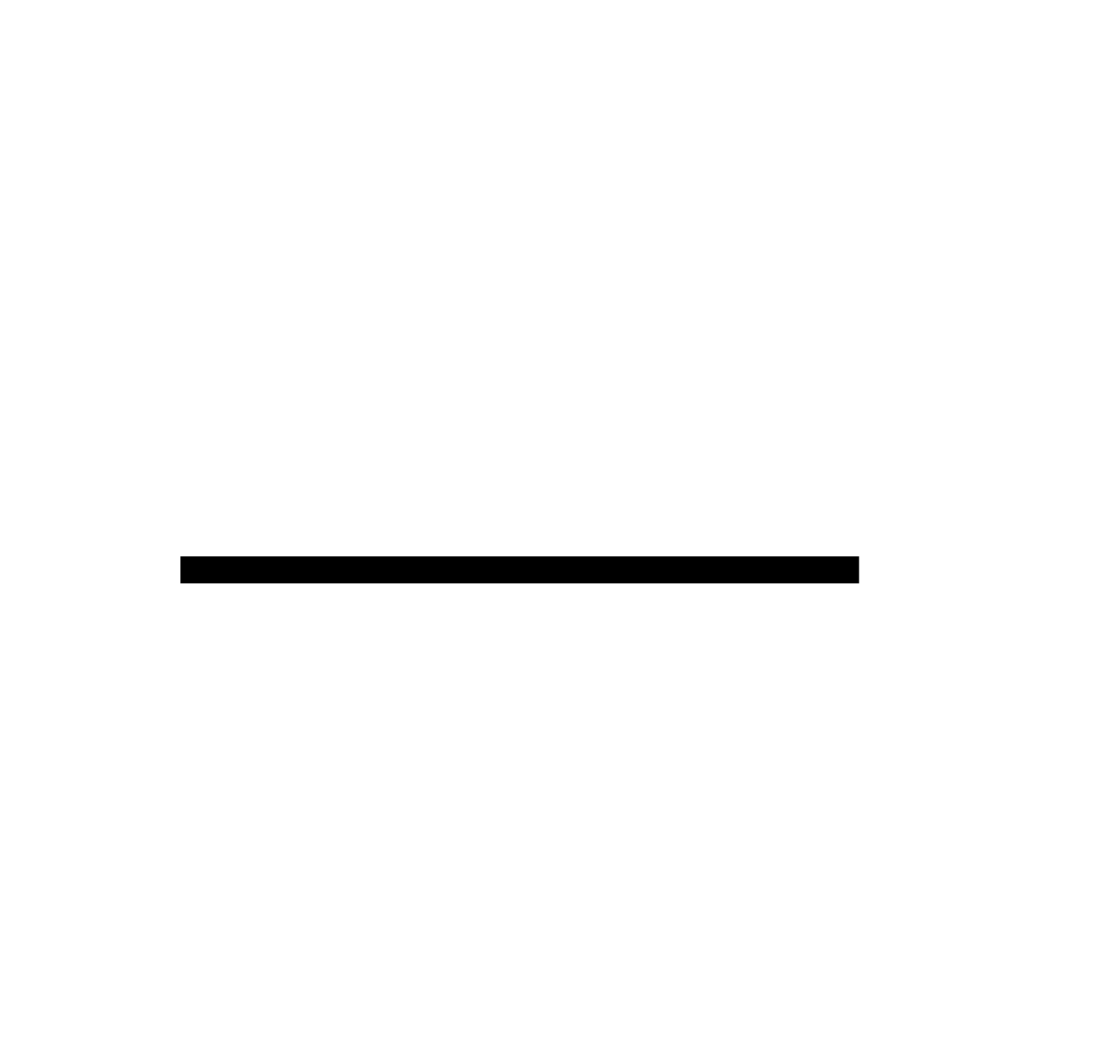} & \includegraphics[width=0.2\linewidth]{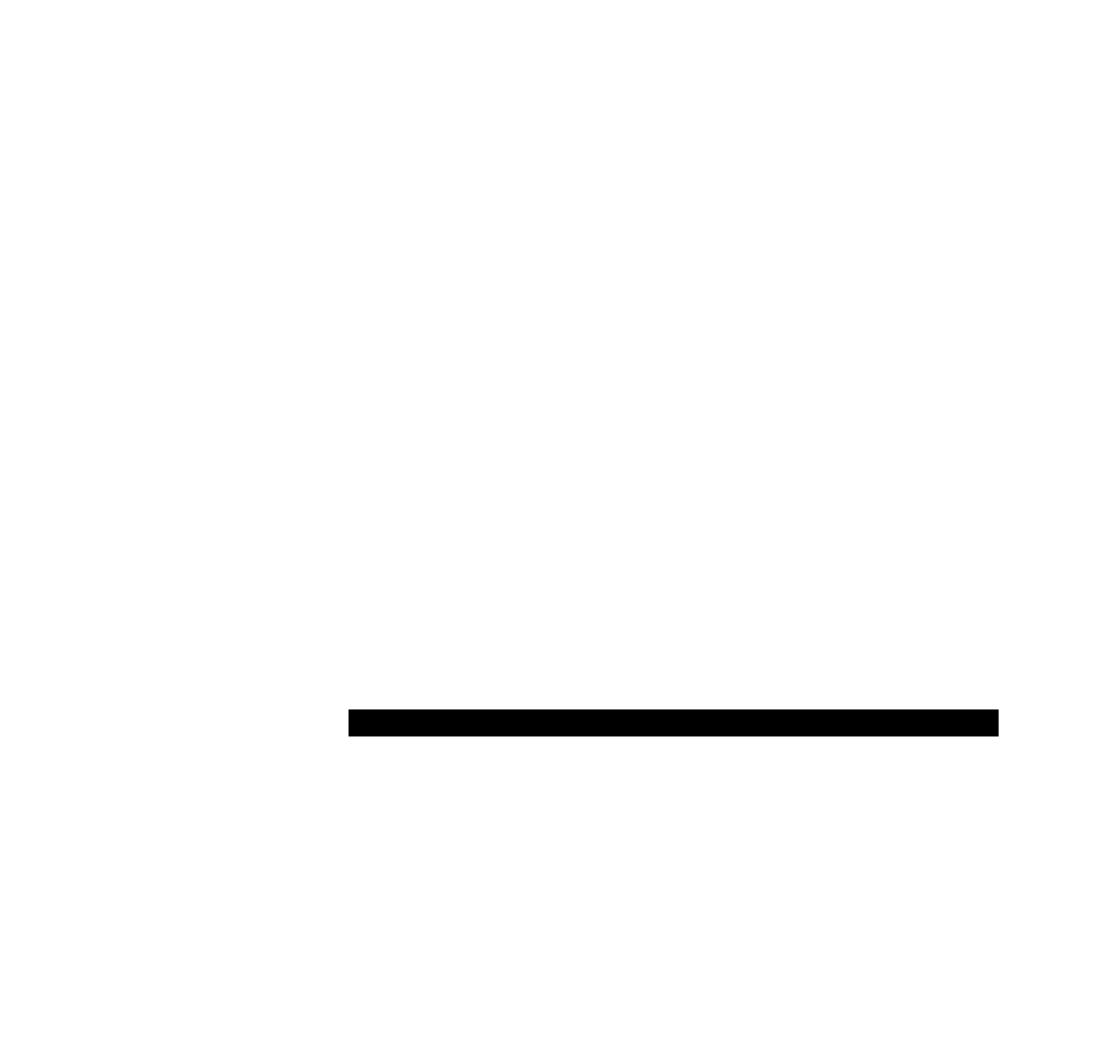} \\
\includegraphics[width=0.2\linewidth]{original_image_32.pdf} & \includegraphics[width=0.2\linewidth]{reconstructed_image_32.pdf} & \includegraphics[width=0.2\linewidth]{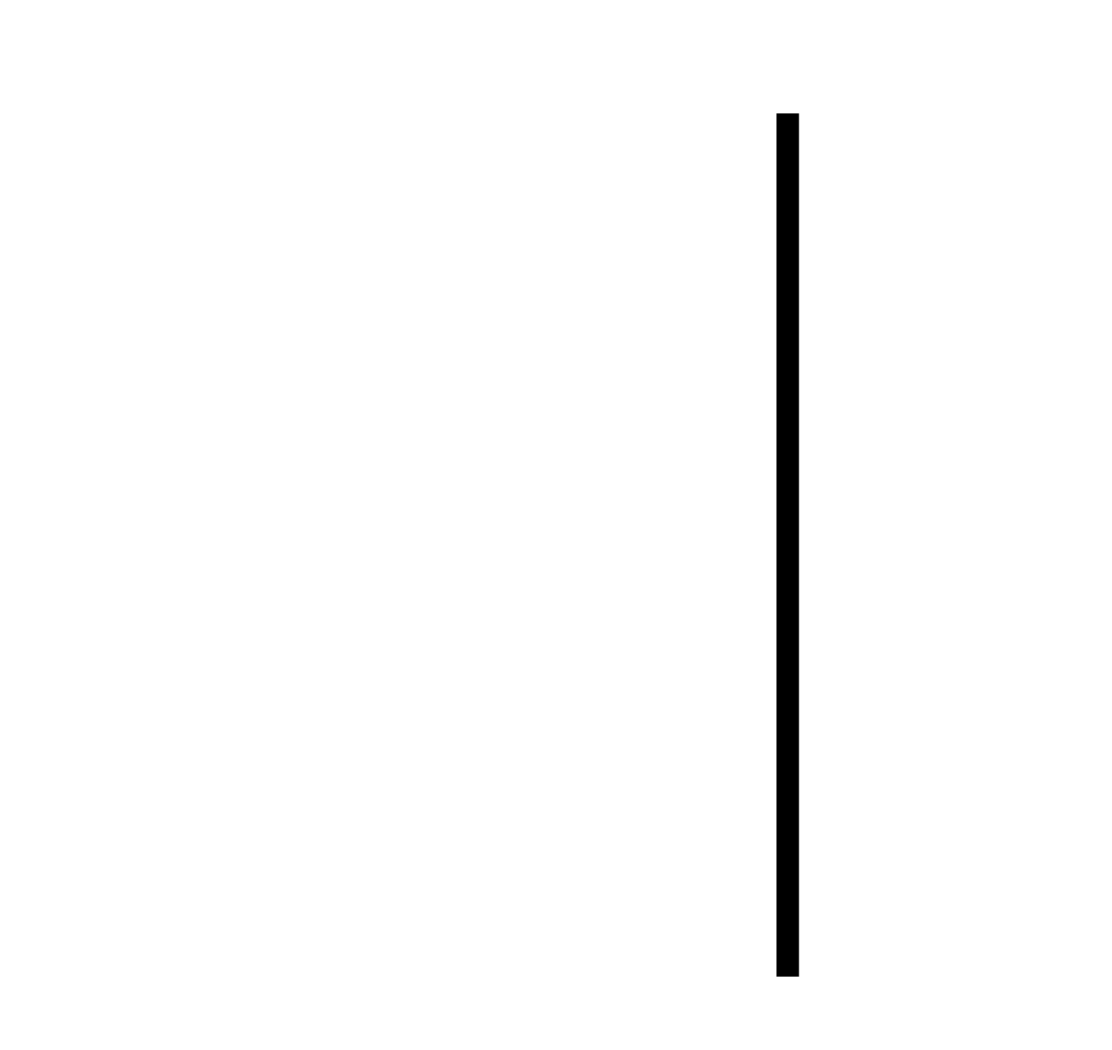} & \includegraphics[width=0.2\linewidth]{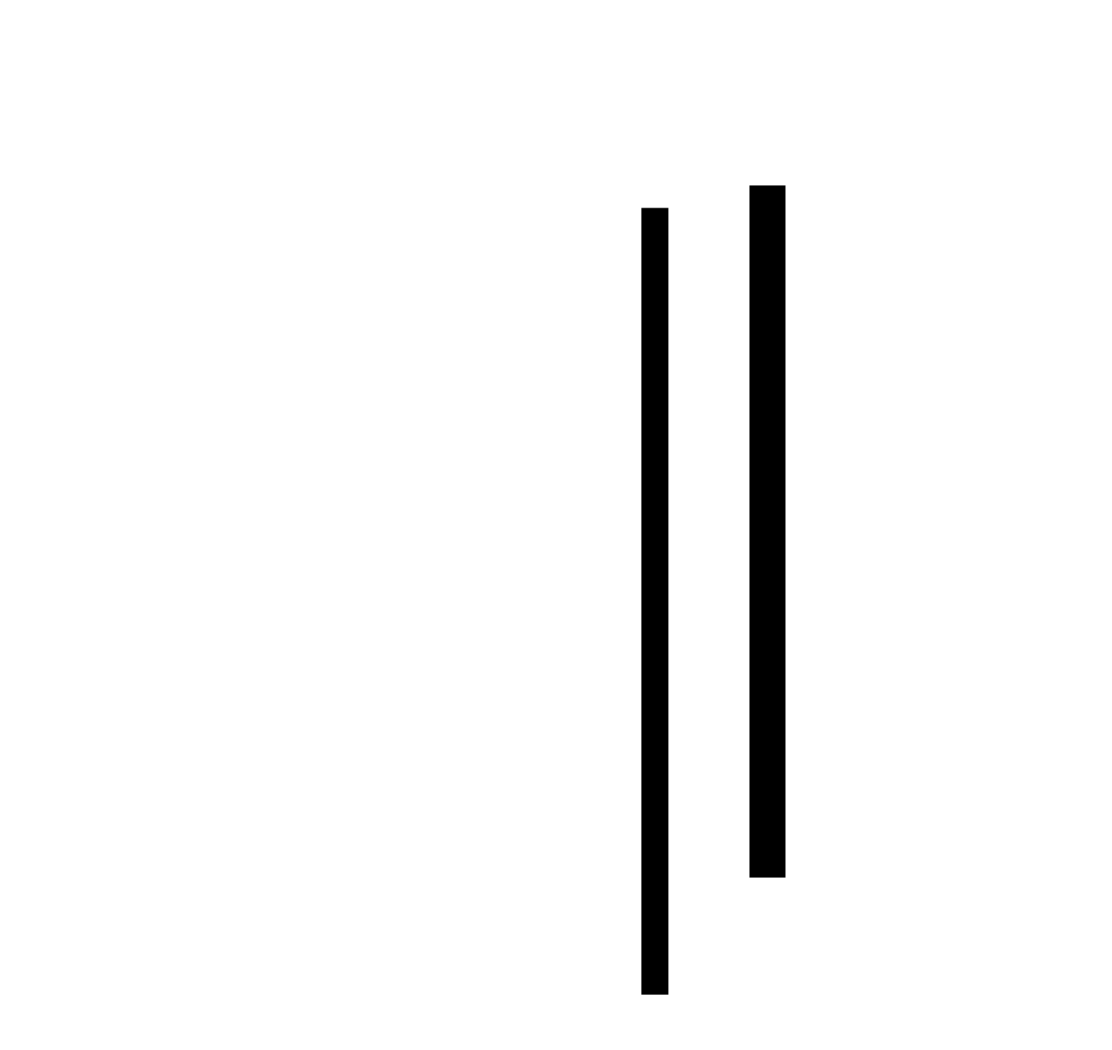} \\
\includegraphics[width=0.2\linewidth]{original_image_30.pdf} & \includegraphics[width=0.2\linewidth]{reconstructed_image_30.pdf} & \includegraphics[width=0.2\linewidth]{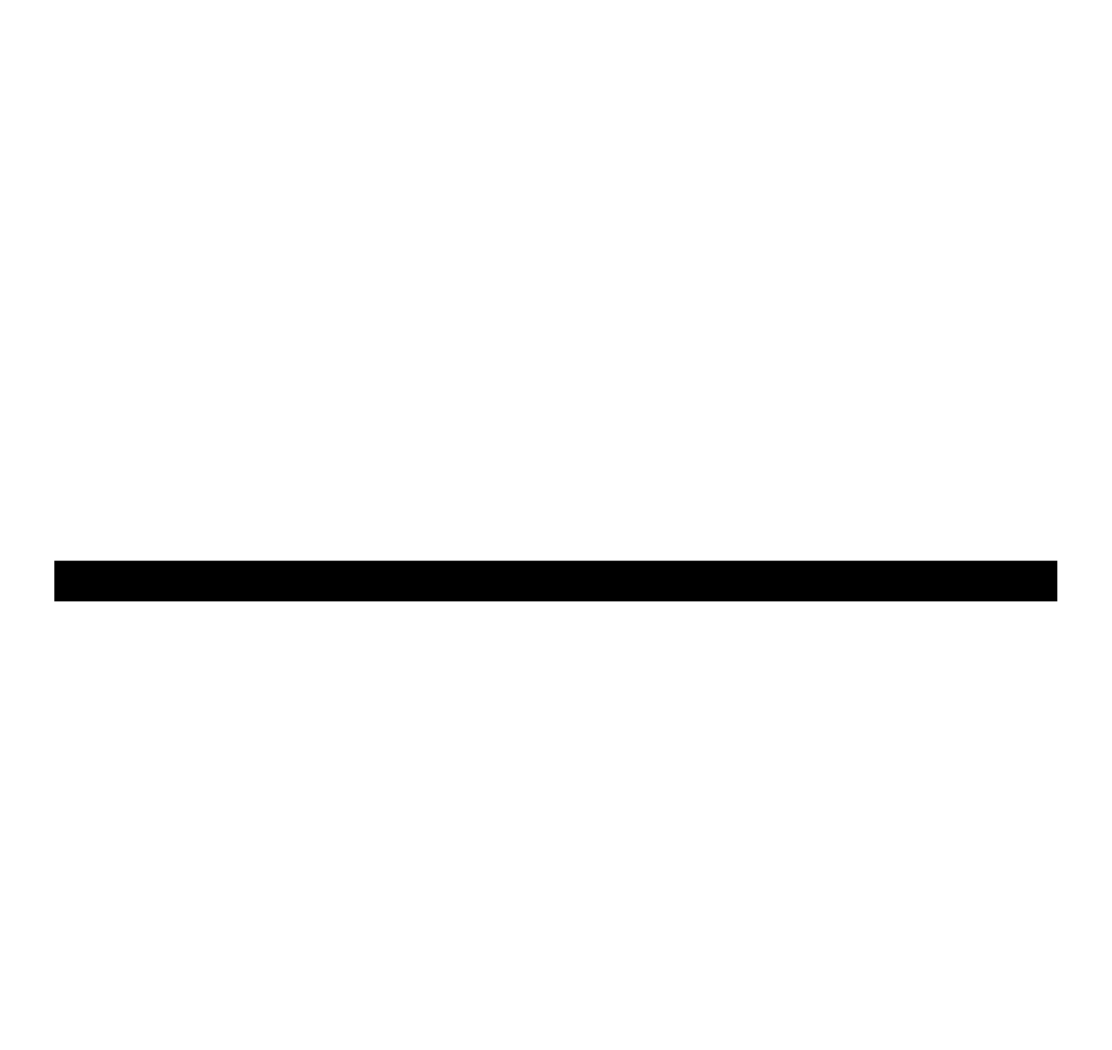} & \includegraphics[width=0.2\linewidth]{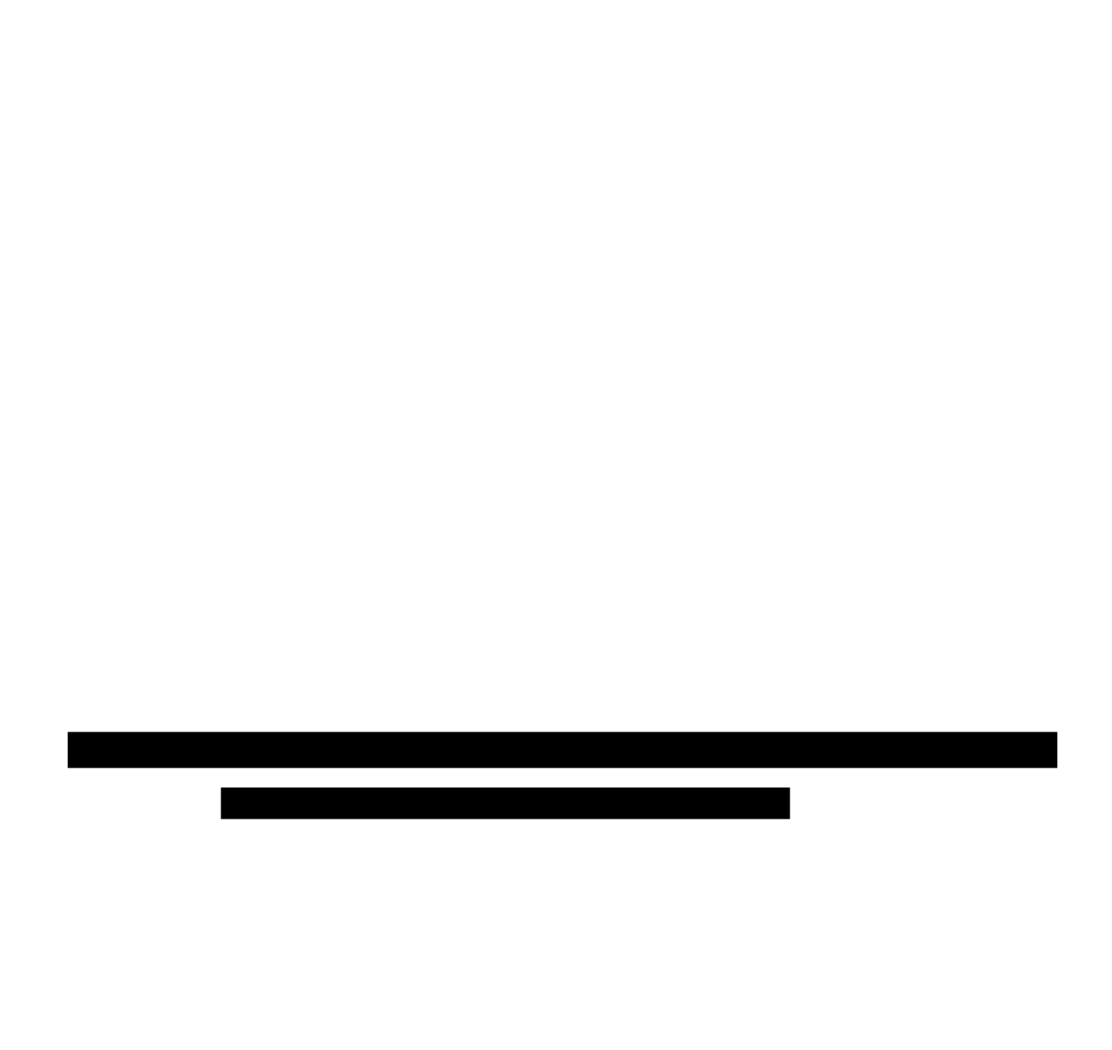} \\
\includegraphics[width=0.2\linewidth]{original_image_25.pdf} & \includegraphics[width=0.2\linewidth]{reconstructed_image_25.pdf} & \includegraphics[width=0.2\linewidth]{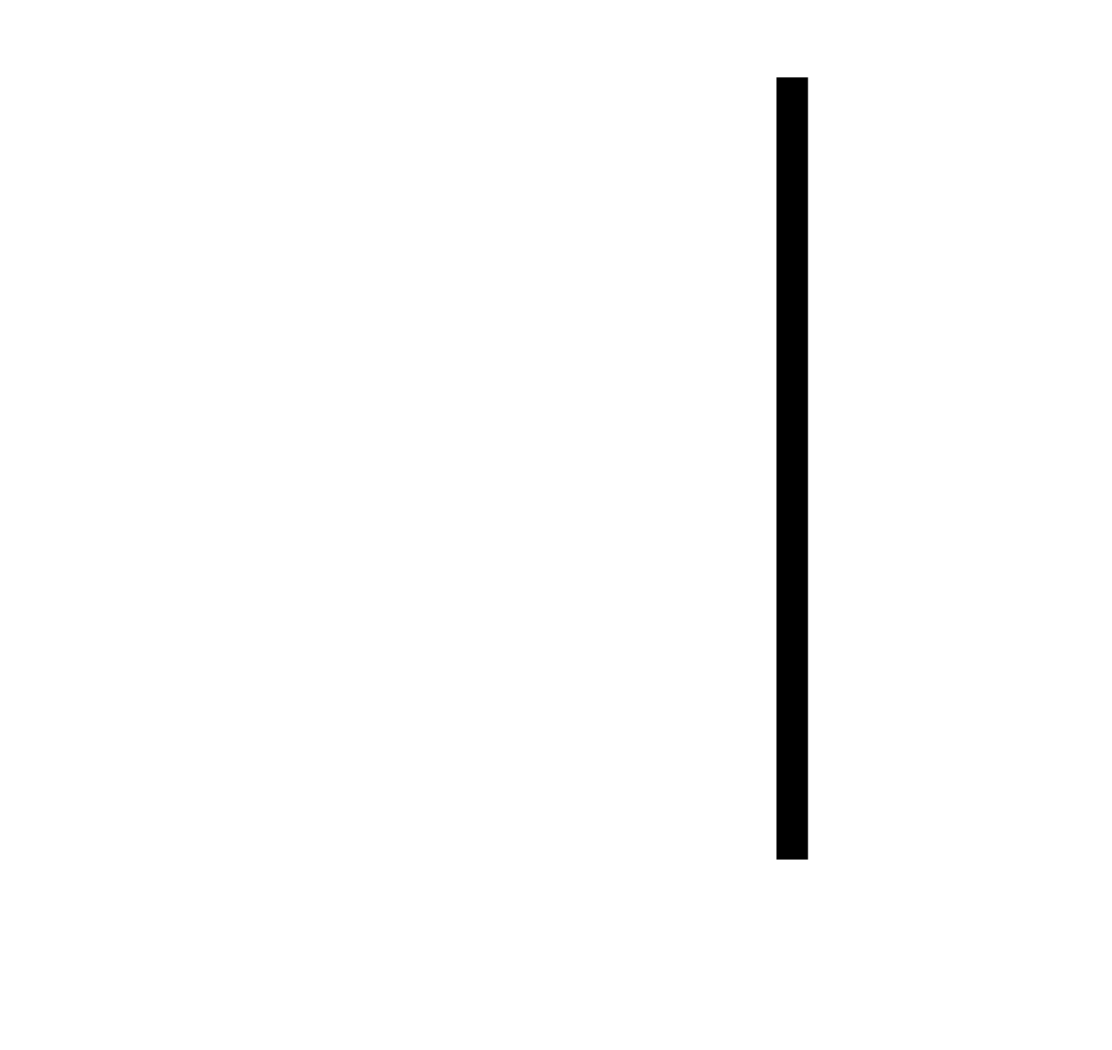} & \includegraphics[width=0.2\linewidth]{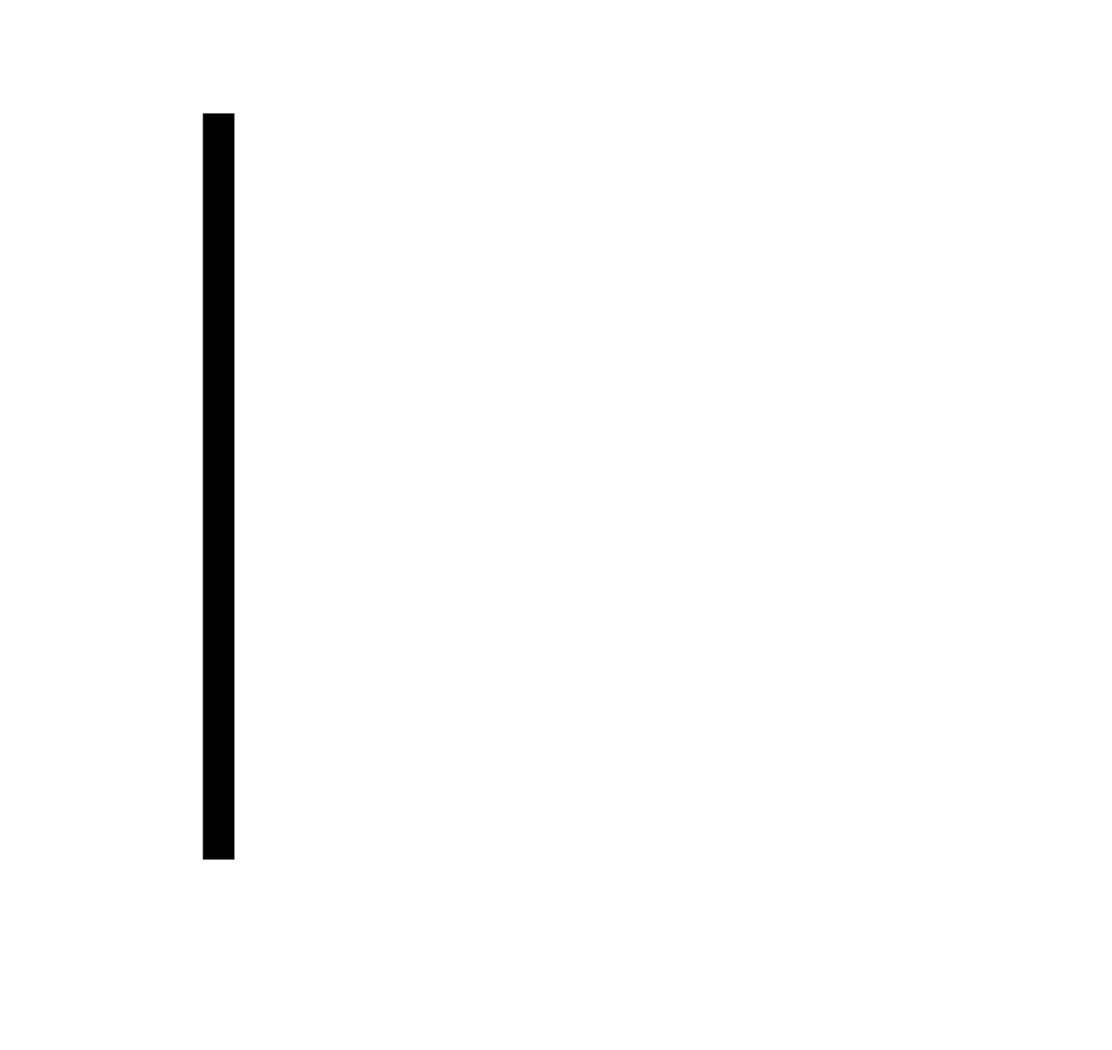} \\
\hline
\end{tabular}
\end{center}
\end{table}

\begin{table}[H]
\caption{Original, reconstructed, and spatial statistics difference}
\begin{center}
\begin{tabular}{cccc}
\hline
\multicolumn{2}{c}{\bf Data} & \multicolumn{2}{c}{\bf Most similar images} \\
\hline
{\bf Original} & {\bf Reconstructed} & {\bf Based on image MSE} & {\bf Based on auto-corr. MSE} \\
\hline
\includegraphics[width=0.2\linewidth]{original_image_26.pdf} & \includegraphics[width=0.2\linewidth]{reconstructed_image_26.pdf} & \includegraphics[width=0.2\linewidth]{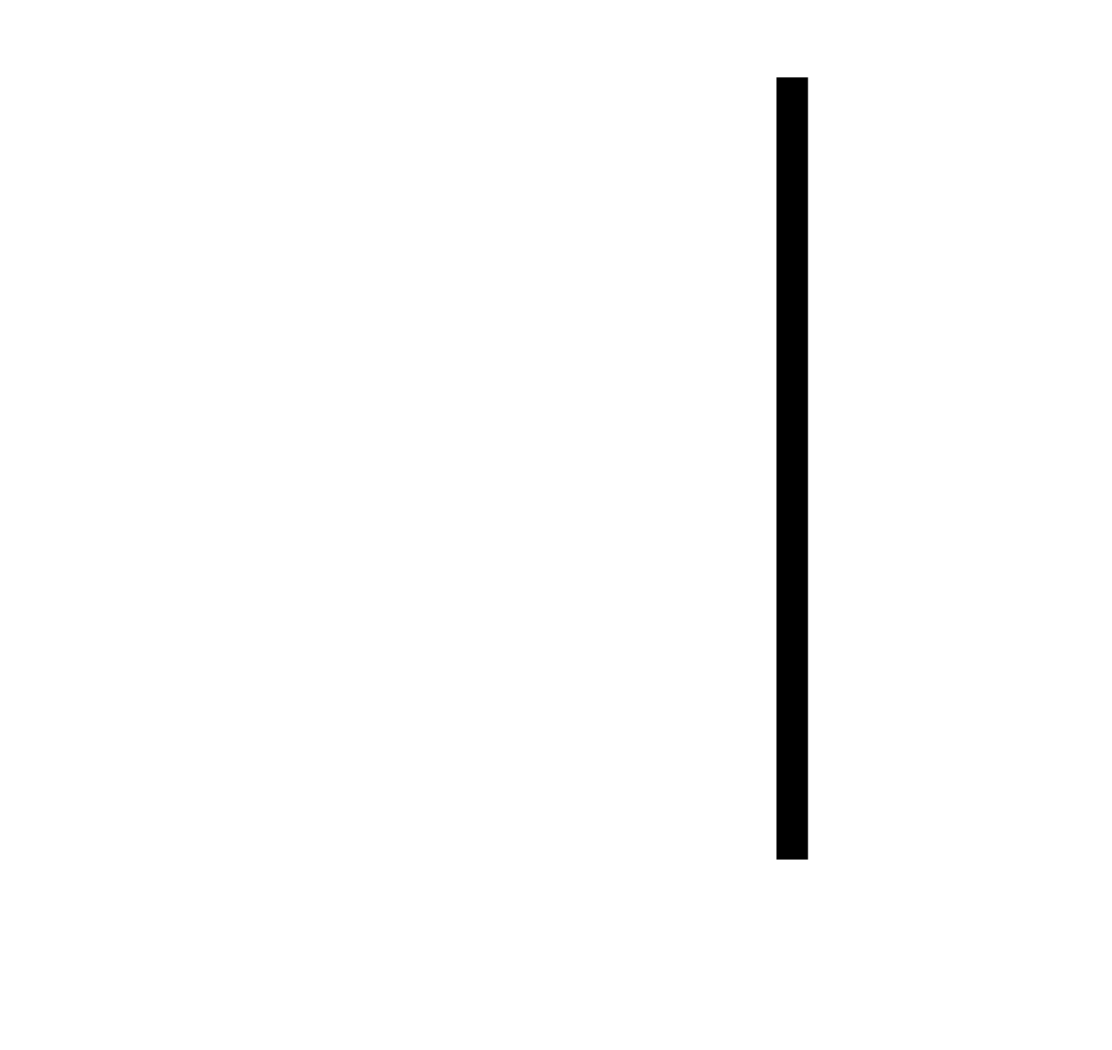} & \includegraphics[width=0.2\linewidth]{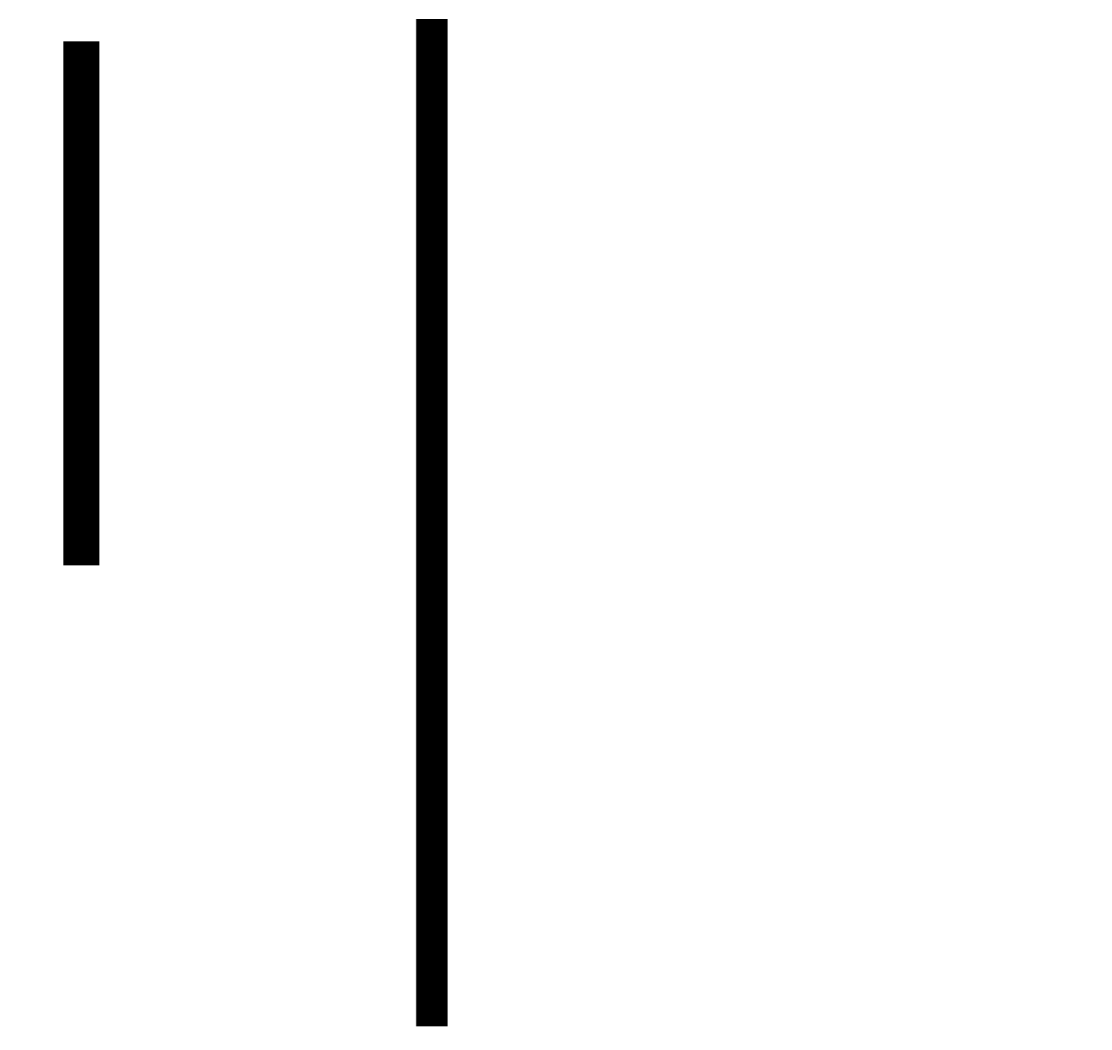} \\
\includegraphics[width=0.2\linewidth]{original_image_24.pdf} & \includegraphics[width=0.2\linewidth]{reconstructed_image_24.pdf} & \includegraphics[width=0.2\linewidth]{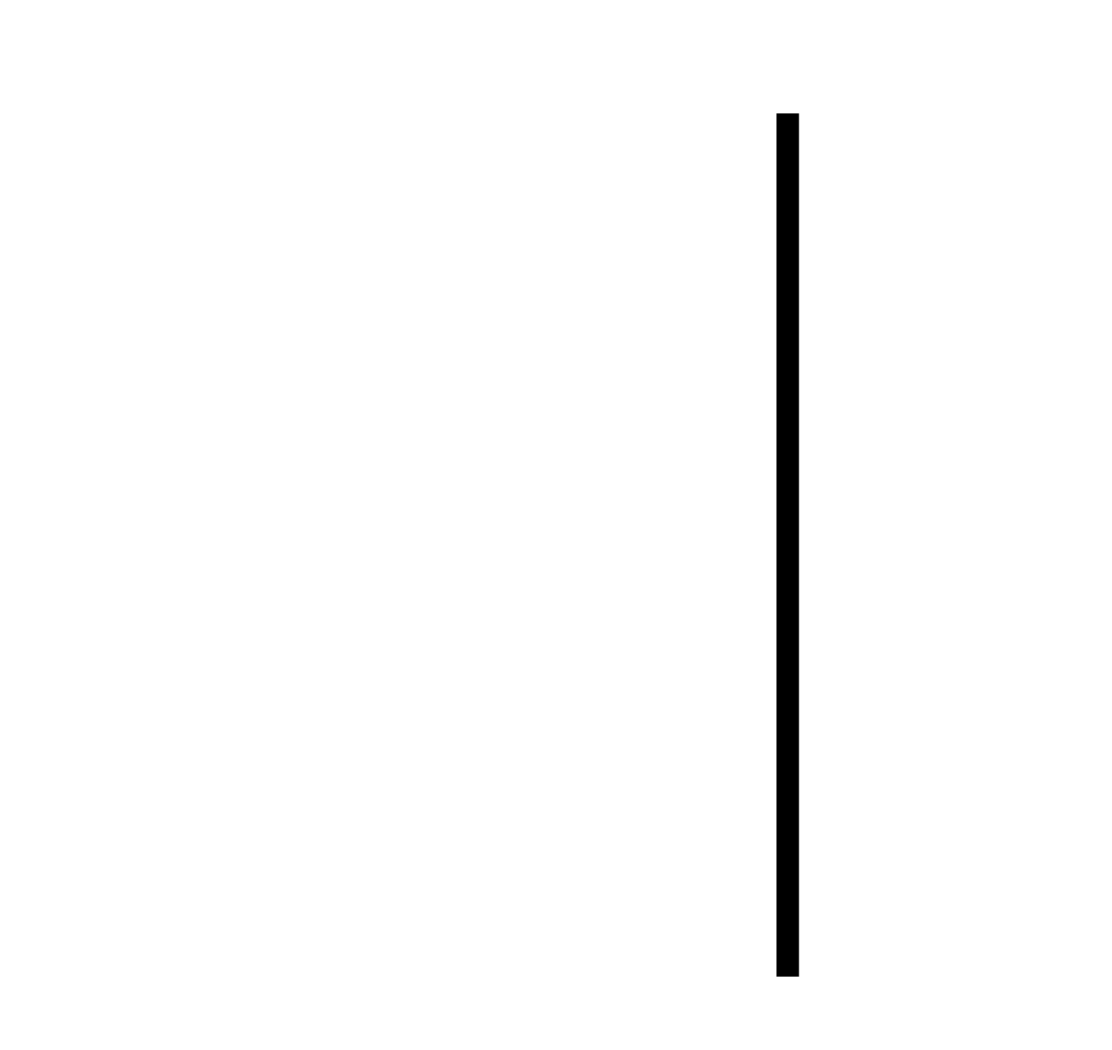} & \includegraphics[width=0.2\linewidth]{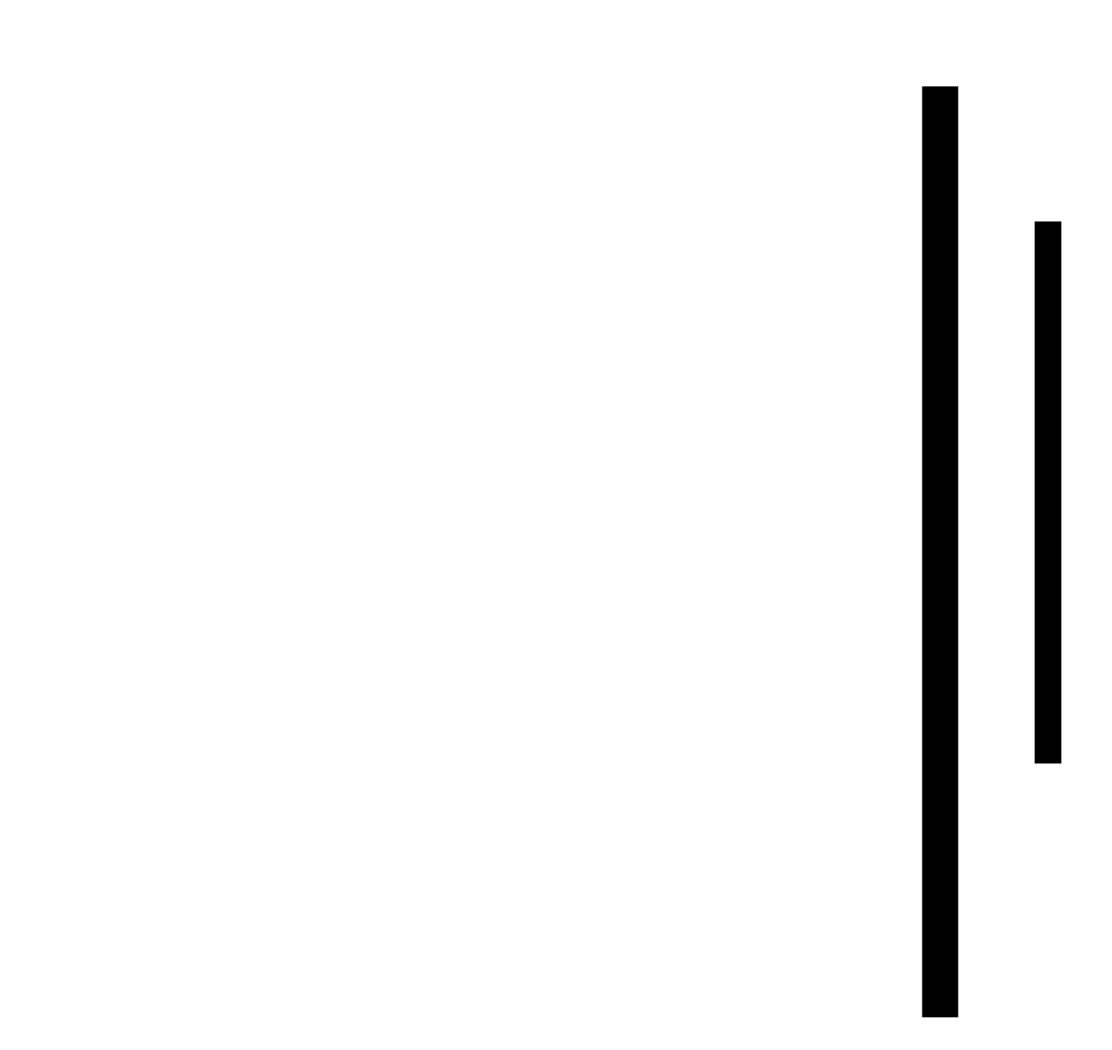} \\
\includegraphics[width=0.2\linewidth]{original_image_19.pdf} & \includegraphics[width=0.2\linewidth]{reconstructed_image_19.pdf} & \includegraphics[width=0.2\linewidth]{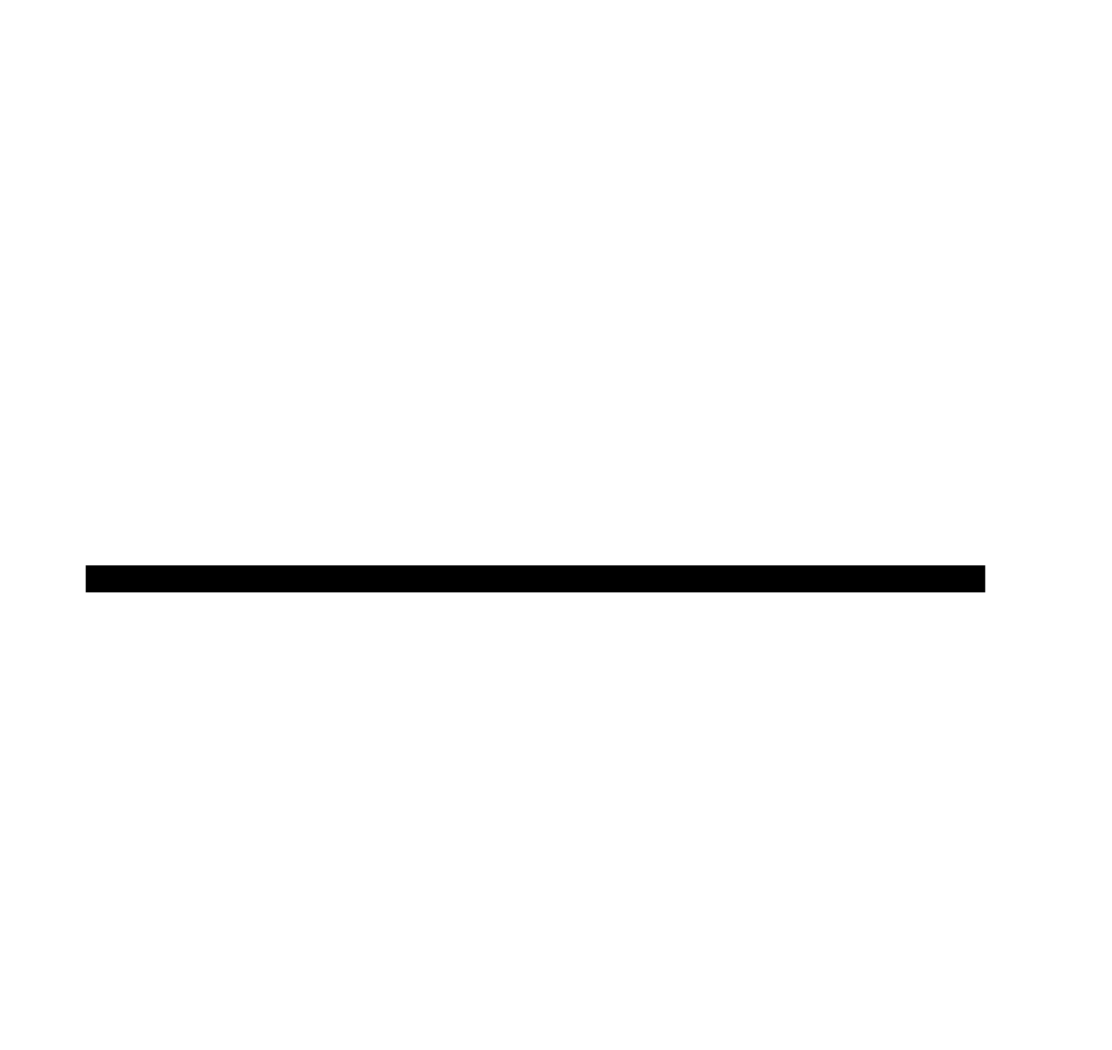} & \includegraphics[width=0.2\linewidth]{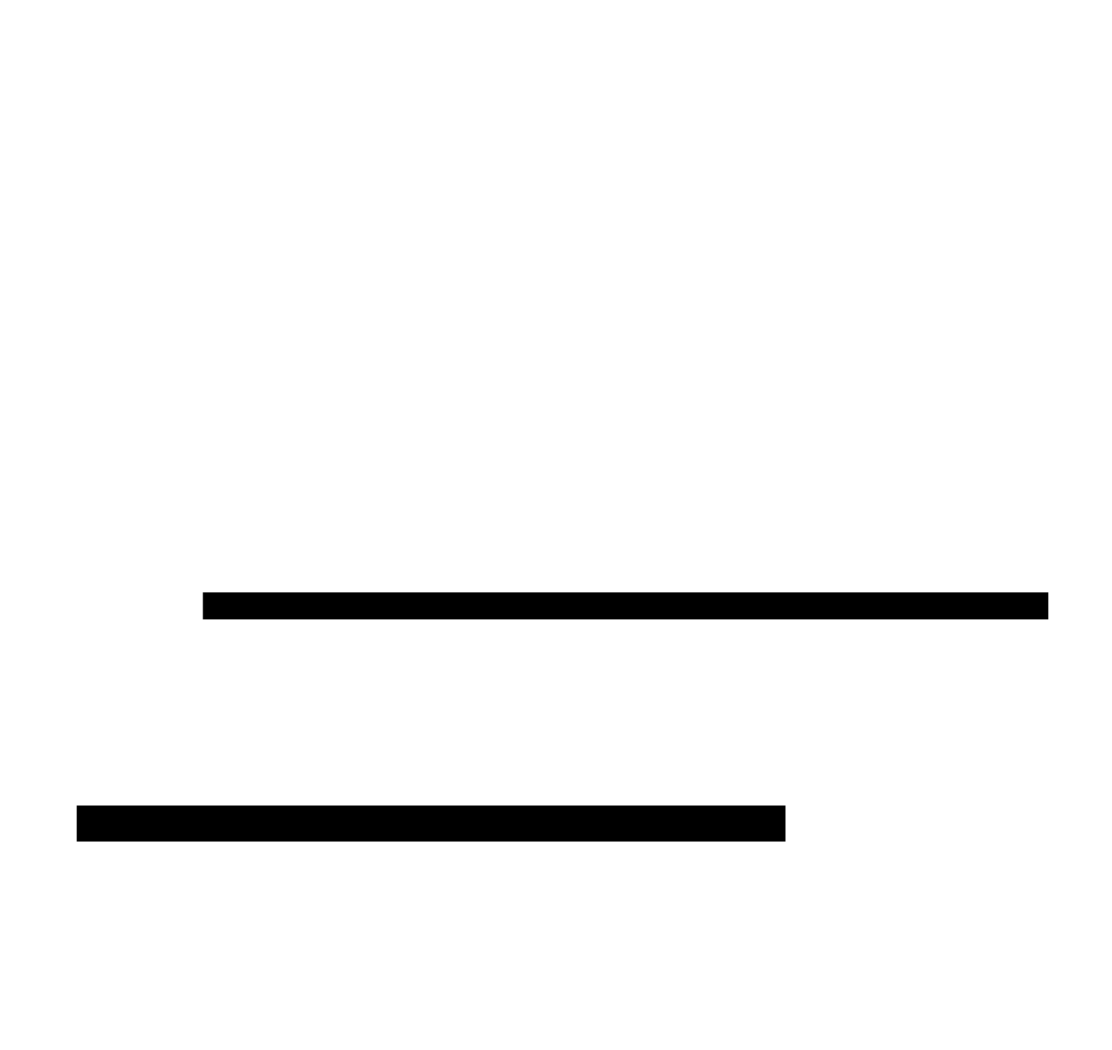} \\
\includegraphics[width=0.2\linewidth]{original_image_23.pdf} & \includegraphics[width=0.2\linewidth]{reconstructed_image_23.pdf} & \includegraphics[width=0.2\linewidth]{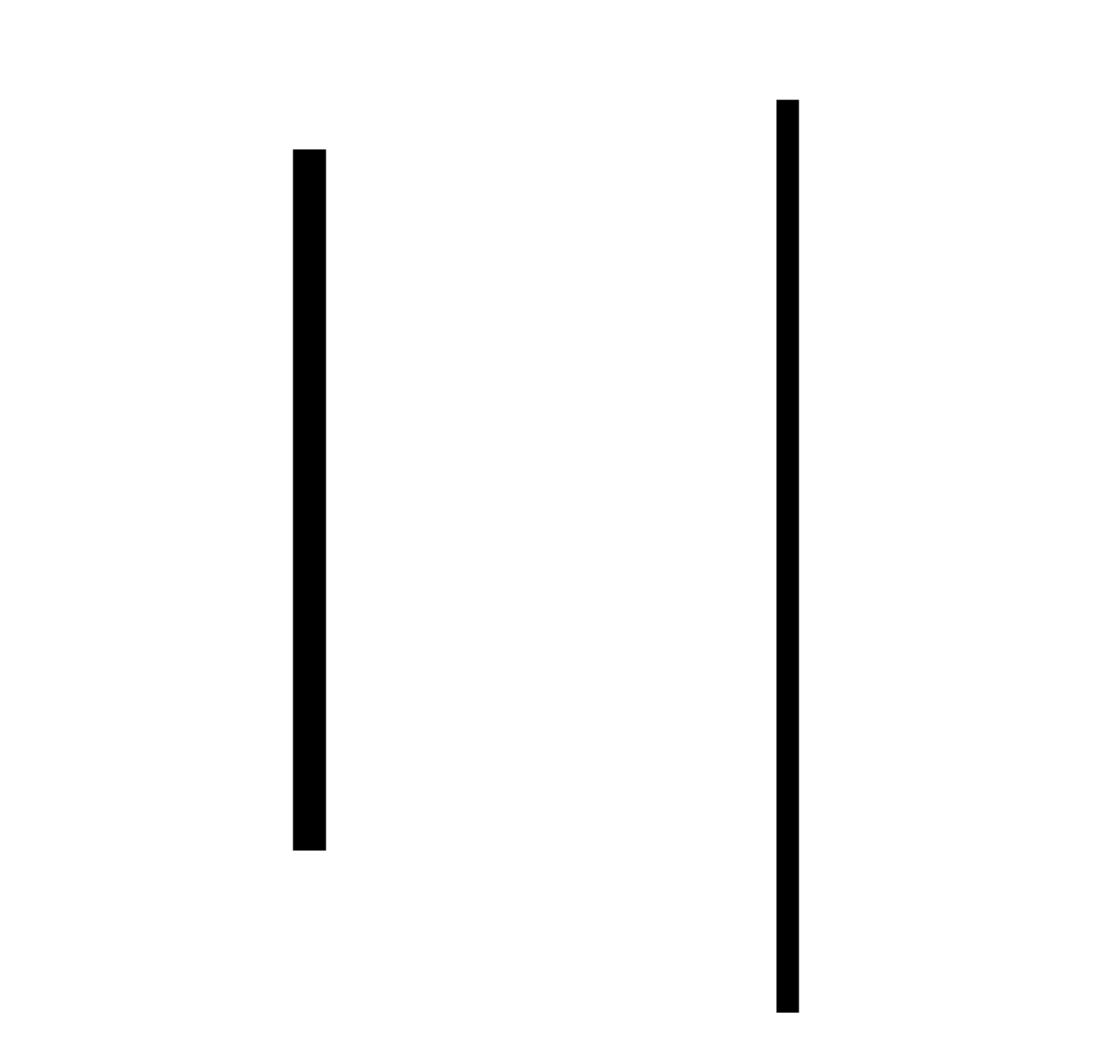} & \includegraphics[width=0.2\linewidth]{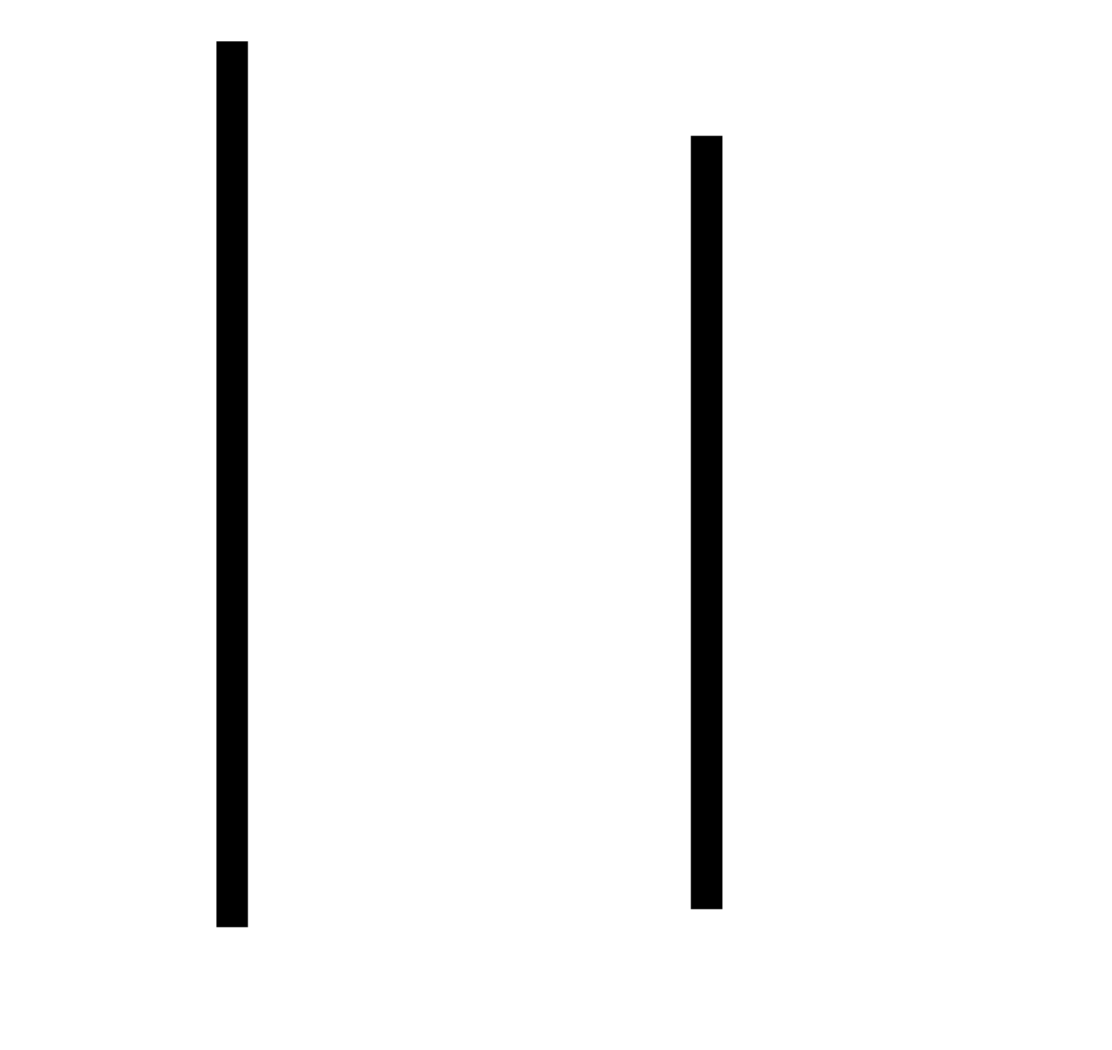} \\
\includegraphics[width=0.2\linewidth]{original_image_17.pdf} & \includegraphics[width=0.2\linewidth]{reconstructed_image_17.pdf} & \includegraphics[width=0.2\linewidth]{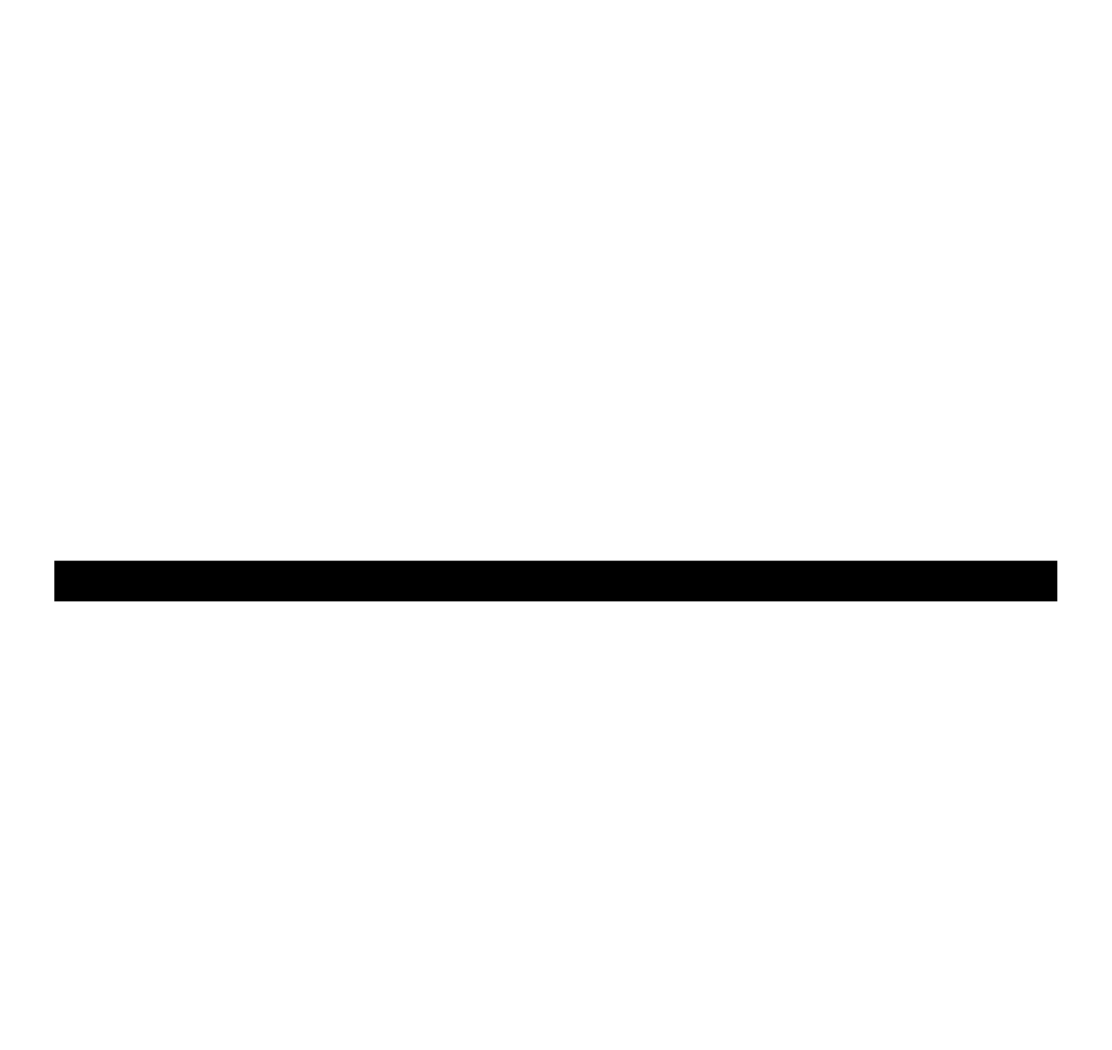} & \includegraphics[width=0.2\linewidth]{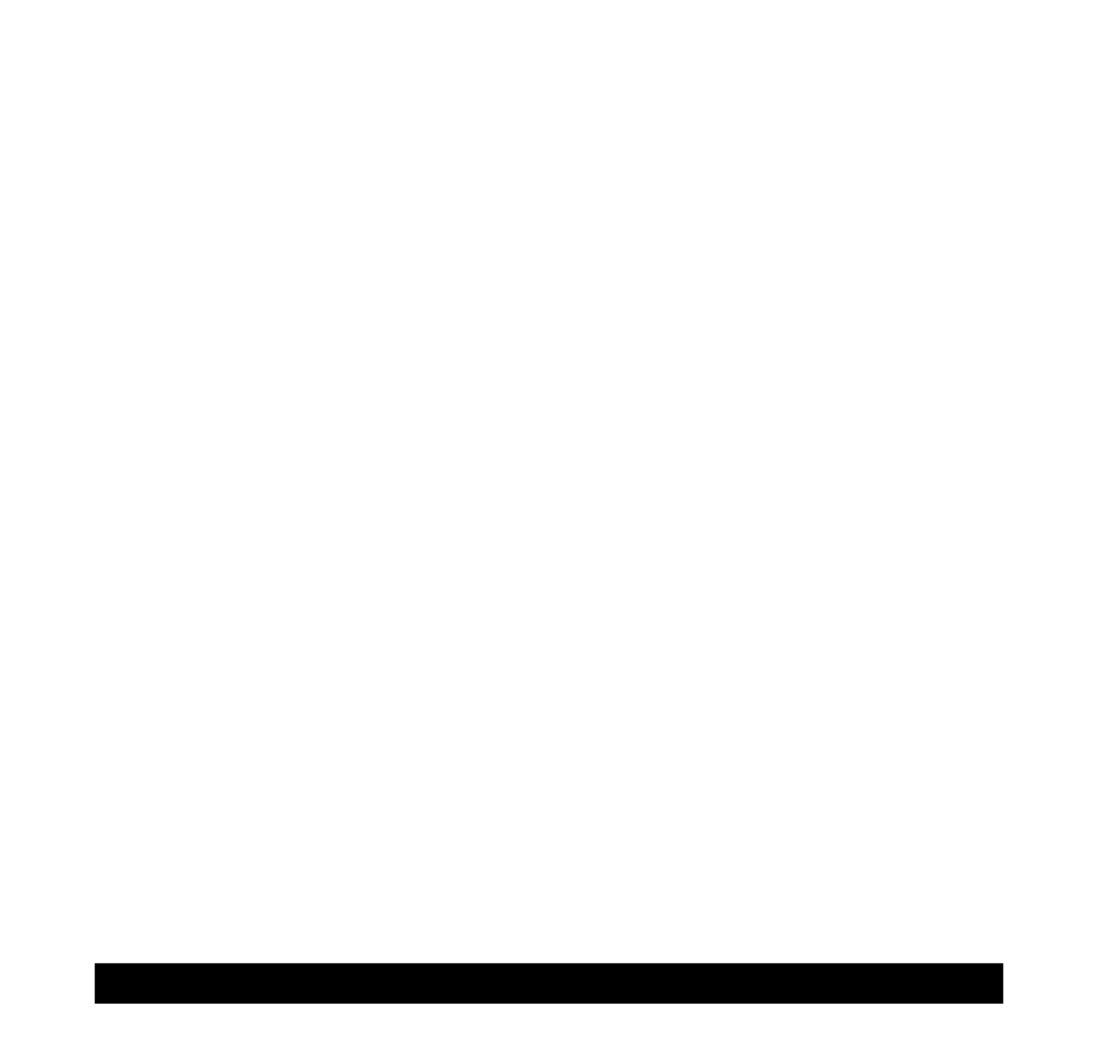} \\
\includegraphics[width=0.2\linewidth]{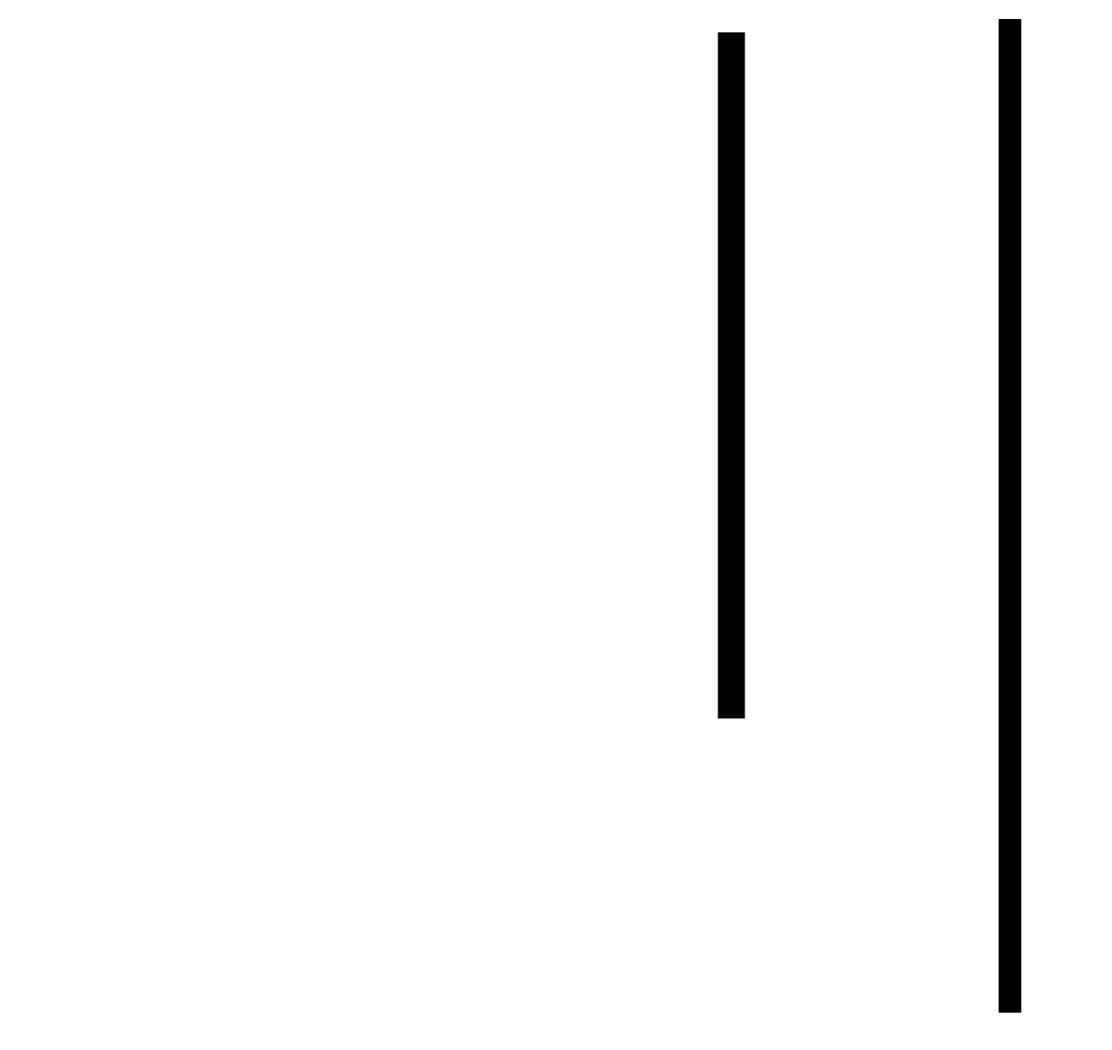} & \includegraphics[width=0.2\linewidth]{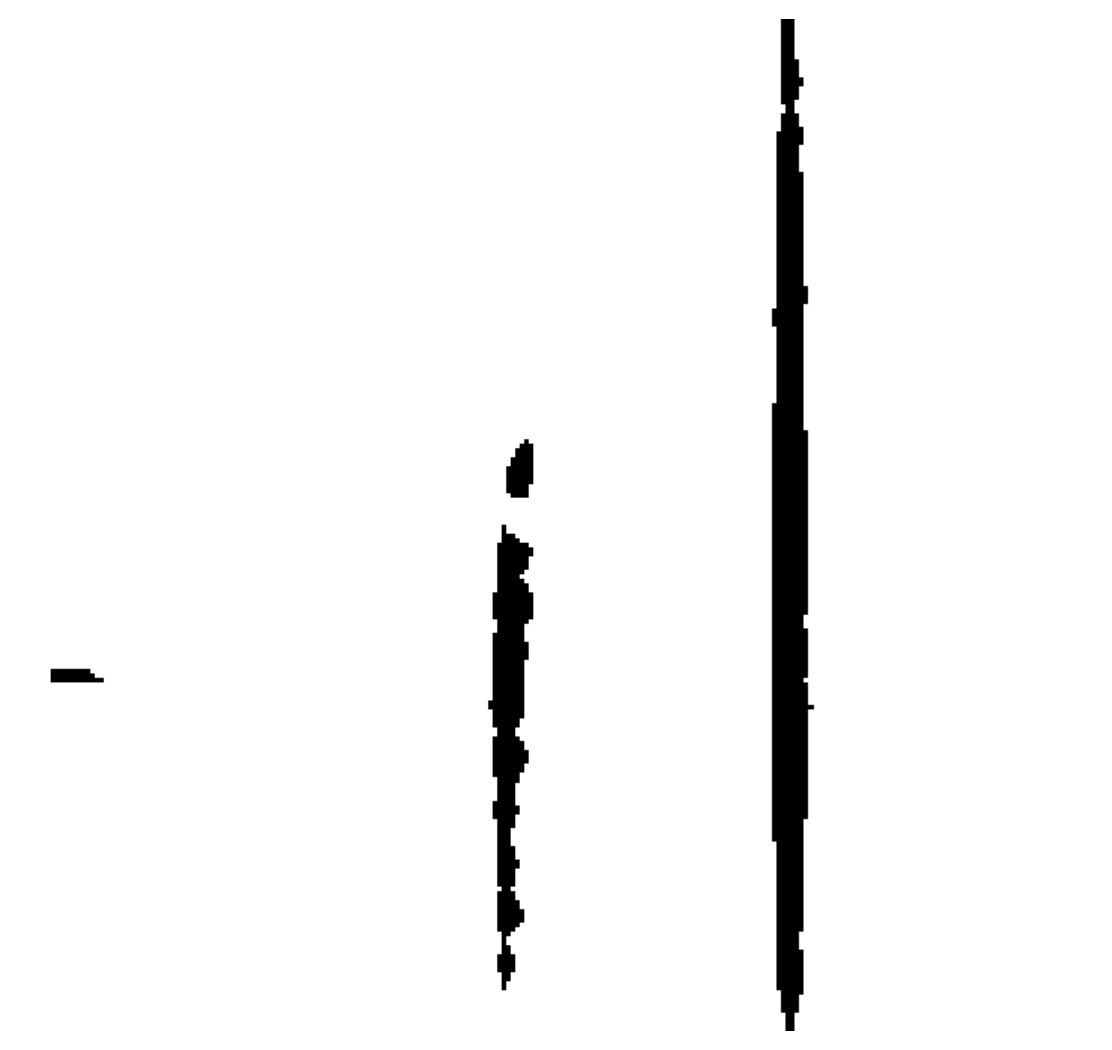} & \includegraphics[width=0.2\linewidth]{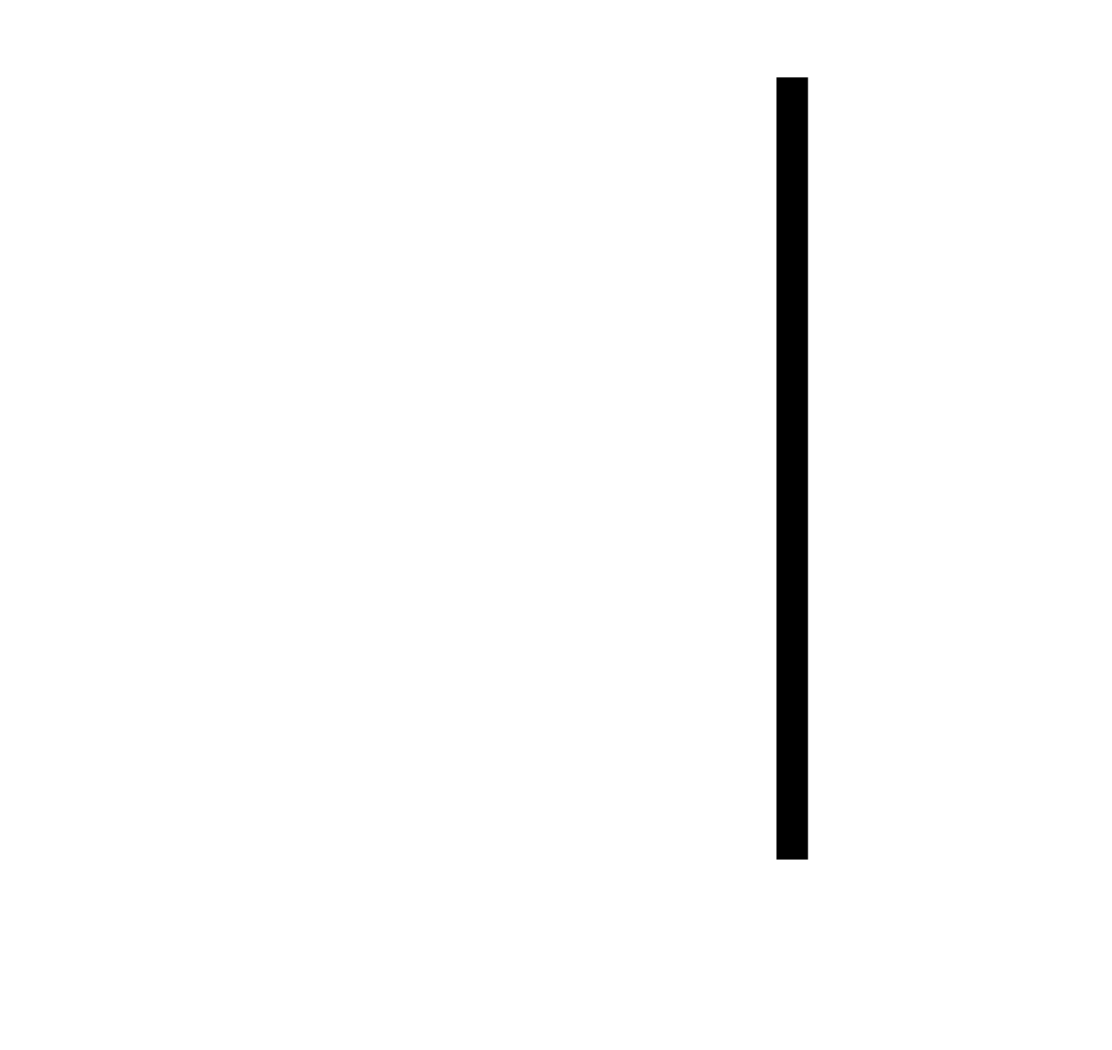} & \includegraphics[width=0.2\linewidth]{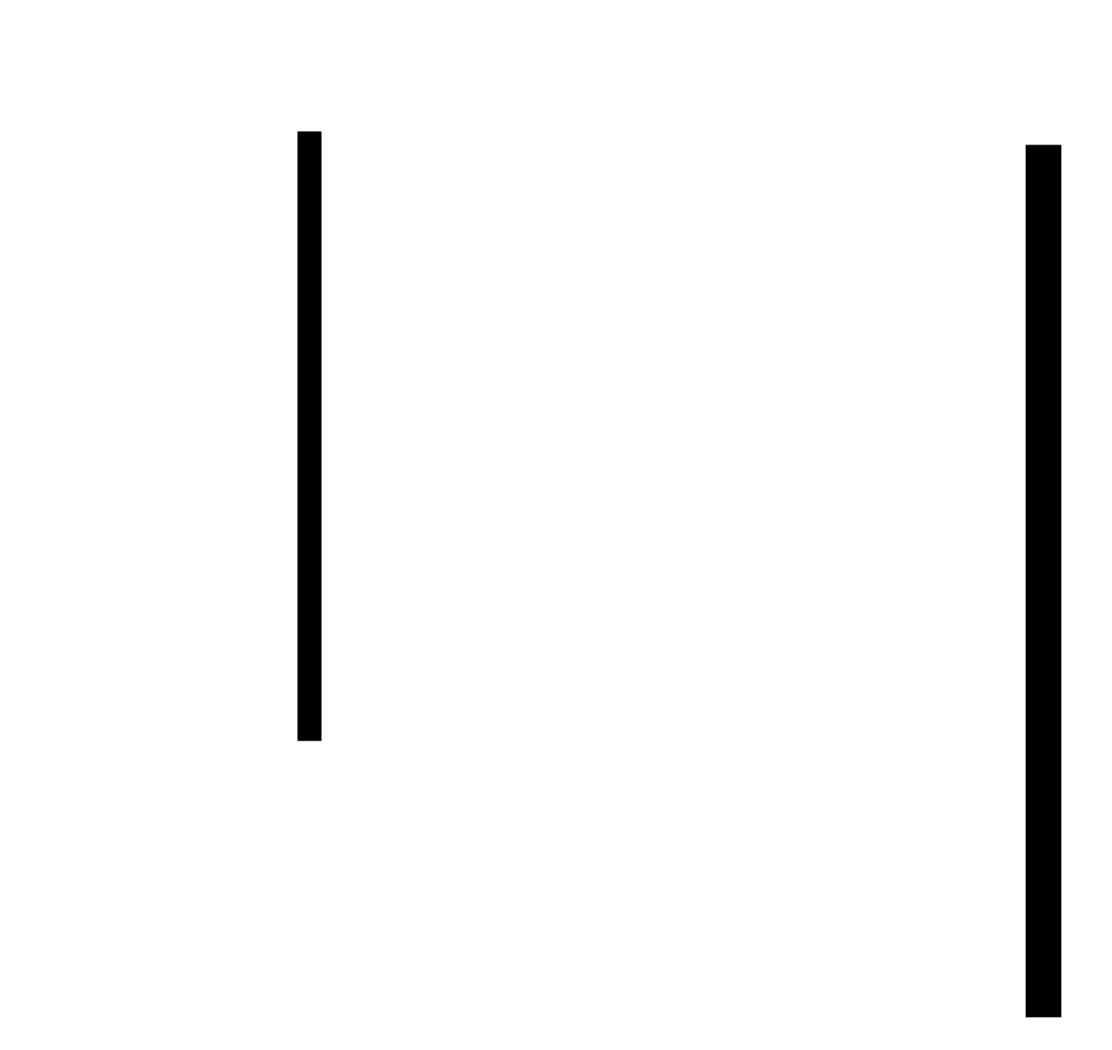} \\
\includegraphics[width=0.2\linewidth]{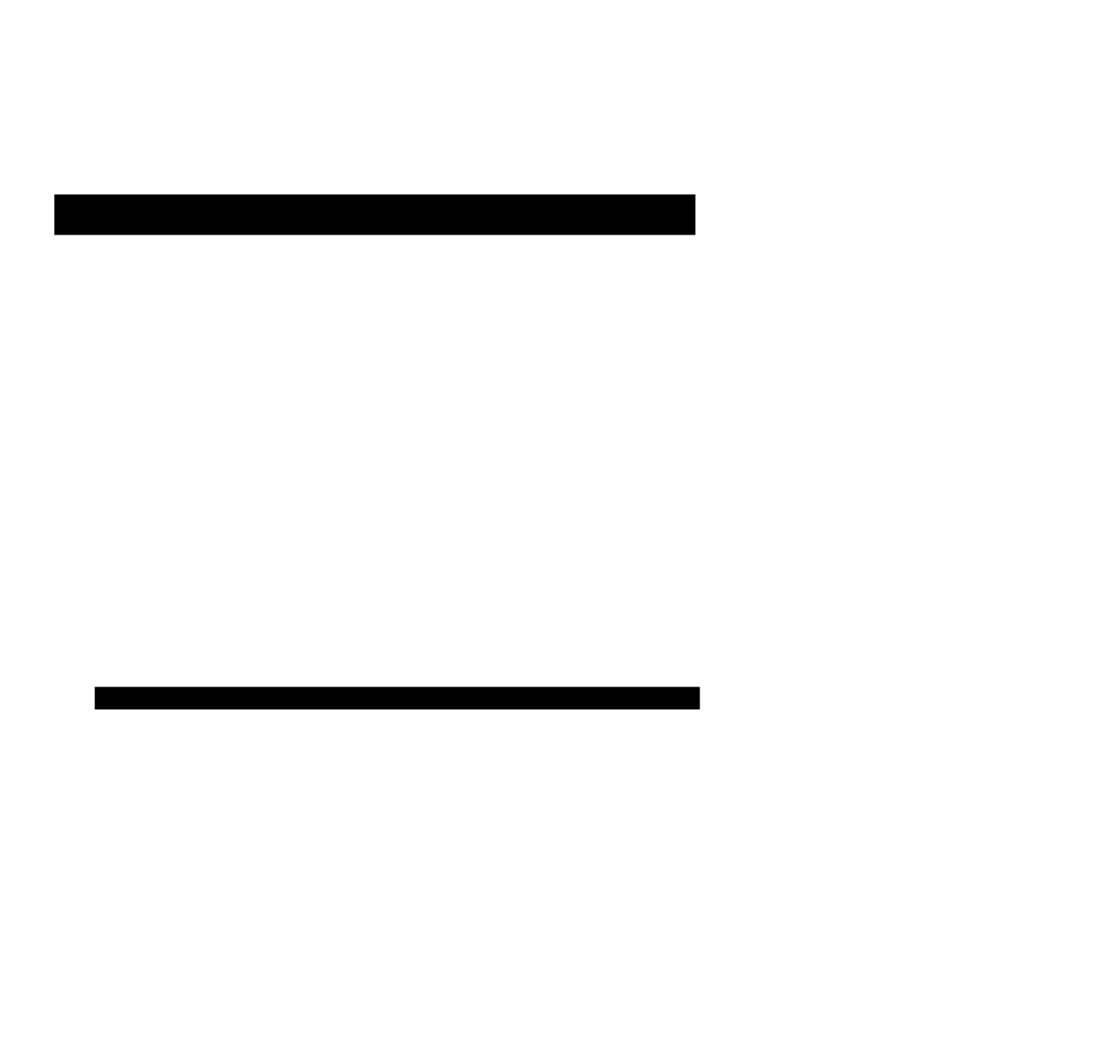} & \includegraphics[width=0.2\linewidth]{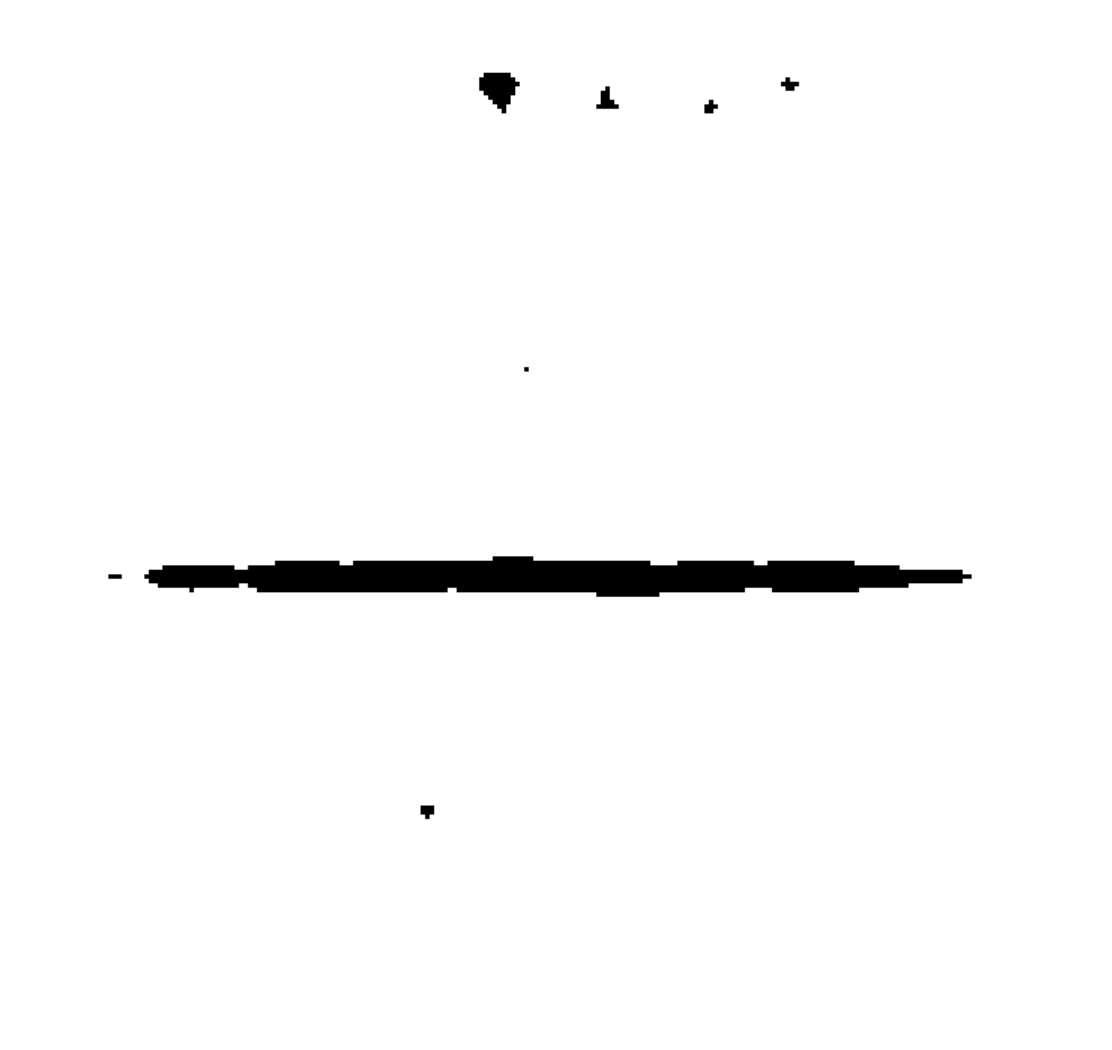} & \includegraphics[width=0.2\linewidth]{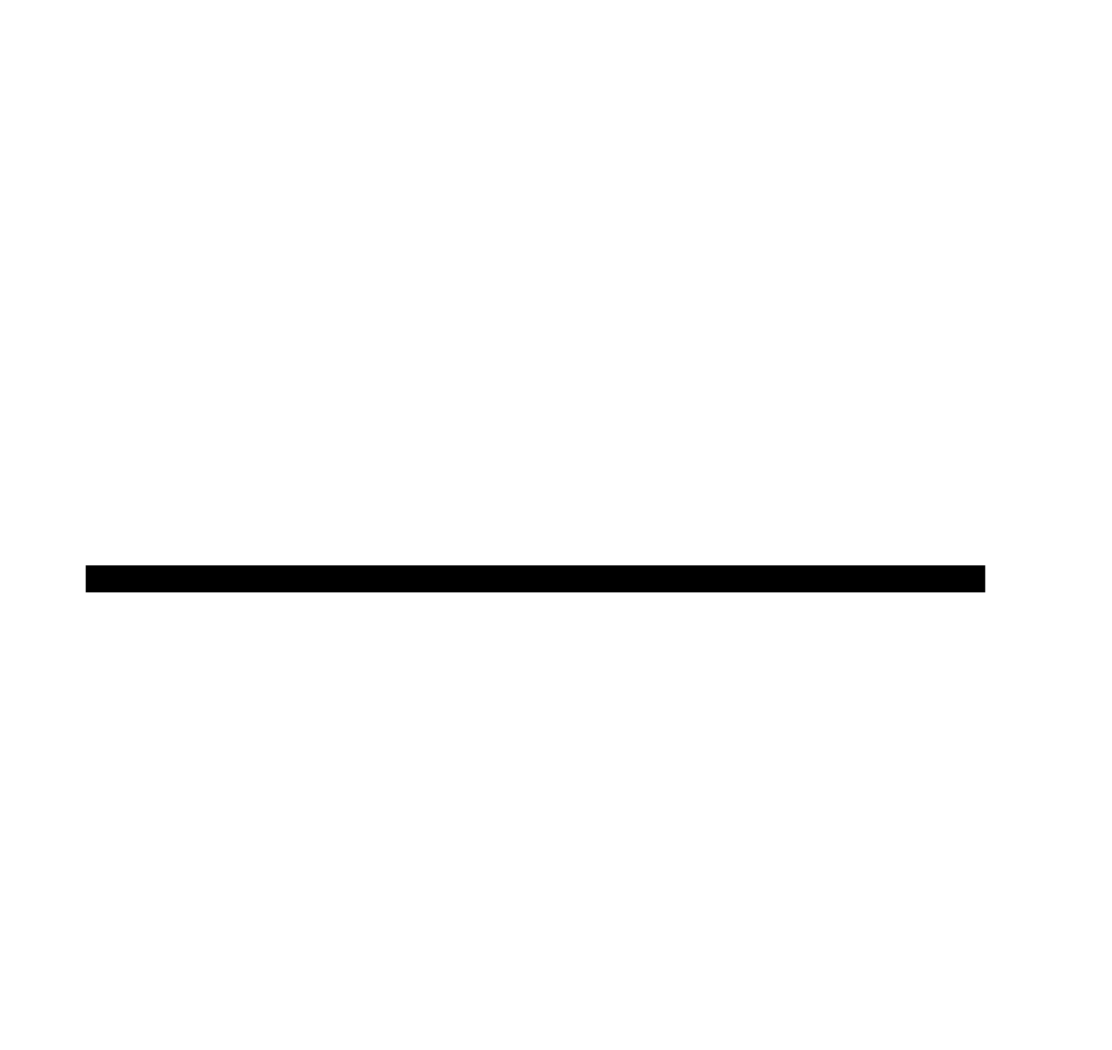} & \includegraphics[width=0.2\linewidth]{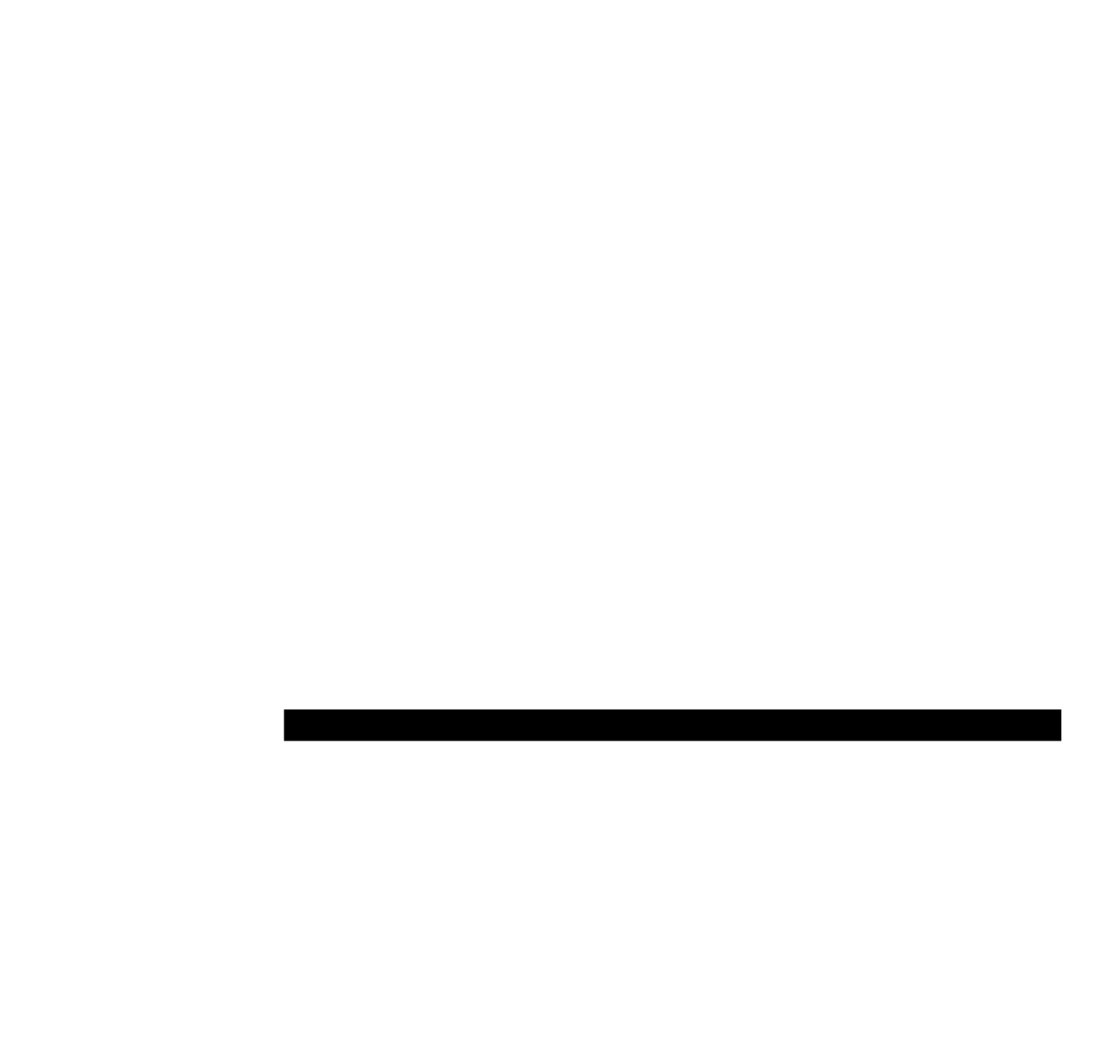} \\
\hline
\end{tabular}
\end{center}
\end{table}
\end{document}